\documentclass{article} 
\usepackage{iclr2026_conference,times}

\usepackage{amsmath,amsfonts,bm}

\def\eqref#1{equation~\ref{#1}}

\def\1{\bm{1}}

\DeclareMathAlphabet{\mathsfit}{\encodingdefault}{\sfdefault}{m}{sl}
\SetMathAlphabet{\mathsfit}{bold}{\encodingdefault}{\sfdefault}{bx}{n}

\usepackage{booktabs}
\usepackage{hyperref}
\usepackage{url}
\usepackage{amsfonts}
\usepackage{amssymb} 
\usepackage[table,xcdraw]{xcolor}
\usepackage{graphicx} 
\usepackage{arydshln}
\usepackage{multirow}
\usepackage{wrapfig} 
\usepackage{float} 
\usepackage{caption}
\usepackage{rotating}
\usepackage{subcaption}  
\usepackage{tabularx}    
\newcolumntype{C}{>{\centering\arraybackslash}X}
\usepackage{makecell}
\usepackage{enumitem}
\usepackage{etoc} 
\usepackage{titletoc}

\makeatletter
\renewcommand{\@oddhead}{}
\makeatother

\title{Unleashing Scientific Reasoning for Bio-experimental Protocol Generation via Structured Component-based Reward Mechanism}

\author{
Haoran Sun$^{1,2}$, Yankai Jiang$^{1}$\thanks{Corresponding authors.}, 
Zhenyu Tang$^{1,3}$, Yaning Pan$^{1,2}$, Shuang Gu$^{1,3}$, Zekai Lin$^{2}$,\\
\textbf{Lilong Wang$^{1}$, Wenjie Lou$^{1}$, Lei Liu$^{2}$, Lei Bai$^{1}$, 
Xiaosong Wang$^{1}$\footnotemark[1]} \\
$^{1}$Shanghai Artificial Intelligence Laboratory \\
$^{2}$Fudan University \\
$^{3}$Shanghai Jiao Tong University \\
\texttt{\{jiangyankai,wangxiaosong\}@pjlab.org.cn}
}

\iclrfinalcopy 
\begin{document}

\maketitle
\begin{abstract}
The foundation of reproducible science lies in protocols that are precise, logically ordered, and executable. The autonomous generation of these protocols through natural language queries could greatly improve the efficiency of the reproduction process. However, current leading large language models (LLMs) often generate incomplete or inconsistent protocols, limiting their utility. To address this limitation, we first introduce SciRecipe, a large-scale dataset of over 12K structured protocols spanning 27 biological subfields and encompassing both comprehension and problem-solving tasks. To further improve protocol generation, we propose the “Sketch-and-Fill” paradigm, which separates analysis, structuring, and expression to ensure each step is explicit and verifiable. Complementing this, the structured component-based reward mechanism evaluates step granularity, action order, and semantic fidelity, aligning model optimization with experimental reliability. Building on these components, we develop Thoth, trained through a staged Knowledge-to-Action process that progresses from knowledge acquisition to operational reasoning and ultimately to robust, executable protocol generation. Across multiple benchmarks, Thoth consistently surpasses both proprietary and open-source LLMs, achieving significant improvements in step alignment, logical sequencing, and semantic accuracy. Our approach paves the way for reliable scientific assistants that bridge knowledge with experimental execution.
Code will be available at \href{https://github.com/manglu097/Thoth}{https://github.com/manglu097/Thoth}
\end{abstract}

\section{Introduction}
The planning and execution of scientific experiments hinge on protocols that serve not merely as textual guidelines but as operational blueprints detailing procedures, materials, and logical dependencies \citep{freedman2015economics, goodman2016does}. A well-structured protocol ensures that experiments are reproducible, safe, and scientifically valid \citep{national2019reproducibility}, which is essential for cumulative progress in the life sciences. While recent advances in large language models (LLMs) have greatly expanded their role in biomedical research, ranging from literature-based discovery to domain-specific question answering \citep{devlin2019bert, brown2020language, lee2020biobert, beltagy2019scibert}, their ability to produce reliable experimental protocols remains underdeveloped. Existing datasets and benchmarks are typically confined to comprehension tasks \citep{liu2025bioprobench, jiang2024protocode}, thereby neglecting the dimensions of planning and problem solving required to support reproducibility and practical execution. As a result, researchers often find that models offer fragmented recommendations on experimental procedures but fall short of producing concise, logically ordered protocols that can be directly implemented in laboratory workflows.


Currently, proprietary models \citep{achiam2023gpt, comanici2025gemini, jaech2024openai, gpt5chat} such as GPT-5 demonstrate strong capabilities in procedural reasoning, while domain-specific scientific systems complement LLMs with curated knowledge bases or tool-assisted pipelines \citep{huang2025biomni, jin2025stella}. Despite these advances, the generated protocols often contain unordered steps, redundant operations, factual inconsistencies, or hallucinated actions, undermining both reproducibility and scientific credibility. Moreover, evaluation remains a central bottleneck: metrics such as BLEU \citep{papineni2002bleu}, ROUGE \citep{lin2004rouge}, and BERTScore \citep{zhang2019bertscore} capture only superficial lexical overlap, failing to reflect whether the generated action sequence is logically consistent, semantically faithful, and practically executable. While ``LLM-as-a-judge'' frameworks align better with human preferences, they introduce prohibitive costs to reinforcement learning (RL) pipelines, limiting scalability \citep{liu2023g, zheng2023judging}. Existing reward designs further neglect the structured and verifiable nature of protocols, producing outputs that are linguistically fluent but experimentally unreliable \citep{zeng2025reviewrl}. These limitations highlight the need for an approach that integrates structured and efficient evaluation.

In this work, we introduce a comprehensive framework that advances both data and modeling for protocol generation. At its core lies the “Sketch-and-Fill” paradigm, which formulates protocol generation as a structured reasoning process: each step is decomposed into essential components and expressed in natural language with explicit correspondence, ensuring logical coherence and experimental verifiability. To support this paradigm, we curate SciRecipe, a large-scale dataset of over 12K protocols spanning diverse domains and covering both Protocol-Comprehension and Problem-Solving tasks.
Building on this foundation, we propose the \textbf{S}tructured \textbf{CO}mponent-based \textbf{RE}ward (SCORE) mechanism, the central innovation of our framework. SCORE provides a structured reward and evaluation scheme that captures three complementary dimensions: step granularity (controlling scale and avoiding redundancy), action ordering (ensuring logically consistent sequences for reproducibility), and semantic fidelity (verifying alignment between predicted and reference actions, objects, and parameters). By jointly modeling these dimensions, SCORE moves beyond conventional text-based metrics to directly assess whether protocols are executable, interpretable, and scientifically sound. It serves both as an effective RL training signal and as a reliable metric for evaluation.
Based on these components, we develop Thoth, a protocol-generation model trained to combine structured reasoning with SCORE-guided evaluation. Extensive experiments demonstrate that Thoth outperforms SOTA models, particularly in step alignment, logical sequencing, and semantic accuracy. Just as importantly, the protocols it generates are concise and reproducible, which are qualities often absent from existing systems.
Our main contributions are as follows:

\begin{itemize}
\item We curate \textbf{SciRecipe}, a large-scale, multi-task dataset covering over 27 biological subfields, designed to serve as a foundation for both training and evaluating on protocol generation.
\item We introduce the \textbf{Sketch-and-Fill paradigm}, a reasoning framework that aligns with the logic of experimental design by converting open-ended queries into verifiable protocols.
\item We propose the \textbf{SCORE mechanism}, a structured reward and evaluation framework that jointly measures step granularity, order consistency, and semantic fidelity, ensuring that protocols are not only linguistically fluent but also experimentally executable.
\item We develop \textbf{Thoth}, a protocol-generation model with strong reasoning abilities, which achieves SOTA performance across protocol-specific and broader scientific benchmarks.  
\end{itemize}
\section{Related Works}
\textbf{LLMs in the Life Science}
Recent progress in life science LLMs has shifted from general-purpose systems to domain-specific models. BioBERT and SciBERT \citep{lee2020biobert, beltagy2019scibert}, pretrained on large-scale biomedical corpora, substantially improved tasks such as information extraction and question answering (QA). Autoregressive models like BioGPT \citep{luo2022biogpt} further advanced generative abilities, achieving strong and consistent results on PubMedQA \citep{jin2019pubmedqa}. These works collectively demonstrate the effectiveness of domain-specific pretraining and task adaptation, yet remain limited to knowledge-based tasks (e.g., QA, summarization) and cannot produce executable protocols. In contrast, proprietary models \citep{gpt5chat, comanici2025gemini, jaech2024openai} such as GPT-5 leverage massive corpora, parameter scales, and reasoning mechanisms to generate preliminary multi-step protocols. More recent efforts \citep{huang2025biomni, jin2025stella, zhang2025origene, gao2025txagent}, including Biomni and STELLA, integrate external knowledge and tools to support task-oriented solutions. Nonetheless, current models still struggle with redundancy, misordered steps, and hallucinations, thereby limiting their practical usability. To address this gap, we propose a biologically oriented model trained on real protocols, designed to generate rigorous experimental procedures through a structured scientific reasoning paradigm.
\begin{figure}[th]
    \centering 
    \includegraphics[width=\textwidth]{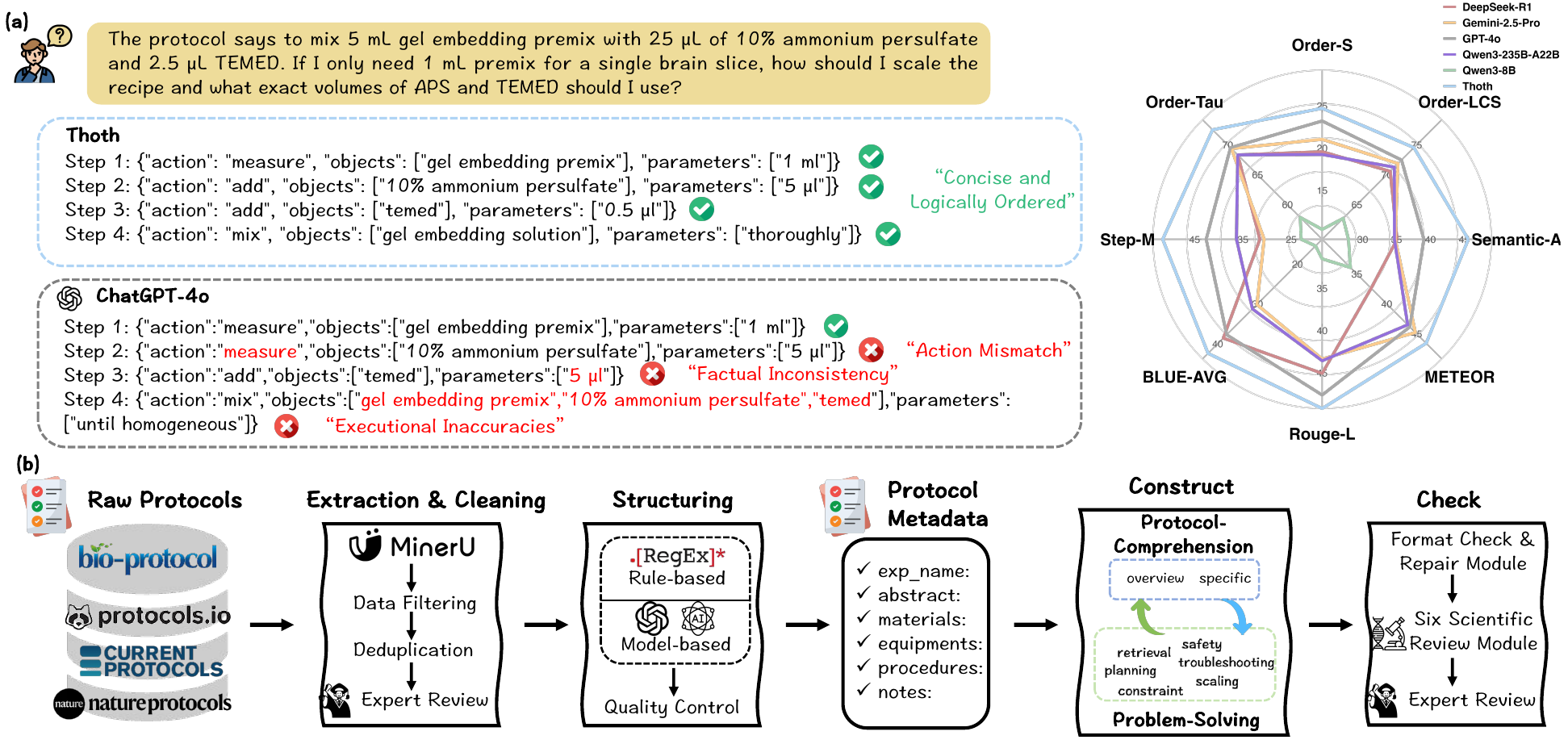} 
    \caption{(a) Thoth’s performance advantages over other models, shown through qualitative evaluation (left) and quantitative evaluation (right). (b) The construction pipeline of the SciRecipe dataset.} 
    \label{fig1} 
    \vspace{-5pt}
\end{figure}

\textbf{Reward Design and Evaluation for Open-Ended Generation}  
Designing effective evaluation remains a key challenge in open-ended generation. Unlike short-text tasks with clear ground truth, long-text generation such as experimental protocols lacks definitive answers and must therefore be assessed along multiple dimensions, including coherence, factuality, and executability \citep{becker2024text, que2024hellobench}. Metrics such as ROUGE, BLEU, and BERTScore \citep{lin2004rouge, papineni2002bleu, zhang2019bertscore} capture surface similarity but fail to fully reflect reasoning depth and often diverge from human judgment. The ``LLM-as-a-judge'' paradigm provides closer alignment with human preferences \citep{gu2024survey, zheng2023judging}, yet it usually introduces prohibitive computational costs in RL training \citep{shao2024deepseekmath, schulman2017proximal, ahmadian2024back, hu2025reinforce++}. Alternative approaches, including smaller evaluator models or reward mechanisms such as PrefBERT, Direct Reasoning Optimization, and LoVeC \citep{li2025semantically, xu2025direct, zhang2025reinforcement}, can improve efficiency or reliability but still require repeated model calls. To address these issues, we propose a rule-based evaluation that extracts key protocol elements through structured reasoning, thereby offering direct and verifiable reward signals with efficiency and interpretability.

\section{Methods}
This section outlines our approach to leveraging high-quality experimental protocol data and a component-wise reward mechanism to enhance the model’s ability to generate experimental solutions. We first propose the ``Sketch-and-Fill'' reasoning paradigm, which aligns with experimental execution logic to guide the model in organizing experimental steps effectively. Building on this, we construct SciRecipe, a dataset encompassing diverse experimental tasks and scenarios. Subsequently, we develop the SCORE mechanism to evaluate protocol quality across multiple dimensions, serving as a reward signal to improve the model’s generalization in protocol generation tasks. 
Ultimately, by leveraging SciRecipe and a multi-stage Knowledge-to-Action learning strategy, we train the Thoth model, which exhibits strong reasoning capabilities for protocol generation.

\subsection{SciRecipe Dataset}

In life science research, protocols serve as essential documents to ensure experimental reproducibility and reliability, providing detailed records of materials, equipment, procedures, and critical notes. Recently, platforms such as Nature Protocols, Bio-protocol, and Protocols.io \citep{enwiki:1296318228, bio-protocol, teytelman2016protocols} have compiled a vast number of standardized workflows, offering rich resources for the scientific community. From these sources, we collected over 23K protocols spanning 27 subfields including neuroscience, molecular biology, and cancer biology. After cleaning and structural processing, approximately 12K high-quality data were retained as the foundation of our dataset (the preprocessing pipeline is detailed in the Appendix \ref{Preprocessing of Original Scientific Protocols}). Building on this foundation, we introduce the \textbf{SciRecipe} dataset, designed to improve and evaluate LLMs in experimental protocol understanding and generation. SciRecipe comprises eight task types, grouped into two categories: Protocol-Comprehension Tasks (overview and specific), targeting global summarization and fine-grained analysis, and Problem-Solving Tasks (retrieval, planning, troubleshooting, constraint, scaling, and safety), simulating typical challenges encountered throughout experimental workflows. Together, these tasks form a complementary “understanding–application” loop.
The construction pipeline of SciRecipe is illustrated in Figure \ref{fig1} and further elaborated in the Appendix \ref{Pipeline_app}. Notably, benchmarks focused on scientific experimental protocol generation are currently scarce \citep{liu2025bioprobench}. To address this gap, we introduce \textbf{SciRecipe-Eval}, built with a similar pipeline, with dataset splits and difficulty levels detailed in the Appendix \ref{SciRecipe-Eval_app}.

\subsection{``Sketch-and-Fill'' Reasoning Paradigm}
\label{Reasoning Paradigm}
\begin{figure}[th]
    \centering 
    \includegraphics[width=\textwidth]{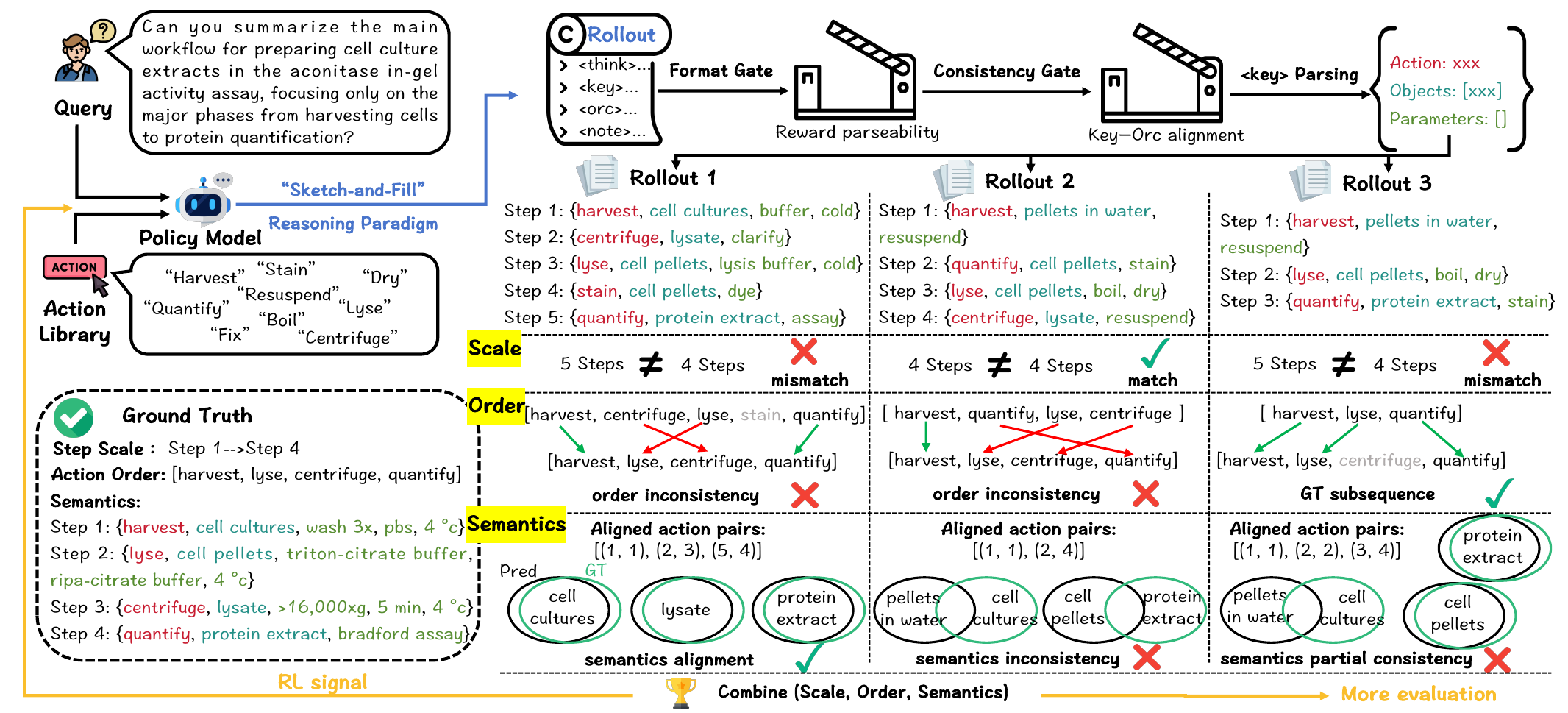} 
    \caption{Illustration of the SCORE mechanism and the “Sketch-and-Fill” reasoning paradigm. Three representative rollouts are subjected to a concise analysis across step scale, order consistency, and semantic consistency, while detailed computations are provided in Section~\ref{Reward Design}.} 
    \label{fig2} 
\end{figure}

To transform protocol generation tasks into an executable and evaluable form, we propose the ``Sketch-and-Fill'' reasoning paradigm. This paradigm is centered on a structured three-stage output: \texttt{<think>} $\to$ \texttt{<key>} $\to$ \texttt{<orc>}, with an additional \texttt{<note>} section dedicated to laboratory safety precautions. The core idea is to organize outputs in the sequence of reasoning, structuring, and expression, thereby enabling the design of a structured reward mechanism for RL. An overview of this paradigm is provided in Figure~\ref{fig2}.  
Specifically, \texttt{<think>}, \texttt{<key>}, and \texttt{<orc>} denote the reasoning process, the extraction of key information, and the final natural language output, respectively. In \texttt{<think>}, the model decomposes sub-goals, identifies sequential dependencies, and justifies the necessity of the proposed experimental steps, ensuring that the protocol is grounded in scientific reasoning. The parsed steps are then organized through a two-phase process of ``Sketch'' and ``Fill.''  
In the ``Sketch'' phase, represented by \texttt{<key>}, the strategies from \texttt{<think>} are transformed into a sequence of atomic, machine-readable steps. Each step is constrained to a single JSON dictionary representing one action unit: \texttt{\{"action": verb, "objects": [...], "parameters": [...]\}}. This abstraction reformulates natural language steps into predicate–object–adverbial triplets, bridging free-form instructions with structured sequences. The subsequent ``Fill'' phase, represented by \texttt{<orc>}, expands these steps into fluent natural language instructions, ensuring readability and executability. Formally, \texttt{<key>} is defined as $\mathbf{Y}$:
\vspace{-3pt}
\begin{equation}
\mathbf{Y} = \left(y_1, \ldots, y_m\right), \quad y_i = \left(a_i, \mathcal{O}_i, \mathcal{P}_i\right)
\vspace{-3pt}
\end{equation}
where $a_i$ is the experimental operation, $\mathcal{O}_i$ denotes the objects being acted upon, and $\mathcal{P}_i$ specifies the parameters (e.g., temperature, concentration). Consistency constraints, such as enforcing ``One-Action-Per-Step'' and maintaining uniform parameter application across objects, are applied to standardize the data format.    
In the ``Fill'' phase, represented by \texttt{<orc>}, the elements of \texttt{<key>} are rendered into human-readable natural language. A strict one-to-one correspondence in step count and semantics is enforced, ensuring no information is added or omitted, with the focus solely on readability. Overall, the ``Sketch-and-Fill'' paradigm standardizes scientific protocol generation by grounding open-ended language output in an executable structural space. This not only supports stable RL training but also provides a consistent foundation for automatic evaluation.

\subsection{SCORE Mechanism}
\subsubsection{Optimization Objective}
In scientific research, experimental protocols are not merely narrative texts but highly structured action guidelines refined through long-term practice. A qualified protocol must describe steps completely while conveying experimental logic: what to do (action), on what objects (objects), and under what conditions (parameters). Traditional text generation metrics (e.g., ROUGE, BLEU, BERTScore) often overlook such structure, rewarding lexical overlap even when action sequences are disordered. To address this, we propose the SCORE mechanism. Its optimization objective emphasizes both local alignment of protocol content and global logical consistency, defined as:
\begin{equation}
\max_{\theta}\;
\mathbb{E}_{(x,\mathcal{A},y^*)\sim\mathcal{D}}\,
\mathbb{E}_{y\sim\pi_{\theta}(\cdot\mid x,\mathcal{A})}\!\left[
\sum_{i=1}^{n}\;\sum_{j=1}^{m}
R_{\mathrm{local}}\!\big(y_{i},\,\, y^{*}_{j}\big)\otimes
R_{\mathrm{global}}\!\big(y,\,\, y^{*}\big)
\right]
\vspace{-5pt}
\end{equation}
which defines $\pi_{\theta}$ as the policy model with parameters $\theta$, $x$ as the context, $\mathcal{A}$ as the action library, and $y^*$ as the ground-truth protocol. $R_{\mathrm{local}}$ and $R_{\mathrm{global}}$ denote reward functions at different granularities, combined by $\otimes$ (details in Section \ref{Reward Design}). The core idea is to shift evaluation from superficial similarity to multidimensional alignment, including logical order and execution precision. Unlike metrics focusing only on text overlap or embeddings, SCORE directly measures the operability of generated protocols, providing scientific, interpretable, experiment-aligned optimization signals.

\subsubsection{Reward Design}
\label{Reward Design}
The SCORE mechanism adopts a progressive design (see Figure~\ref{fig2}). First, a gating stage ensures basic structural and consistency requirements. Then, fine-grained rewards for step scale and step semantics are combined into the final reward signal. This approach encourages the model to follow a rational reasoning paradigm and enhances both coherence and executability of the generated protocols.
\textbf{Format Gate}: The output must contain the four sections \texttt{<think>}, \texttt{<key>}, \texttt{<orc>}, and \texttt{<note>} in the correct order. Each step in \texttt{<key>} must follow the format \texttt{Step x:\{json\}}, explicitly specifying \texttt{action}, \texttt{objects}, and \texttt{parameters}, ensuring parseability for subsequent reward calculation.  
\textbf{Consistency Gate}: This gate verifies step-by-step correspondence between \texttt{<key>} and \texttt{<orc>}. Every action, object, and parameter in \texttt{<key>} must appear in \texttt{<orc>} with at least 95\% coverage (see Appendix \ref{Consistency Gate_app} for details), ensuring the protocol is a practical guide rather than a hollow framework. Only protocols passing both gates are used for reward computation.

\textbf{Step Scale}: The quality of a protocol depends on proper granularity, as too few steps cause omissions while too many lead to redundancy. We design a step scale reward that measures the gap between generated and gold step counts, with one atomic action per step guaranteed by the ``Sketch-and-Fill'' paradigm. Matching counts yield a score of 1, and deviations are penalized by cosine decay that drops to zero beyond a threshold. A length penalty is also applied, and when the average word length per step exceeds a limit, the score is proportionally reduced to discourage verbosity. Formally:
\vspace{-7pt}

{\small
\begin{equation}
f(d) = \left\{
\begin{array}{cc}
\cos \!\left(\tfrac{\pi d}{2 M}\right), & d < M \\
0, & d \geq M
\end{array}
\right\}, \quad
g(\bar{L}) = \left\{
\begin{array}{ll}
1, & \bar{L} \leq L \\
\bar{L}, & \bar{L} > L
\end{array}
\right\}, \quad
r_{\text{scale}} = 
\dfrac{f(|N_{\text{pred}} - N_{\text{gold}}|)}{g(\bar{L})}
\end{equation}
}
where $N_{\text{pred}}$ and $N_{\text{gold}}$ are predicted and gold step counts, and $M = \max\!\left(1, \lfloor 0.6N_{\text{gold}} \rfloor\right)$ is the deviation threshold. $\bar{L}$ denotes the average text length per step. $f(d)$ penalizes step mismatch via cosine decay, while $g(\bar{L})$ penalizes verbosity. Together, they yield the step scale reward $r_{\text{scale}}$.

\textbf{Step Semantics}  
The step semantics reward is central to SCORE, as it reflects the core execution logic of protocols. It consists of two components: Order Consistency and Semantic Consistency.  

\textbf{a) Order Consistency}, denoted as $Order(\cdot)$, assesses whether the generated sequence of experimental actions matches the ground truth. We adopt a ``Strict'' mode (see details in Appendix \ref{Order Consistency_app}), rewarding only if the predicted and ground-truth sequences are identical or mutually subsequences, and zero otherwise. This design mirrors laboratory reality since some steps may be repeated or omitted, but a disordered sequence renders the protocol invalid regardless of textual similarity.

\textbf{b) Semantic Consistency} is then calculated on the basis of action alignment, since actions function as the anchor of experimental steps (see Appendix \ref{Action Anchor_app}). In practice, researchers first decide what operation to perform (e.g., ``incubate'', ``add''), and only then specify the relevant objects and conditions. Without the action anchor, textual similarity cannot guarantee executability. Thus, we align predicted and gold actions step by step. For each aligned pair $(i,j)$, the overlap of object sets is measured using intersection-over-union, with subword-based similarity used as compensation when objects differ but are semantically related. Parameters are compared only if object overlap $\geq 0.5$, reflecting the principle that condition descriptions are meaningless for incorrect objects. Formally:
\begin{equation}
\mathrm{Obj}(i,j) = \frac{|\hat{\mathcal{O}}_i \cap \mathcal{O}_j^*|}{|\hat{\mathcal{O}}_i \cup \mathcal{O}_j^*|}, \quad
\operatorname{Par}(i,j) =
\begin{cases}
\quad\quad\,\,1, & \hat{\mathcal{P}}_i=\varnothing, \mathcal{P}_j^*=\varnothing \\
\quad\quad\,\,0, & \hat{\mathcal{P}}_i=\varnothing \vee \mathcal{P}_j^*=\varnothing \\
\frac{|\mathcal{K}(\hat{\mathcal{P}}_i)\cap \mathcal{K}(\mathcal{P}_j^*)|}{|\mathcal{K}(\hat{\mathcal{P}}_i)\cup \mathcal{K}(\mathcal{P}_j^*)|}, & \text{otherwise}
\end{cases}
\vspace{-5pt}
\end{equation}
where $\hat{\mathcal{O}}_i$ and $\hat{\mathcal{P}}_i$ are the predicted object and parameter sets at step $i$, $\mathcal{O}_j^*$ and $\mathcal{P}_j^*$ are the corresponding ground truth sets at step $j$, and $\mathcal{K}(\cdot)$ extracts subword sets for reliable comparison. To better incorporate positional fidelity, we apply a decay factor $m_{ij}=\max\{0,1-(|i-j|/D)^{\lambda}\}$, where $D$ is the number of ground truth steps. This reduces the score when predicted actions, though correct in type, appear far from their reference positions. The final score is:
\begin{equation}
r_{\mathrm{semantics}} = Order(\hat{a}, a^*) +
\frac{1}{|\mathcal{W}|} \sum_{(i,j)\in\mathcal{W}} m_{ij}\!\left(\mathrm{Obj}(i,j)+\tfrac{1}{2}\mathrm{Par}(i,j)\right)
\vspace{-5pt}
\end{equation} 
where $\mathcal{W}$ is the set of aligned action pairs. Unlike the step scale (equation \ref{eq6}), we adopt an additive combination to avoid over-penalization. This allows partial credit when objects or parameters are imperfectly matched, as long as the action sequence remains reasonable. Such tolerance improves training stability and aligns with how real researchers judge protocols, where minor detail errors may be acceptable if the experimental logic is preserved.
Finally, the overall SCORE is defined as:
\begin{equation}
\label{eq6}
\operatorname{SCORE}(y,y^*)=\mathbb{I}_{\text{format}}(y)\cdot\mathbb{I}_{\text{cons}}(y)\cdot r_{\text{scale}}(y,y^*) \cdot r_{\text{semantics}}(y,y^*)
\end{equation}
where $\mathbb{I}_{\text{format}}$ and $\mathbb{I}_{\text{cons}}$ are gating functions. This design ensures that only protocols satisfying structural, step scale, and semantic requirements obtain high rewards \citep{wang2025acting}, thereby mitigating reward hacking and more faithfully simulating how protocols are evaluated in practice. SCORE also serves as a multidimensional evaluation framework, with details in Appendix \ref{Evaluation Metric_add}.

\subsection{Knowledge-to-Action Learning Strategy}
Inspired by curriculum learning \citep{bengio2009curriculum, wang2021survey}, we propose a three-stage Knowledge-to-Action Learning framework that progressively enables the transition from textual knowledge to protocol generation. The framework parallels human learning, progressing from knowledge accumulation to standardized operations and finally to exploratory optimization. In the first stage, pre-training, the model learns the semantic structure and operational logic of experimental language from large-scale protocol texts (see Appendix \ref{pretrained} for further experiments). The second stage, supervised instruction tuning (SFT), is conducted on data following the ``Sketch-and-Fill'' paradigm, incorporating subtasks such as parameter filling, step ordering, and error correction \citep{liu2025bioprobench}. SFT both injects domain knowledge and provides a cold start for RL, aligning outputs with the designated paradigm. The third stage applies RL with the GRPO algorithm (details in Appendix \ref{RL Algorithm} \& \ref{RL Algorithms}) \citep{guo2025deepseek}. By removing entropy loss and reducing the KL penalty, we enhance exploration and avoid premature convergence. Combined with SCORE rewards, this stage improves generalizability and robustness, ensuring more reliable and executable protocol generation.

\section{Experiments and Results}
\subsection{Experiment Settings}
\textbf{Benchmarks \& Baselines}
To evaluate our model, we adopt two types of benchmarks. The first focuses on real-world protocol generation, represented by SciRecipe-Eval, which emphasizes linguistic quality and practical executability. The second covers broader scientific reasoning and question answering across domains, including Humanity's Last Exam (HLE), LAB-Bench, and PubMedQA \citep{phan2025humanity, laurent2024lab, jin2019pubmedqa}. (Additionally, BioProBench sub-tasks \citep{liu2025bioprobench} are used as supplementary experiments for protocol comprehension.) We further compare against proprietary, open-source, reasoning, and scientific LLMs (see Appendix \ref{Baseline Models_add} for details).

\textbf{Evaluation Metrics}  
We designed metrics tailored to different task types. For natural language generation, BLEU, ROUGE, METEOR, and keyword matching \citep{liu2025bioprobench} were used to assess surface-level semantic similarity \citep{papineni2002bleu, lin2004rouge, banerjee2005meteor, grootendorst2020keybert}. Based on the SCORE mechanism, we introduced five executability metrics, namely Step-MATCH (Step-M), Order-LSC/S/Tau, and Semantic-Alignment (Semantic-A), to evaluate step scale, order, and semantic fidelity (see Appendix \ref{Evaluation Metric_add} for details). For multiple-choice, classification, and ranking tasks, we employed Accuracy, F1, Exact Match, and Kendall’s Tau \citep{abdi2007kendall}.

\textbf{Implementation Details}  
We used GPT-5 Chat \citep{gpt5chat} (temperature 0.6) to construct SciRecipe and Gemini 2.5 Flash \citep{comanici2025gemini} (temperature 0.2) for validation. In the SCORE mechanism, hyperparameters $L=30$ and $\lambda=1.5$ were determined through iterative experiments. For training Thoth, we adopted Qwen3-8B as the base model \citep{yang2025qwen3}. Pre-training and SFT used LoRA fine-tuning via LLaMA-Factory \citep{zheng2024llamafactory}, while RL full-parameter tuning was conducted with the VeRL framework \citep{sheng2025hybridflow}. All experiments ran on eight Nvidia H100 GPUs. Further hyperparameters and data ratios are given in Appendix \ref{Implementation Details_add}.

\subsection{Main Results}
\renewcommand{\arraystretch}{1.1}
\begin{table}[h]
\centering
\caption{Main results on SciRecipe-Eval. Metrics left of the dashed line evaluate executability, those on the right measure lexical similarity. \textbf{Bold} denotes the best score (see Appendix \ref{Fine-grained Results on SciRecipe} for details).}
\label{TABLE1}
\resizebox{\textwidth}{!}{%
\begin{tabular}{@{}lccccc:ccccc@{}}
\toprule
\textbf{Methods} &
  \textbf{Semantic-A} &
  \textbf{Order-LCS} &
  \textbf{Order-S} &
  \textbf{Order-Tau} &
  \textbf{Step-M} &
  \textbf{BLUE-AVG} &
  \textbf{ROUGE-L} &
  \textbf{METEOR} &
  \textbf{KW-F1} &
  \textbf{AVG} \\ \midrule
\multicolumn{11}{l}{\textit{Close-Source SOTA}}                                                                            \\
\textbf{ChatGPT-4o}                   & 40.04 & 73.27 & 24.00 & 70.33 & 44.00 & 38.95 & 48.42 & 44.66 & 52.05 & 48.41 \\
\textbf{GPT-5}                    & 27.79 & 58.12 & 11.35 & 53.55 & 18.79 & 21.31 & 32.96 & 32.55 & 39.17 & 32.84 \\
\textbf{GPT-5 Chat}               & 36.30 & 73.21 & 21.17 & 65.67 & 25.00 & 29.57 & 42.04 & 41.95 & 47.87 & 42.53 \\
\textbf{Claude Sonnet 4}          & 39.35 & 71.97 & 20.83 & 70.00 & 35.83 & 34.24 & 44.27 & 40.97 & 49.40 & 45.21 \\
\textbf{Claude Opus 4.1}          & 41.32 & 71.70 & 21.80 & 71.93 & 34.59 & 34.69 & 44.42 & 40.36 & 50.00 & 45.65 \\
\textbf{Gemini 2.5 Flash}         & 36.35 & 70.61 & 20.00 & 70.33 & 32.33 & 33.19 & 42.91 & 39.26 & 48.07 & 43.67 \\
\textbf{Gemini 2.5 Pro}           & 35.80 & 72.68 & 21.83 & 70.17 & 32.00 & 31.37 & 44.16 & 45.59 & 48.58 & 44.69 \\
\textbf{Doubao-1.5-pro}           & 33.33 & 73.29 & 23.67 & 70.00 & 47.50 & 38.16 & 46.88 & 38.71 & 48.74 & 46.70 \\
\textbf{Qwen2.5-Max}              & 40.34 & 72.88 & 21.83 & 71.33 & 47.50 & 30.81 & 48.02 & 43.82 & 51.98 & 47.61 \\ \midrule
\multicolumn{11}{l}{\textit{Open-Source SOTA}}                                                                             \\
\textbf{Qwen2.5-72B-Instruct}     & 36.40 & 70.82 & 21.00 & 69.17 & 42.17 & 29.55 & 46.06 & 43.44 & 49.73 & 45.37 \\
\textbf{Qwen3-235B-A22B-Instruct} & 35.68 & 72.07 & 20.03 & 69.12 & 37.73 & 32.48 & 44.37 & 44.30 & 47.89 & 44.85 \\
\textbf{DeepSeek-V3}              & 41.72 & 73.97 & 21.44 & 70.54 & 41.71 & 38.18 & 48.49 & 45.08 & 52.33 & 48.16 \\
\textbf{GPT-OSS-120B}             & 32.86 & 69.97 & 17.67 & 64.17 & 27.83 & 30.72 & 43.44 & 42.90 & 49.92 & 42.16 \\
\textbf{Llama-3.1-405B-Instruct}  & 35.92 & 69.46 & 18.03 & 67.78 & 39.23 & 36.42 & 44.86 & 42.17 & 48.09 & 44.66 \\
\textbf{Kimi-K2-Instruction}      & 36.99 & 71.83 & 20.83 & 69.83 & 40.00 & 33.81 & 44.49 & 42.99 & 49.00 & 45.53 \\ \midrule
\multicolumn{11}{l}{\textit{Reasoning Models}}                                                                             \\
\textbf{DeepSeek-R1}              & 36.07 & 71.38 & 20.37 & 69.12 & 32.89 & 39.80 & 45.86 & 38.19 & 49.83 & 44.83 \\
\textbf{Grok 3}                   & 37.40 & 73.27 & 21.92 & 69.73 & 39.25 & 34.72 & 46.04 & 46.21 & 48.59 & 46.35 \\
\textbf{Grok 4}                   & 36.73 & 72.08 & 20.25 & 65.34 & 34.66 & 37.21 & 46.17 & 40.18 & 51.81 & 44.94 \\
\textbf{OpenAI-o1}                & 34.74 & 73.40 & 18.53 & 67.45 & 35.39 & 35.68 & 46.29 & 43.82 & 50.13 & 45.05 \\
\textbf{OpenAI-o3}                & 35.40 & 70.38 & 15.38 & 65.05 & 24.08 & 28.62 & 43.08 & 44.33 & 50.26 & 41.84 \\ \midrule
\textbf{Qwen3-4B}                 & 24.37 & 53.55 & 13.67 & 50.50 & 28.83 & 14.52 & 24.74 & 23.95 & 27.69 & 29.09 \\
\rowcolor[HTML]{FFFFC7} 
\textbf{Thoth-mini}               & 44.28 & 74.68 & 25.33 & 70.83 & 52.67 & 43.32 & 49.23 & 46.41 & 53.13 & 51.10 \\
\textbf{Qwen3-8B}                 & 28.89 & 63.51 & 11.17 & 58.67 & 24.33 & 16.66 & 32.31 & 34.72 & 38.63 & 34.32 \\
\rowcolor[HTML]{FFFFC7} 
\textbf{Thoth}                    & \textbf{46.60} & \textbf{75.34} & \textbf{25.50} & \textbf{73.33} & \textbf{53.00} & \textbf{43.62} & \textbf{50.02} & \textbf{47.39} & \textbf{54.13} & \textbf{52.10} \\ \bottomrule
\end{tabular}}
\end{table}
\vspace{-5pt}
\textbf{Protocol Generation}  
~We evaluated protocol generation across two dimensions: executability, which emphasizes consistency in objects and step order, and lexical similarity, which reflects surface-level matching. As shown in Table \ref{TABLE1}, Thoth achieves SOTA results across all metrics. Compared to baselines, Thoth and Thoth-mini improve average performance by 17.78\% and 22.01\%, underscoring the effectiveness of our approach. Thoth also surpasses much larger proprietary models, outperforming ChatGPT-4o by 3.69\% on average. Against the strongest open-source model, DeepSeek-V3, Thoth achieves gains of 4.88\%, 4.06\%, and 11.29\% on Semantic-Alignment, Order-S, and Step-MATCH, respectively, showing clear advantages in step alignment, logical order, and action fidelity. General-purpose SOTA models show notable capability, likely due to large-scale pre-training, but mainly achieve superficial similarity and lack real-world executability. Reasoning-oriented models also perform poorly, often producing overly complex outputs unsuited for laboratory workflows (see Appendix \ref{Qualitative Study of Reasoning Models} for qualitative cases). Overall, Thoth demonstrates consistent advantages across metrics, bridging knowledge-based text generation with executable protocols.

\begin{wrapfigure}[16]{r}{0.5\textwidth} 
  \centering
  \includegraphics[width=\linewidth]{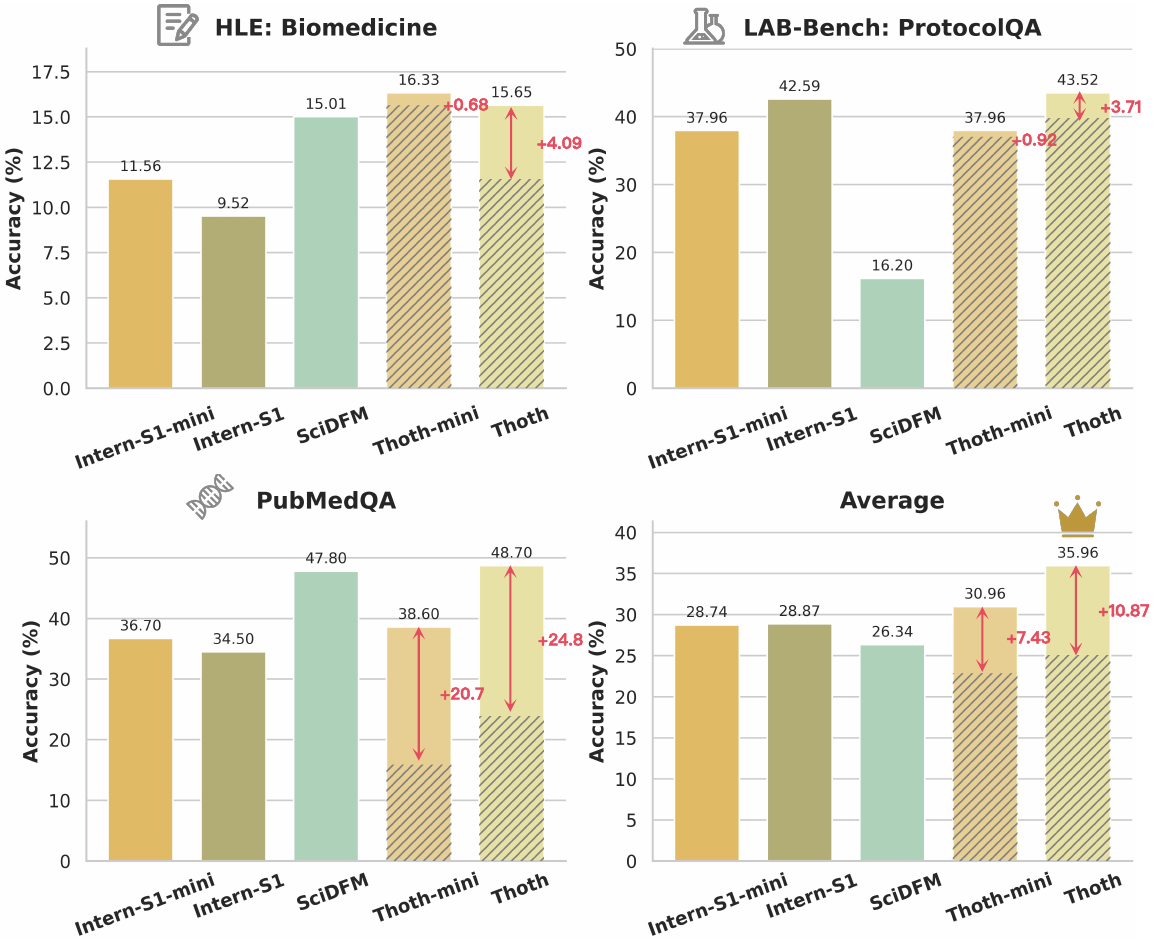} 
  \caption{Results on other scientific benchmarks. Slash shading denotes baseline of Thoth models.}
  \label{fig3}
\end{wrapfigure}

\textbf{Fundamental Scientific Tasks}
~Beyond protocol generation, we evaluated Thoth on three out-of-domain biomedical benchmarks against recent scientific LLMs (Intern-S1 and SciDFM), with results shown in Figure \ref{fig3}. Thoth consistently outperforms models fine-tuned on biomedical corpora, exceeding Intern-S1 and Intern-S1-mini by 7.09\% and 2.22\% on average, and notably both Thoth and Intern-S1-mini share Qwen3-8B as their base model. As illustrated in Figure \ref{fig3}, Thoth also shows clear improvements over its baseline models (slash shading), achieving an average improvement of 10.87\%, while Thoth-mini improves by 7.43\%. The performance gap between Thoth and Thoth-mini further confirms the benefits of our approach. These results demonstrate that knowledge and reasoning skills acquired from scientific protocols generalize effectively to broader biomedical tasks. Additional experiments on protocol comprehension are provided in Appendix \ref{Other Protocol Benchmark_add}.
\subsection{Analyses}
\textbf{SciRecipe}
~To assess the impact of different data components, we conducted ablation studies on SciRecipe (Table~\ref{TABLE3}). Models trained only on QA or generation QA showed weak results, with BLEU-AVG below 40\%. Adding Protocol-Comprehension tasks markedly improved executability, with Order-LCS reaching 74.50\%. Joint training on both task types further enhanced alignment, with Step-M rising to 52.33\%. The best overall performance was obtained when combining QA with scientific review, achieving 46.60\% on Semantic-A and 49.45\% on AVG. These findings highlight that diverse task coverage and systematic quality checks enhance the reliability of protocol generation.
\vspace{-5pt}
\renewcommand{\arraystretch}{1.1}
\begin{table}[th]
\centering
\caption{Ablation studies on training set. $\clubsuit$: trained only on SciRecipe Protocol-Comprehension tasks; $\spadesuit$: trained on both task types. QA and SciCheck denote inclusion of basic protocol QA and scientific review of SciRecipe, respectively. $\star$: trained only on protocol generation QA.}
\label{TABLE3}
\resizebox{\textwidth}{!}{%
\begin{tabular}{@{}ccc|ccccccccc@{}}
\toprule
\textbf{SciRecipe} &
\textbf{QA} &
\textbf{SciCheck} &
\textbf{Semantic-A} &
\textbf{Order-LCS} &
\textbf{Order-S} &
\textbf{Step-M} &
\textbf{BLUE-AVG} &
\textbf{ROUGE-L} &
\textbf{METEOR} &
\textbf{KW-F1} &
\textbf{AVG} \\ \midrule
 & $\checkmark$ & \textemdash & \textemdash & \textemdash & \textemdash & \textemdash & 23.88 & 31.50 & 29.53 & 34.50 & 29.85 \\
 & $\star$ & \textemdash & \textemdash & \textemdash & \textemdash & \textemdash & 23.71 & 32.08 & 27.92 & 37.78 & 30.37 \\
$\clubsuit$ &  & $\checkmark$ & 42.34 & 74.50 & 24.33 & 50.50 & 41.33 & 47.34 & 44.81 & 51.72 & 47.11 \\
$\clubsuit$ & $\checkmark$ & $\checkmark$ & 44.19 & 74.85 & 23.83 & 43.67 & 42.42 & 47.59 & 45.44 & 52.29 & 46.79 \\
$\spadesuit$ &  & $\checkmark$ & 44.51 & 74.67 & 25.00 & 52.33 & 42.74 & 48.53 & 46.84 & 52.30 & 48.37 \\
$\spadesuit$              & $\checkmark$ & \multicolumn{1}{c|}{}  & 44.54 & 75.14 & 23.50 & 47.00 & 41.79 & 48.09 & 47.01 & 51.89 & 47.37 \\
$\spadesuit$              & $\checkmark$ & \multicolumn{1}{c|}{$\checkmark$} & 46.60 & 75.34 & 25.50 & 53.00 & 43.62 & 50.02 & 47.36 & 54.13 & 49.45 \\
\bottomrule
\end{tabular}%
}
\end{table}
\vspace{-7pt}

\begin{wrapfigure}[20]{r}{0.5\textwidth} 
\vspace{-20pt}
  \centering
  \includegraphics[width=\linewidth]{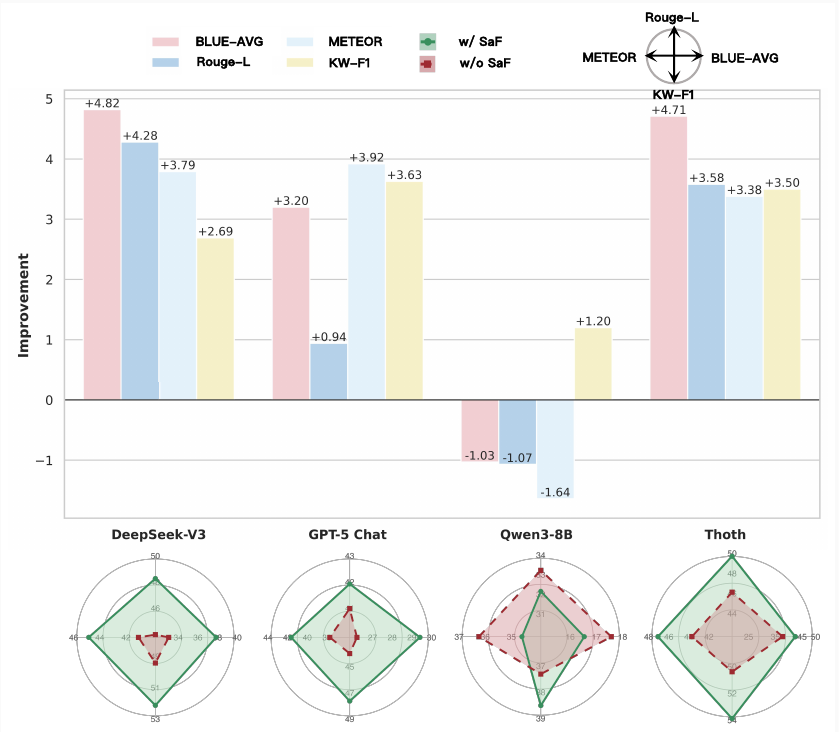} %
  \caption{Ablation studies on reasoning paradigm. Performance differences with/without Sketch-and-Fill (SaF) are shown in the top panel, and a visual comparison is provided in the bottom panel.}
  \label{fig4}
\end{wrapfigure}

\textbf{Reasoning Paradigm}  
~To align model outputs with experimental logic and generate evaluable protocols, we propose the ``Sketch-and-Fill'' reasoning paradigm. Its effect was systematically tested through ablation experiments on four representative models: DeepSeek-V3, GPT-5 Chat, Thoth, and its base model Qwen3-8B. Since outputs are in natural language, executability metrics were not applicable, and Figure \ref{fig4} reports results based on lexical similarity. The results show that ``Sketch-and-Fill'' improves performance for all models except the base model, with notable gains on BLEU-AVG. DeepSeek-V3, GPT-5 Chat, and Thoth achieve average improvements of 3.89\%, 2.92\%, and 3.79\% over counterparts without the paradigm. For the base model, the lack of improvement is attributed to reasoning failure due to missing knowledge injection from scientific protocols.

\renewcommand{\arraystretch}{1.1}
\begin{table}[ht]
\centering
\caption{Ablation studies on the SCORE mechanism. Evaluation covers step scale, semantic alignment, and overall reward design. $KL(\cdot)$ denotes the KL-divergence penalty in the loss function, and “Vanilla” refers to using standard semantic similarity metrics as the reward signal.}
\label{TABLE4}
\resizebox{\textwidth}{!}{%
\begin{tabular}{@{}c c| cccccccccc@{}}
\toprule
\multicolumn{2}{c|}{\textbf{SCORE}} &
  \textbf{Semantic-A} & \textbf{Order-LCS} & \textbf{Order-S} &
  \textbf{Order-Tau} & \textbf{Step-M} & \textbf{BLUE-AVG} &
  \textbf{ROUGE-L} & \textbf{METEOR} & \textbf{KW-F1} & \textbf{AVG} \\ \midrule

\multirow{3}{*}{\centering\arraybackslash \textbf{Step Scale}} &
\textbf{w/o $f(d)$}      & 43.67 & 55.97 & 6.83  & 51.83 & 10.00 & 43.09 & 37.28 & 26.08 & 43.31 & 35.34 \\
& \textbf{w/o $g(\bar{L})$}      & 38.89 & 74.37 & 23.67 & 71.00 & 51.00 & 43.18 & 46.14 & 44.92 & 52.80 & 49.55 \\
& \textbf{w/o $f(d)$+$g(\bar{L})$} & 45.35 & 49.50 & 3.00  & 34.17 & 4.17  & 42.49 & 33.44 & 21.12 & 40.41 & 30.41 \\ \cmidrule(r){1-2}

\multirow{3}{*}{\centering\arraybackslash \textbf{\begin{tabular}[c]{@{}l@{}}\\Semantic\\Alignment\end{tabular}}} &
\textbf{w/o $m_{ij}$}      & 38.68 & 73.70 & 23.17 & 67.67 & 50.17 & 42.84 & 48.91 & 45.58 & 52.85 & 49.29 \\
& \textbf{w/o $Order(\cdot)$}  & 40.93 & 61.27 & 12.83 & 48.00 & 33.33 & 37.26 & 45.35 & 43.27 & 49.46 & 41.30 \\
& \textbf{\begin{tabular}[c]{@{}l@{}}w/o $Order(\cdot)$\\+$f(d)$+$g(\bar{L})$\end{tabular}}
                        & 44.15 & 58.22 & 7.67  & 42.83 & 43.33 & 38.45 & 45.78 & 43.52 & 48.74 & 41.41 \\ \cmidrule(r){1-2}

\multirow{2}{*}{\centering\arraybackslash \textbf{Reward}} &
\textbf{w/o $KL(\cdot)$}     & 39.96 & 74.43 & 22.67 & 69.33 & 46.67 & 42.73 & 46.03 & 44.82 & 53.36 & 48.89 \\
& \textbf{Vanilla}      & 38.74 & 63.41 & 21.50 & 52.36 & 44.50 & 45.52 & 50.12 & 43.90 & 50.54 & 45.62 \\ \cmidrule(r){1-2}

\multicolumn{2}{c|}{\textbf{Thoth}} &
46.60 & 75.34 & 25.50 & 73.33 & 53.00 & 43.62 & 50.02 & 47.39 & 54.13 & 52.10 \\ \bottomrule
\end{tabular}}
\vspace{-5pt}
\end{table}

\textbf{SCORE Mechanism}  
In Section \ref{Reward Design}, we introduced the SCORE reward mechanism, consisting of step scale, order consistency, and semantic consistency. To evaluate its role in RL training, we conducted ablation studies, including variants that removed individual components, dropped the KL divergence penalty, or replaced SCORE with a vanilla reward based only on BLEU-AVG, ROUGE-L, and BERTScore. Results are summarized in Table \ref{TABLE4}.  
Omitting the step scale reward leads to sharp declines in Order-S and Step-M (6.83\%/10\% and 3\%/4.17\%, rows 1 and 3), indicating failures to generate executable protocols and producing either verbose or incomplete outputs. Excluding the positional decay factor (row 4) weakens penalties for misaligned steps and reduces semantic consistency, under which Thoth still achieves a 7.92\% gain in Semantic-A. The order consistency reward is also critical because without enforcing execution order, steps become misarranged and semantic coherence breaks down. Thoth surpasses the variants in rows 5 and 6 by 6.36\% and 5.17\% on BLEU-AVG, respectively.
Removing the KL penalty further degrades performance, with Table \ref{TABLE4} showing a 3.21\% average drop compared to Thoth. Although the vanilla reward improves BLEU-AVG and ROUGE-L, the generated protocols exhibit poor executability with a 10.65\% average reduction. These results demonstrate the effectiveness of SCORE in guiding the model to generate coherent and executable protocols. For further analyses of reward, see Appendix~\ref{Reward Computations} \& ~\ref{Reward Magnitudes}.

\begin{figure}[ht]
    \centering 
    \includegraphics[width=\textwidth]{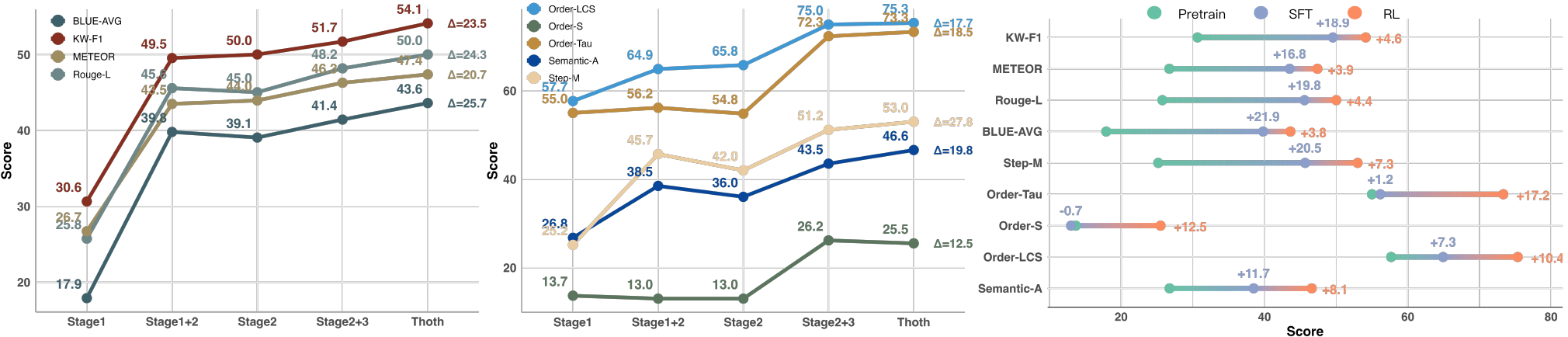} 
    \caption{Ablation studies on the training strategy. Left and middle: rightmost $\Delta$ marks extreme values across strategies. Right: Thoth’s training trajectory across stages.} 
    \label{fig5} 
\end{figure}
\textbf{Training Strategy}  
~To progressively equip the model for solving complex scientific problems, we propose the three-stage Knowledge-to-Action Learning strategy. We compared models trained at different stages with Thoth to validate its effectiveness. As shown in Figure \ref{fig5} (left and middle), Thoth achieves SOTA performance after staged training and consistently outperforms models trained with only Stage 1+2 (pre-training and SFT) or without pre-training (Stage 2+3 or Stage 2 alone). Specifically, Thoth improves by an average of 11.08\% on executability and 4.2\% on lexical similarity over Stage 1+2. It further achieves mean gains of 1.44\% and 8.79\% over Stage 2+3 and Stage 2 across both dimensions. Figure \ref{fig5} (right) further demonstrates that the three-stage process progressively enhances performance, with SFT improving lexical similarity and RL strengthening rationality and executability.  

\section{Conclusion}
In this paper, we introduced Thoth, a protocol-generation model grounded in the SciRecipe dataset, the ``Sketch-and-Fill'' reasoning paradigm, and the SCORE mechanism. By integrating structured reasoning with verifiable rewards, Thoth achieves SOTA performance on both protocol-specific and broader scientific benchmarks. Equally important, the protocols it generates are concise, logically coherent, and experimentally executable. This work provides a blueprint for building reliable scientific assistants and demonstrates that structured reasoning frameworks combined with targeted rewards are critical for advancing protocol generation and improving reproducibility.


\section*{Reproducibility Statement}
We have made extensive efforts to ensure the reproducibility of our results. 
Details on training data distributions, hyperparameter settings for pre-training, SFT, and RL, as well as implementation frameworks, are provided in Appendix~\ref{Implementation Details_add}. 
We will also release all code, models, and processed datasets to facilitate faithful reproduction of our findings.

\bibliography{iclr2026_conference}
\bibliographystyle{iclr2026_conference}

\clearpage

\appendix
\startcontents[appendix]
\section*{Appendix Contents}
\printcontents[appendix]{l}{1}{\setcounter{tocdepth}{2}}
\clearpage

\section{SciRecipe}
\label{SciRecipe_app}

\subsection{Preprocessing of Original Scientific Protocols}
\label{Preprocessing of Original Scientific Protocols}
In the preprocessing phase, we first employed MinerU \citep{wang2024mineru} to identify and extract protocol texts, as the majority of online experimental protocols are distributed in non-editable PDF format. This step was essential because protocols published in PDF form often contain heterogeneous content such as figures, tables, references, and lengthy narratives, which substantially increase the difficulty of accurate parsing and subsequent question–answer pair construction. To mitigate this issue, we analyzed both the page count and the number of figures/tables in each document, and applied empirically determined thresholds to filter out overly long or structurally complex protocols.
Another important consideration was the removal of redundancy, since many protocols share highly similar structures or represent slightly modified versions of the same experiment. Without careful filtering, these near-duplicate entries could cause information leakage across training and evaluation splits. To address this, we applied similarity detection and clustering. Specifically, we used Qwen3-Embedding-8B \citep{zhang2025qwen3} to vectorize protocol titles, built an efficient similarity search index using FAISS \citep{douze2024faiss}, and conducted Top-K nearest-neighbor searches. Similar pairs were identified by thresholding cosine similarity scores, after which a union–find algorithm was applied to partition protocols into connected components. This process produced both pairwise and grouped similarity sets, effectively eliminating redundant samples while preserving data diversity and representativeness.  
In addition, some protocols were accompanied by electronic supplementary materials, which frequently contained fragmented or ambiguous instructions. Because these supplementary files risked introducing inconsistencies and annotation difficulties, they were uniformly removed from the dataset. To further guarantee the reliability of the corpus, we recruited three doctoral students with extensive experimental experience to manually review a subset of the data. Their verification ensured not only correctness but also balanced coverage across experimental domains, ultimately enhancing both the diversity and representativeness of the retained corpus.

\subsection{Structured Integration of Scientific Protocols}
\label{Structured Integration of Scientific Protocols}
Because experimental protocols vary widely in writing style and formatting, we extracted common elements to form a unified structured representation. The core fields included experiment name, research background, reagents and materials, experimental equipment, and experimental procedures.  
This integration was carried out in two complementary steps. First, we applied a rule-based method that relied on regular expressions to match predefined keywords and extract the corresponding fields. During this stage, Markdown tables were removed to avoid parsing errors, while figure and table captions were preserved as valuable contextual information. The rule-based approach achieved reliable performance for protocols written in a highly standardized format, but its applicability was limited for documents with more complex or irregular structures.  
To address these limitations, we further adopted a model-based strategy. Specifically, we leveraged Grok-4 \citep{xai2025grok4} to summarize and restructure the original protocols, ensuring accurate extraction of key fields without introducing fabricated content. The model also adjusted the ordering of experimental steps to maintain logical and procedural consistency. This hybrid two-step pipeline allowed us to combine the precision of rule-based parsing with the flexibility of model-based restructuring.  
Through this process, we obtained over 12K structured protocols. Quality control measures were subsequently applied to filter out entries with missing fields, excessively lengthy procedures, or dependencies on external online resources. The resulting corpus provides a clean, logically consistent, and representative collection of protocols suitable for downstream training and evaluation.

\subsection{Supplementary Details on SciRecipe}
When designing the eight task types in SciRecipe, we defined explicit objectives and representative scenarios for each. Together, these tasks cover the entire lifecycle of experimental research, ensuring that the dataset trains and evaluates both general reasoning and practical execution.  

\begin{itemize}
    \item \textbf{Overview}: Requires the model to generate a hierarchically organized summary of the protocol, enabling rapid comprehension of its global structure.  
    \item \textbf{Specific}: Focuses on step-by-step decomposition of procedures, ensuring the model captures the micro-level operational logic.  
    \item \textbf{Retrieval}: Targets precise extraction of experimental parameters (e.g., temperature, pH, concentration), emphasizing accuracy in reporting.  
    \item \textbf{Planning}: Involves transforming high-level objectives into coherent, logically ordered experimental steps.  
    \item \textbf{Troubleshooting}: Simulates diagnosing experimental failures, identifying error sources, and proposing corrective strategies.  
    \item \textbf{Constraint}: Tests adaptability under resource limitations, such as restricted equipment or reagent availability.  
    \item \textbf{Scaling}: Requires numerical adjustments and unit conversions to adapt experiments across different scales.  
    \item \textbf{Safety}: Focuses on compliance with safety standards and identification of potential laboratory risks.  
\end{itemize}

By encompassing comprehension, planning, error handling, adaptability, scalability, and safety, SciRecipe ensures a comprehensive and targeted task design. See Figure \ref{fig21} for detailed prompts.

\subsection{Supplementary Details on SciRecipe Construction Pipeline}
\label{Pipeline_app}
During the construction of SciRecipe, we first generated QA pairs for multiple task types using the structured protocol corpus as the foundation. Particularly, for Protocol-Comprehension tasks, we relied on hierarchical tree extraction (e.g., ``Step 1: xxx; Step 1.1: xxx; Step 1.1.1: xxx''), which preserved both the hierarchical structure and the stepwise dependencies of experimental procedures. This design ensured that the model could capture not only surface-level descriptions but also the nested logic inherent in complex experiments.  
Then, we introduced the ``Sketch-and-Fill'' reasoning paradigm. This approach first summarizes core elements in an outline-like format and then fills in the specific operations, thereby enhancing both interpretability and training effectiveness. The paradigm was particularly effective in decomposing high-level objectives into concrete experimental actions while maintaining logical consistency. 
Because LLMs may produce outputs with randomness and formatting inconsistencies, directly discarding such samples would risk losing high-quality data. To address this, we designed a format validation and repair module. This module automatically detected non-compliant outputs and prompted the model to regenerate them under low-temperature settings, ensuring structural consistency while retaining valid content.  

After passing the format checks, each QA pair underwent content review across six scientific dimensions:
\begin{enumerate}
    \item \textbf{Scientific Accuracy}: correctness of operations, parameters, and outcomes.  
    \item \textbf{Safety \& Compliance}: adherence to laboratory safety standards and ethical guidelines.  
    \item \textbf{Logical Coherence \& Actionability}: stepwise consistency and rational flow of procedures.  
    \item \textbf{Clarity \& Ambiguity}: precision and unambiguity of textual descriptions.  
    \item \textbf{Generality \& Specificity}: balanced coverage of broad applicability and detailed context.  
    \item \textbf{Efficiency \& Resource Optimization}: optimization of time, materials, and experimental resources.  
\end{enumerate}

For this step, we employed Gemini 2.5 Flash \citep{comanici2025gemini} as the validation model, which performed dimension-wise evaluations and produced review reports. QA pairs that failed to meet requirements were systematically discarded. 
Finally, to guarantee the reliability of the constructed dataset, we performed manual verification. Doctoral students from interdisciplinary backgrounds conducted thorough inspections, ensuring that the resulting data were not only structurally correct but also scientifically feasible and practically reasonable. Through this multi-stage pipeline, which combines automated validation, targeted repair, model-assisted review, and expert-level manual checks, SciRecipe achieved both rigor and diversity, providing a robust foundation for protocol-oriented training and evaluation.

\begin{table}[ht]
\centering
\caption{Disciplinary distribution of the SciRecipe dataset.}
\label{table1_app}
\resizebox{0.6\textwidth}{!}{%
\begin{tabular}{lcc}
\hline
\textbf{Subdomain}                  & \textbf{Count} & \textbf{Percentage} \\ \hline
Cell Biology                        & 2366           & 19.82\%             \\
Biochemistry                        & 1935           & 16.21\%             \\
Molecular Biology                   & 1788           & 14.98\%             \\
Microbiology                        & 1163           & 9.74\%              \\
Plant Science                       & 936            & 7.84\%              \\
Immunology                          & 717            & 6.01\%              \\
Neuroscience                        & 688            & 5.77\%              \\
General Laboratory Procedure        & 521            & 4.37\%              \\
Bioinformatics                      & 319            & 2.67\%              \\
Bioimaging Technologies             & 284            & 2.38\%              \\
Cancer Biology                      & 275            & 2.30\%              \\
Model Organism-Specific Techniques  & 246            & 2.06\%              \\
Others                              & 127            & 1.06\%              \\
Genomics Technologies               & 111            & 0.93\%              \\
Structural Biology Techniques       & 79             & 0.66\%              \\
Biophysics                          & 77             & 0.65\%              \\
Pharmacology \& Drug Development    & 67             & 0.56\%              \\
Developmental Biology               & 64             & 0.54\%              \\
Genetics                            & 55             & 0.46\%              \\
Histology Techniques                & 31             & 0.26\%              \\
Synthetic Biology \& Bioengineering & 31             & 0.26\%              \\
Stem Cells                          & 18             & 0.15\%              \\
Bioengineering                      & 11             & 0.09\%              \\
Toxicology \& Safety Testing        & 11             & 0.09\%              \\
Systems Biology                     & 8              & 0.07\%              \\
Medicine                            & 4              & 0.03\%              \\
Drug Discovery                      & 3              & 0.03\%              \\ \hline
\end{tabular}%
}
\end{table}

\begin{figure}[ht]
    \centering 
    \includegraphics[width=0.5\textwidth]{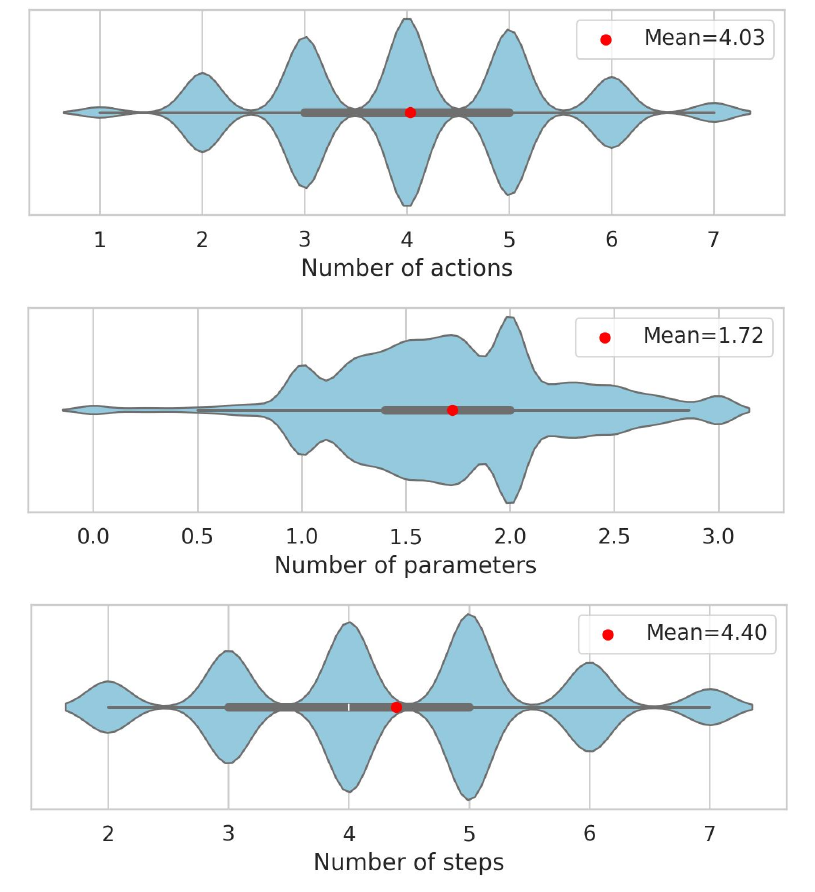} 
    \caption{Data distribution of experimental actions, parameters (each step), and step counts in SciRecipe after quality control.} 
    \label{fig6} 
\end{figure}

\section{SciRecipe-Eval}
\label{SciRecipe-Eval_app}
Cuurrently, benchmarks specifically dedicated to experimental protocol generation remain scarce, and most existing efforts are limited to tasks that simply restate the original protocol content. Such designs make it difficult to evaluate a model’s ability to generalize in realistic scientific contexts. Furthermore, the reliance on natural language alone as the ground truth introduces additional challenges for systematic evaluation.  
To address this gap, we constructed SciRecipe-Eval by following the design principles of SciRecipe while independently processing 400 protocols. The benchmark is organized into two major task categories, each containing approximately 300 QA pairs with a balanced distribution across subtasks.

\begin{itemize}
    \item \textbf{Protocol-Comprehension}: focuses on understanding and restructuring protocol content.  
    \begin{itemize}
        \item \textit{Overview}: generating hierarchical summaries of experimental procedures.  
        \item \textit{Specific}: decomposing steps to capture micro-level logic.  
    \end{itemize}  

    \item \textbf{Problem-Solving}: emphasizes reasoning and execution in realistic laboratory contexts.  
    \begin{itemize}
        \item \textit{Retrieval}: extracting precise parameters such as temperature, pH, or concentration.  
        \item \textit{Planning}: transforming objectives into coherent and logically ordered plans.  
        \item \textit{Troubleshooting}: diagnosing potential causes of experimental failure.  
        \item \textit{Constraint}: adapting protocols under resource limitations.  
        \item \textit{Scaling}: performing numerical adjustments and conversions for different scales.  
        \item \textit{Safety}: ensuring compliance and risk control.  
    \end{itemize}  
\end{itemize}  

Because the QAs were designed under specific reasoning paradigms, experimental actions (predicates) were treated as the core elements of the ground truth. The number of actions defines the size of the action space and thereby provides a direct measure of experimental complexity. Based on this criterion, tasks containing fewer than four actions were categorized as \textbf{Level 1}, whereas those with four or more actions were categorized as \textbf{Level 2}. For Level 2 tasks, we further inserted one to four randomly sampled distractor actions, enabling more comprehensive evaluation of a model’s ability to discriminate between correct and incorrect operations.  
In line with the data quality control standards established for SciRecipe, we recruited three doctoral students in biology-related fields to perform manual sampling and review of the dataset. Their verification ensured that the final benchmark was not only structurally consistent but also scientifically accurate and representative of real laboratory practices.

\section{SCORE Mechanism}
\subsection{Consistency Gate}
\label{Consistency Gate_app}
In protocol evaluation, \textbf{consistency} refers to the step-by-step correspondence between the structured outline (\texttt{<key>}) and its natural language expansion (\texttt{<orc>}). If an action, object, or parameter declared in \texttt{<key>} is missing from the corresponding \texttt{<orc>} description, the generated protocol is semantically incomplete and cannot serve as a reliable experimental guide. Therefore, the consistency gate is adopted as the second checkpoint in SCORE, and only protocols passing this constraint are eligible for reward computation.  

Concretely, we first require that the number of steps in \texttt{<key>} matches exactly with \texttt{<orc>}, with consecutive numbering covering $1$ through $N$. For each step $i$, we construct a token set $T_i$ by concatenating the declared action $a_i$, object set $\mathcal{O}_i$, and parameter set $\mathcal{P}_i$. After normalization (e.g., case folding, unit standardization, removal of subscript/superscript variants), we compute the coverage rate of these tokens in the natural language step $\texttt{<orc>}_i$ as:  
\begin{equation}
\mathrm{cov}_i = 
\frac{\big|\{t \in T_i : t \subseteq \mathrm{norm}(\texttt{<orc>}_i)\}\big|}{|T_i|}
\end{equation}

If all steps satisfy $\mathrm{cov}_i \geq \tau$ (we set $\tau = 0.95$ in this work) and the step indices are strictly aligned, the consistency gate is considered passed:  
\begin{equation}
\mathbb{I}_{\mathrm{cons}}(y) =
\begin{cases}
1, & \hat{N} = N \;\land\;
\{\hat{n}_1,\dots,\hat{n}_{\hat{N}}\} = \{1,\dots,N\} \;\land\;
\min_i \mathrm{cov}_i \ge \tau \\[6pt]
0, & \text{otherwise}
\end{cases}
\end{equation}
where $\hat{N}$ and $N$ denote the step counts of the generated and reference protocols, respectively, and $\hat{n}_i$ denotes the predicted step indices. If the gate fails, all subsequent rewards are set to zero. This hard gating design reflects a fundamental experimental principle: when the structured outline and its natural language realization are inconsistent, even partially correct fragments cannot make the protocol executable.

\subsection{Order Consistency}
\label{Order Consistency_app}
Let $\hat{\boldsymbol{a}}=(\hat{a}_1,\ldots,\hat{a}_{n})$ denote the predicted action sequence extracted from \texttt{<key>}, and $\boldsymbol{a}^*=(a^*_1,\ldots,a^*_{m})$ the reference sequence. The goal of order consistency is to evaluate whether the predicted execution order preserves the logical progression of the reference protocol, since deviations in ordering can undermine reproducibility even when individual steps appear correct. Two complementary scoring modes are provided.

The first mode, Strict Subsequence, enforces a conservative executability constraint. The score equals 1 if the predicted sequence is identical to the reference, or if one is a subsequence of the other; otherwise it is 0:
\begin{equation}
\mathrm{Order}_{\mathrm{strict}}(\hat{\boldsymbol{a}},\boldsymbol{a}^*)=
\begin{cases}
1, & \hat{\boldsymbol{a}}=\boldsymbol{a}^*
\;\lor\;\mathrm{subseq}(\hat{\boldsymbol{a}},\boldsymbol{a}^*)
\;\lor\;\mathrm{subseq}(\boldsymbol{a}^*,\hat{\boldsymbol{a}}) \\[6pt]
0, & \text{otherwise}
\end{cases}
\end{equation}
where $\mathrm{subseq}(\mathbf{x},\mathbf{y})$ holds when $\mathbf{x}$ can be embedded in $\mathbf{y}$ with order preserved. This criterion accepts insertions or omissions as long as the preserved actions remain in order, but strictly rejects any reordering. It is also computationally efficient, requiring only $O(n+m)$ time.

The second mode, Longest Common Subsequence (LCS), provides a graded similarity that reflects how much of the reference order is retained. The score is defined as:
\begin{equation}
\mathrm{Order}_{\mathrm{LCS}}(\hat{\boldsymbol{a}},\boldsymbol{a}^*)=
\frac{\mathrm{LCS}(\hat{\boldsymbol{a}},\boldsymbol{a}^*)}{m}
\end{equation}
where $m = |\boldsymbol{a}^*|$ is the length of the reference sequence and $\mathrm{LCS}(\cdot,\cdot)$ is the length of the longest common subsequence between the predicted and reference sequences. This normalization yields a value in $[0,1]$ that represents the fraction of the reference order preserved. Unlike the strict mode, LCS assigns partial credit when the prediction maintains some correct subsequences even if other steps are misplaced. For example, given the ground truth [harvest, lyse, centrifuge, quantify], a prediction [harvest, lyse, quantify] achieves a strict score of 1 and an LCS score of $3/4=0.75$ (omission but order preserved), whereas [harvest, centrifuge, lyse, quantify] scores 0 under the strict rule but $3/4=0.75$ under LCS, since most of the order is still preserved. By contrast, [lyse, harvest, quantify, centrifuge] is more severely misordered, yielding an LCS score of $2/4=0.5$.

For completeness, the LCS length is computed via dynamic programming. 
Let $L(i,j)$ denote the LCS length between the prefix subsequences 
$(\hat{a}_1,\dots,\hat{a}_i)$ of the predicted sequence $\hat{\boldsymbol{a}}$ 
and $(a^*_1,\dots,a^*_j)$ of the reference sequence $\boldsymbol{a}^*$. Then
\begin{equation}
L(i,j)=
\begin{cases}
0, & i=0 \ \,\, \text{or}\ \,\, j=0 \\[6pt]
L(i-1,j-1)+1, & \hat{a}_i=a^*_j \\[6pt]
\max\{L(i-1,j),\,L(i,j-1)\}, & \hat{a}_i \ne a^*_j
\end{cases}
\end{equation}
with $\mathrm{LCS}(\hat{\boldsymbol{a}},\boldsymbol{a}^*)=L(n,m)$. 
The algorithm runs in $O(nm)$ time and requires $O(nm)$ space.

Together, these two modes serve distinct but complementary purposes, with the strict subsequence mode being the primary criterion. Strict alignment best reflects the hard requirements of laboratory executability, where even minor order violations typically render a protocol unusable. The LCS mode is introduced only as a supplementary measure, offering smoother and more informative feedback during training by granting partial credit when some procedural intent is preserved. In practice, strict mode is preferred for evaluation of executable protocols, while LCS can be selectively applied to provide graded optimization signals when softer supervision is required. All formal experimental results in this work are reported under the strict mode. See Appendix \ref{Reward Computations} for further experiments.

\subsection{Action Anchor}
\label{Action Anchor_app}

In semantic consistency evaluation, actions (i.e., step verbs) are treated as anchors to align predicted steps with reference steps before any further comparison. The alignment is performed in a left-to-right, order-preserving manner. Formally, let the predicted action sequence be $\hat{\boldsymbol{a}}=(\hat{a}_1,\dots,\hat{a}_n)$ and the reference action sequence be $\boldsymbol{a}^*=(a^*_1,\dots,a^*_m)$. For each predicted action $\hat{a}_i$, we select the earliest reference index $j$ \emph{after} the previous match such that $a^*_j=\hat{a}_i$. If such a $j$ exists, we record the pair $(i,j)$, otherwise $\hat{a}i$ remains unmatched and contributes no anchor. This procedure yields an ordered set of index pairs $\mathcal{W}=\{(i_1,j_1),\dots,(i_K,j_K)\}$ with $1 \le i_1 < \cdots < i_K \le n$, $1 \le j_1 < \cdots < j_K \le m$, and $\hat{a}_{i_k}=a^*_{j_k}$ for all $(i_k,j_k)\in\mathcal{W}$. The construction is greedy and monotone (first-come–first-match), which guarantees that (i) sequence order is preserved, (ii) repeated actions are matched consistently, with each predicted occurrence aligning only to the next unused reference occurrence, and (iii) the overall alignment runs in linear time $O(n+m)$.

\textbf{Example}  
~For the ground truth sequence $[\textit{harvest},\ \textit{lyse},\ \textit{centrifuge},\ \textit{quantify}]$ and the prediction $[\textit{harvest},\ \textit{centrifuge},\ \textit{lyse},\ \textit{stain},\ \textit{quantify}]$, the alignment proceeds as
\[
\begin{aligned}
\hat{a}_1 &= \textit{harvest}   &&\mapsto\ a^*_1=\textit{harvest}, \\
\hat{a}_2 &= \textit{centrifuge}&&\mapsto\ a^*_3=\textit{centrifuge}, \\
\hat{a}_3 &= \textit{lyse}      &&\mapsto\ a^*_2=\textit{lyse}\ \ (\text{rejected: order violation}), \\
\hat{a}_5 &= \textit{quantify}  &&\mapsto\ a^*_4=\textit{quantify}.
\end{aligned}
\]
Hence the retained alignment is $\mathcal{W}=\{(1,1),\,(2,3),\,(5,4)\}$, while the unmatched action \textit{stain} is discarded. This anchor-based procedure ensures that semantic comparisons focus only on correctly aligned procedural units, avoiding misleading matches across unrelated operations.

\section{RL Algorithm}
\label{RL Algorithm}
During the RL stage, we adopt Group Relative Policy Optimization (GRPO) \citep{shao2024deepseekmath}, a variant of PPO designed to reduce variance and stabilize updates by exploiting multiple responses generated from the same query. Unlike standard PPO, which normalizes the reward across a batch, GRPO performs group-wise normalization within each query group. This strategy ensures that the relative quality of responses to the same query is emphasized, thereby mitigating reward scale differences across queries and improving training stability.
Formally, let $x$ denote a query, and $\{y_i\}_{i=1}^G$ be $G$ responses sampled from the current policy $\pi_\theta(\cdot|x)$. For each response $y_i$, the normalized advantage is defined as
\begin{equation}
\hat{A}_i = \frac{r(x,y_i) - \mathrm{mean}\big(\{r(x,y_j)\}_{j=1}^G\big)}
{\mathrm{std}\big(\{r(x,y_j)\}_{j=1}^G\big)}
\end{equation}
where $r(x,y_i)$ denotes the scalar reward assigned to $y_i$. By centering and scaling rewards within the group, the advantage highlights relative performance among candidate responses.

The GRPO objective combines this normalized advantage with the clipped surrogate function:
\begin{equation}
\mathcal{L}_{\mathrm{GRPO}}(\theta) =
\mathbb{E}_{x\sim\mathcal{D},\,\{y_i\}_{i=1}^G\sim\pi_\theta(\cdot|x)} \left[
\frac{1}{G}\sum_{i=1}^G \min\!\Big( s_i(\theta)\hat{A}_i,\,
\mathrm{clip}\!\big(s_i(\theta),\,1-\epsilon,\,1+\epsilon\big)\hat{A}_i \Big)
\right]
\end{equation}
where $\epsilon$ is a clipping threshold controlling the size of policy updates. This formulation prevents excessively large updates while still encouraging exploration.
The importance sampling ratio $s_i(\theta)$ is computed at the token level to account for the sequential nature of text generation. Specifically, it is defined as the geometric mean of the per-token probability ratios:
\begin{equation}
s_i(\theta) =
\exp\!\left(\tfrac{1}{|y_i|}\sum_{t=1}^{|y_i|}
\log \frac{\pi_\theta(y_{i,t}\mid x,y_{i,<t})}
{\pi_{\theta_{\mathrm{old}}}(y_{i,t}\mid x,y_{i,<t})}\right)
\end{equation}
where $\pi_\theta$ and $\pi_{\theta_{\mathrm{old}}}$ denote the current and reference policy models, respectively, and $y_{i,t}$ is the $t$-th token of response $y_i$. This per-token formulation maintains sensitivity to fine-grained action probabilities and avoids domination by sequence length. 
Overall, GRPO stabilizes training by normalizing advantages within query groups, constrains policy updates through clipping, and ensures robustness at the token level via per-token importance ratios.

\section{Baseline Models}
\label{Baseline Models_add}
Given the lack of models tailored for scientific protocol generation, we compare our approach against a broad set of large-scale language models. The closed-source baselines include the GPT-5 series \citep{gpt5chat} (GPT-5, GPT-5 Chat), the Claude series \citep{anthropic_claude} (Claude Sonnet 4, Claude Opus 4.1), the Gemini 2.5 series \citep{comanici2025gemini} (Flash and Pro), as well as ChatGPT-4o \citep{achiam2023gpt}, Doubao-1.5-Pro \citep{doubao15pro}, and Qwen2.5-Max \citep{qwen2025qwen25technicalreport}. For open-source baselines, we evaluate strong instruction-tuned models such as Qwen2.5-72B \citep{qwen2025qwen25technicalreport}, Qwen3-235B \citep{yang2025qwen3}, DeepSeek-V3 \citep{liu2024deepseek}, GPT-OSS-120B \citep{agarwal2025gpt}, Llama-3.1-405B \citep{grattafiori2024llama}, and Kimi-K2 \citep{team2025kimi}. In addition, we consider reasoning-oriented models including DeepSeek-R1 \citep{guo2025deepseek}, Grok 3/4 \citep{xai2025grok4}, and OpenAI’s o1 and o3 \citep{jaech2024openai}. Finally, to assess scientific capabilities, we further evaluate domain-specific models including Intern-S1-mini, Intern-S1 \citep{bai2025intern}, and SciDFM \citep{sun2024scidfm}. Together, these baselines provide a comprehensive landscape of frontier, open-source, reasoning, and domain-oriented models for comparison.

\section{Evaluation Metrics}
\label{Evaluation Metric_add}
It is noteworthy that SCORE is not merely a reward mechanism for RL but also naturally serves as an evaluation metric. In particular, the step scale reward, action order score, and object and parameter matching scores can all be directly used as assessment criteria for protocol generation tasks. Beyond this setting, the structured component-based comparison approach has the potential to generalize to broader scenarios, such as task planning in dialogue systems, generation of robotic operation scripts, or construction of reasoning chains in complex question-answering. By aligning the predicate–object–adverbial structure, we can evaluate not only the surface similarity of model outputs but also their reasoning depth and logical coherence. This provides a novel perspective for evaluating open-ended problems and highlights the broad applicability of the SCORE mechanism.

Formally, let the predicted key-step sequence be $\hat{\mathbf{s}}=(\hat{s}_1,\dots,\hat{s}_n)$ and the reference sequence be $\mathbf{s}^*=(s^*_1,\dots,s^*_m)$, with corresponding action sequences $\hat{\boldsymbol{a}}=(\hat{a}_1,\dots,\hat{a}_n)$ and $\boldsymbol{a}^*=(a^*_1,\dots,a^*_m)$. We denote by $\mathcal{W}=\{(i_k,j_k)\}_{k=1}^K$ the set of monotone action anchors obtained from the alignment procedure in Appendix~\ref{Action Anchor_app}. On this basis, we introduce five structured metrics for evaluation.

The first, Step-M, measures step-level completeness through strict length matching,
\begin{equation}
\mathrm{Step\mbox{-}M}(\hat{\mathbf{s}},\mathbf{s}^*)=\mathbb{I}\!\left[\,|\hat{\mathbf{s}}|=|\mathbf{s}^*|\,\right]
\end{equation}
The Order-S then enforces exact sequential agreement,
\begin{equation}
\mathrm{Order\mbox{-}S}(\hat{\boldsymbol{a}},\boldsymbol{a}^*)=\mathbb{I}\!\left[\,\hat{\boldsymbol{a}}=\boldsymbol{a}^*\,\right]
\end{equation}
while the more tolerant Order-LCS employs a normalized longest common subsequence,
\begin{equation}
\mathrm{Order\mbox{-}LCS}(\hat{\boldsymbol{a}},\boldsymbol{a}^*)=\tfrac{2\,L(\hat{\boldsymbol{a}},\boldsymbol{a}^*)}{n+m}
\end{equation}
where $L(\cdot,\cdot)$ denotes the LCS length. To further capture order fidelity, Order-Tau evaluates the concordance of action anchors using a Kendall-style correlation coefficient,
\begin{equation}
\mathrm{Order\mbox{-}Tau}(\mathcal{W})=\frac{C-D'}{C+D'}
\end{equation}
with $C$ and $D'$ the counts of concordant and discordant pairs in the aligned index sequence, defaulting to $0$ when $C+D'=0$.

Finally, Semantic-A is defined identically to the semantic consistency score introduced in Section~\ref{Reward Design}. It evaluates object overlap, parameter similarity, and positional fidelity on the basis of aligned action anchors, thereby reflecting both the executability and semantic correctness of generated protocols.

\section{Reproducibility Statement}
\label{Implementation Details_add}
We provide essential information to ensure the reproducibility of our experiments.
Table~\ref{table2_app} reports the training data distribution across stages, covering both SciRecipe tasks and BioProBench sub-tasks.
Key hyperparameters for pre-training, SFT, and RL are listed in Tables~\ref{tab:hyperparams}, \ref{tab:hyperparams_sft}, and \ref{tab:hyperparams_rl}.
All experiments were conducted with the LLaMA-Factory framework for pre-training and SFT, and VeRL for RL training.
We release detailed settings, including LoRA configurations, learning rates, epochs, scheduling, and hardware, so that independent researchers can replicate our pipeline and validate the reported results. All data, code, and models will be released publicly.
\begin{table}[ht]
\centering
\caption{Training data distribution across different stages (one epoch). ``PC'' denotes Protocol-Comprehension tasks and ``PS'' denotes Problem-Solving tasks. Four task types from BioProBench \citep{liu2025bioprobench} are also incorporated to enhance the model’s ability to interpret protocols.}
\label{table2_app}
\small
\begin{tabular}{@{}ccc@{}}
\toprule
Dataset         & \textbf{SFT stage} & \textbf{RL stage} \\ \midrule
SciRecipe\_PC   & 34461              & 6956              \\
SciRecipe\_PS    & 16164              & 3263              \\
BioProBench\_PQA & 52640              & \textemdash              \\
BioProBench\_ERR & 103983             & \textemdash      \\
BioProBench\_ORD & 31039              & \textemdash      \\
BioProBench\_GEN & 59477              & \textemdash      \\ \bottomrule
\end{tabular}%
\end{table}
\vspace{-13pt}
\begin{table}[ht]
\centering
\caption{Key hyperparameters used in LLaMA-Factory for pre-training.}
\label{tab:hyperparams}
\begin{tabular}{cc}
\toprule
\textbf{Parameter} & \textbf{Value} \\
\midrule
finetuning\_type & lora \\
lora\_rank & 8 \\
lora\_target & all \\
preprocessing\_num\_workers & 16 \\
dataloader\_num\_workers & 4 \\
per\_device\_train\_batch\_size & 1 \\
gradient\_accumulation\_steps & 8 \\
learning\_rate & 1e-4 \\
num\_train\_epochs & 6 \\
lr\_scheduler\_type & cosine \\
warmup\_ratio & 0.1 \\
bf16 & true \\
\bottomrule
\end{tabular}
\end{table}
\vspace{-13pt}
\begin{table}[ht]
\centering
\caption{Key hyperparameters used in LLaMA-Factory for SFT training.}
\label{tab:hyperparams_sft}
\begin{tabular}{cc}
\toprule
\textbf{Parameter} & \textbf{Value} \\
\midrule
finetuning\_type & lora \\
lora\_rank & 32 \\
lora\_target & all \\
preprocessing\_num\_workers & 8 \\
dataloader\_num\_workers & 4 \\
per\_device\_train\_batch\_size & 1 \\
gradient\_accumulation\_steps & 8 \\
learning\_rate & 3e-4 \\
num\_train\_epochs & 5 \\
lr\_scheduler\_type & cosine \\
warmup\_ratio & 0.05 \\
bf16 & true \\
\bottomrule
\end{tabular}
\end{table}
\begin{table}[ht]
\centering
\caption{Key hyperparameters used in VeRL for RL training.}
\label{tab:hyperparams_rl}
\begin{tabular}{cc}
\toprule
\textbf{Parameter} & \textbf{Value} \\
\midrule
algorithm.adv\_estimator & grpo \\
data.train\_batch\_size & 1024 \\
data.max\_prompt\_length & 1024 \\
data.max\_response\_length & 1024 \\
actor.optim.lr & 7e-6 \\
actor.ppo\_mini\_batch\_size & 256 \\
actor.ppo\_micro\_batch\_size\_per\_gpu & 32 \\
actor.use\_kl\_loss & true \\
actor.kl\_loss\_coef & 0.001 \\
actor.kl\_loss\_type & low\_var\_kl \\
actor.entropy\_coeff & 0 \\
rollout.tensor\_model\_parallel\_size & 2 \\
rollout.gpu\_memory\_utilization & 0.6 \\
rollout.n & 5 \\
trainer.total\_epochs & 15 \\
trainer.n\_gpus\_per\_node & 8 \\
trainer.nnodes & 1 \\
\bottomrule
\end{tabular}
\end{table}

\vspace{50pt}
\section{Limitations}
\label{Limitations}
While our work presents a structured framework for scientific protocol generation and shows consistent improvements across benchmarks, some limitations remain. First, although SciRecipe spans 27 biological subfields, it is still skewed toward widely used laboratory techniques, with rare or highly specialized protocols underrepresented. Second, the SCORE metrics mainly emphasize step completeness and order fidelity, but do not yet capture finer aspects such as stylistic variation, cross-step dependencies, or long-range experimental context. Third, our experiments were conducted in relatively controlled settings, which may not fully reflect the complexity and variability of real laboratory environments. We believe these issues can be progressively addressed in future work by expanding the dataset to cover rarer domains, enriching evaluation dimensions beyond structural executability, and validating the framework under more diverse practical scenarios.

\section{Usage of LLMs on Paper Writing}
\label{Usage of LLMs}
We used LLMs (e.g., GPT-5) to aid in polishing the writing of this paper. Specifically, LLMs were employed to refine grammar, improve fluency, and adjust the tone of academic writing (e.g., smoothing transitions between sections, shortening overly long sentences, and rephrasing repetitive expressions). The models were not used for generating technical content, conducting experiments, or formulating research ideas. All substantive contributions remain the work of the authors.
\section{Additional Experimental Results}
\label{Additional Experimental Results}

\subsection{Qualitative Study of Reasoning Models}
\label{Qualitative Study of Reasoning Models}
As discussed in the main text, general-purpose reasoning models often ``overthink'' protocol generation, producing overly long procedures with redundant actions and no clear stopping criteria. In the permeability assay, the gold-standard protocol requires only four steps (wash–remove–incubate–wash) to ensure structural preservation and experimental efficiency. Thoth closely followed this structure, preserving both the correct sequence and action fidelity. In contrast, Grok 4 expanded the task into a 13-step immunostaining workflow, adding permeabilization, blocking, antibody staining, and repeated washes far beyond the intended fixation step. Although logically coherent within its reasoning chain, this output diverges from laboratory practice, where clarity and parsimony are essential. This case underscores a broader limitation of general-purpose reasoning models: their tendency to prioritize exhaustive reasoning over practical executability, reinforcing our quantitative findings that domain-adapted models like Thoth achieve more concise, faithful, and actionable protocols.
\begin{figure}[ht]
    \centering 
    \includegraphics[width=0.9\textwidth]{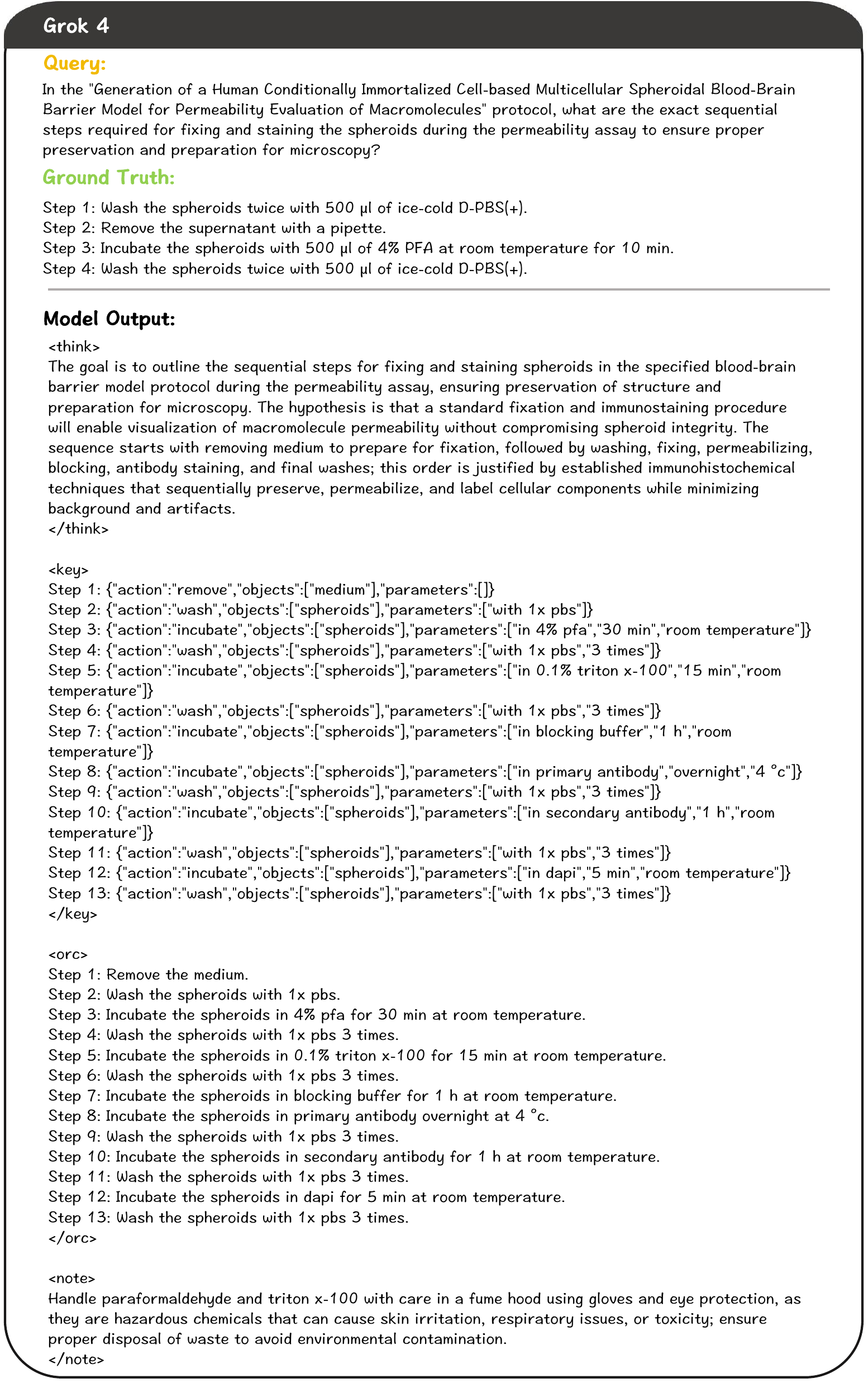} 
    \caption{Case study of the reasoning model Grok 4 on Protocol-Comprehension task.} 
    \label{fig7} 
\end{figure}

\begin{figure}[ht]
    \centering 
    \includegraphics[width=0.9\textwidth]{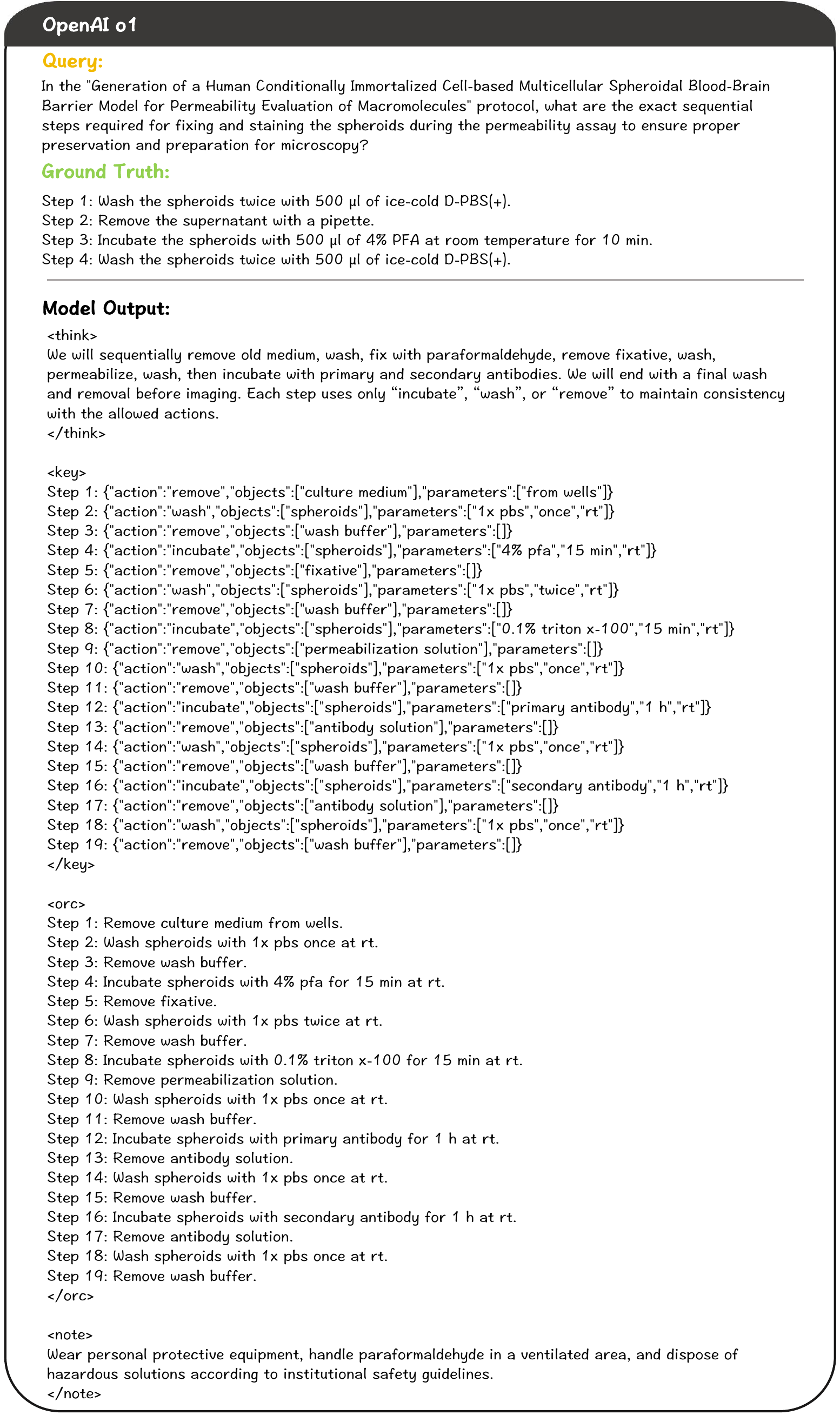} 
    \caption{Case study of the reasoning model o1 on Protocol-Comprehension tasks.} 
    \label{fig8} 
\end{figure}

\begin{figure}[ht]
    \centering 
    \includegraphics[width=0.9\textwidth]{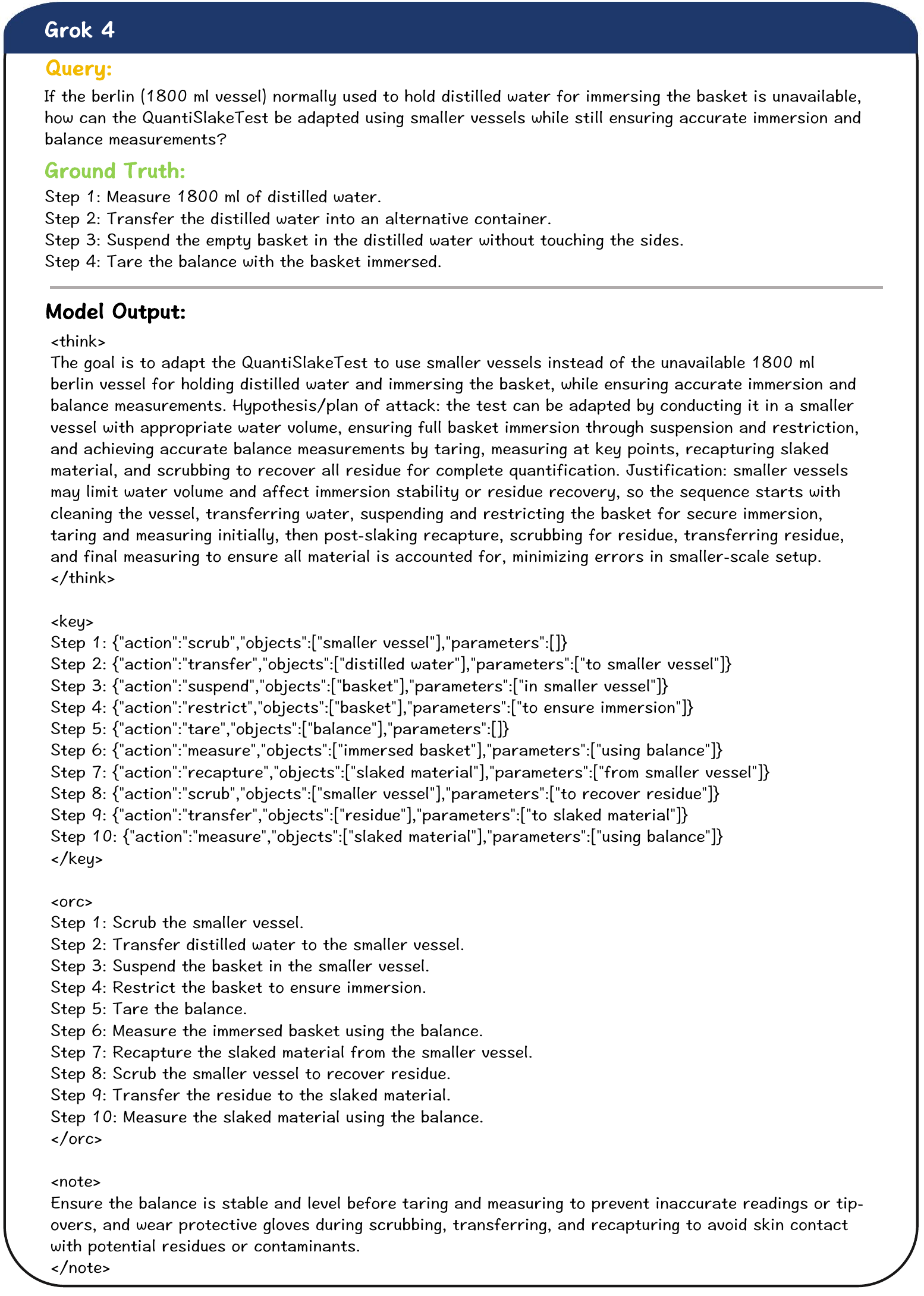} 
    \caption{Case study of the reasoning model Grok 4 on Problem-Solving task.} 
    \label{fig9} 
\end{figure}

\begin{figure}[ht]
    \centering 
    \includegraphics[width=0.9\textwidth]{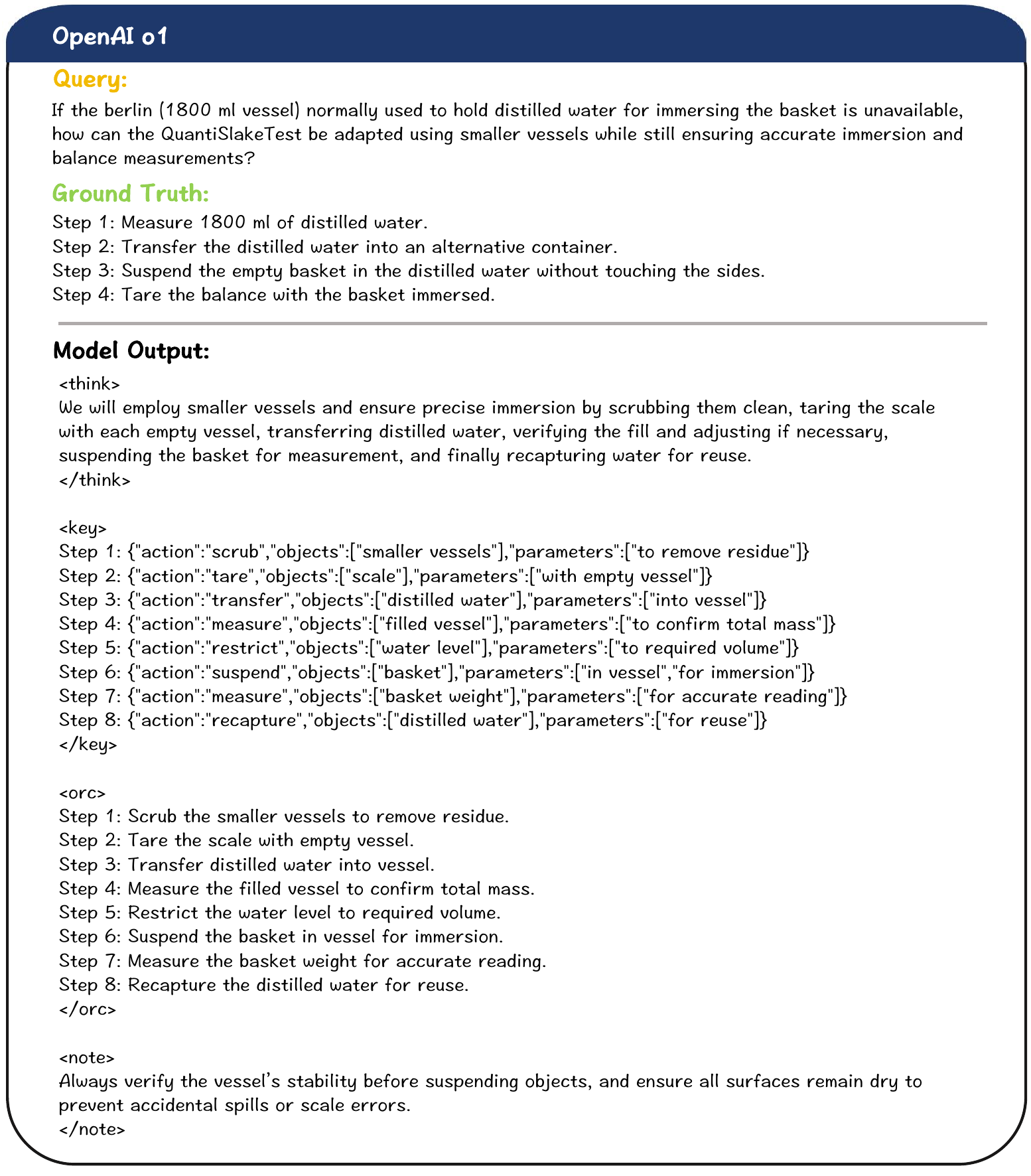} 
    \caption{Case study of the reasoning model o1 on Problem-Solving task.} 
    \label{fig10} 
\end{figure}

\clearpage
\subsection{Fine-grained Results on SciRecipe-Eval}
\label{Fine-grained Results on SciRecipe}

\renewcommand{\arraystretch}{1.1}
\begin{table}[h]
\centering
\caption{Main results on SciRecipe-Eval at level 1. Metrics left of the dashed line evaluate executability, those on the right measure semantic similarity. \textbf{Bold} denotes the best score.}
\label{TABLE9}
\resizebox{\textwidth}{!}{%
\begin{tabular}{@{}lccccc:ccccc@{}}
\toprule
\textbf{Methods} &
  \textbf{Semantic-A} &
  \textbf{Order-LCS} &
  \textbf{Order-S} &
  \textbf{Order-Tau} &
  \textbf{Step-M} &
  \textbf{BLUE-AVG} &
  \textbf{ROUGE-L} &
  \textbf{METEOR} &
  \textbf{KW-F1} &
  \textbf{AVG} \\ \midrule
\multicolumn{11}{l}{\textit{Close-Source SOTA}} \\
\textbf{ChatGPT-4o} & 44.44 & 78.95 & 39.49 & 80.04 & 56.49 & 40.98 & 52.11 & 48.02 & 55.37 & 55.10 \\
\textbf{GPT-5} & 32.82 & 63.40 & 19.27 & 63.63 & 27.62 & 22.34 & 35.92 & 35.41 & 42.61 & 38.11 \\
\textbf{GPT-5 Chat} & 41.50 & 78.91 & 35.75 & 73.50 & 35.06 & 31.74 & 45.77 & 45.80 & 50.60 & 48.74 \\
\textbf{Claude Sonnet 4} & 44.35 & 76.44 & 33.70 & 76.90 & 47.65 & 35.91 & 47.54 & 43.61 & 52.96 & 51.01 \\
\textbf{Claude Opus 4.1} & 45.43 & 75.42 & 34.82 & 76.88 & 46.89 & 36.29 & 46.90 & 42.14 & 52.36 & 50.79 \\
\textbf{Gemini 2.5 Flash} & 39.66 & 75.32 & 33.70 & 75.66 & 43.91 & 34.60 & 46.21 & 42.55 & 50.68 & 49.14 \\
\textbf{Gemini 2.5 Pro} & 38.78 & 77.51 & 36.08 & 76.56 & 40.17 & 32.71 & 46.84 & 47.84 & 50.64 & 49.68 \\
\textbf{Doubao-1.5-pro} & 37.14 & 78.42 & 38.47 & 76.00 & 60.23 & 39.78 & 49.57 & 41.77 & 51.50 & 52.54 \\
\textbf{Qwen2.5-Max} & 45.77 & 78.43 & 37.78 & 79.28 & 61.93 & 31.96 & 51.43 & 47.16 & 55.29 & 54.34 \\ \midrule
\multicolumn{11}{l}{\textit{Open-Source SOTA}} \\
\textbf{Qwen2.5-72B-Instruct} & 40.13 & 77.15 & 37.10 & 77.93 & 59.21 & 31.90 & 49.87 & 46.58 & 52.54 & 52.49 \\
\textbf{Qwen3-235B-A22B-Instruct-2507} & 39.68 & 78.19 & 35.08 & 78.28 & 54.82 & 35.68 & 48.10 & 47.11 & 51.20 & 52.02 \\
\textbf{DeepSeek-V3} & 45.10 & 78.59 & 33.84 & 75.64 & 52.27 & 39.66 & 51.46 & 48.04 & 54.55 & 53.24 \\
\textbf{GPT-OSS-120B} & 36.56 & 76.12 & 31.66 & 74.18 & 44.25 & 34.12 & 47.31 & 46.23 & 53.08 & 49.28 \\
\textbf{Llama-3.1-405B-Instruct} & 39.80 & 74.42 & 28.34 & 73.06 & 49.50 & 37.28 & 47.38 & 45.11 & 50.29 & 49.46 \\
\textbf{Kimi-K2-Instruction} & 42.58 & 76.96 & 34.72 & 78.60 & 52.07 & 35.40 & 47.65 & 45.78 & 51.61 & 51.71 \\ \midrule
\multicolumn{11}{l}{\textit{Reasoning Models}} \\
\textbf{DeepSeek-R1} & 37.86 & 77.08 & 35.43 & 76.35 & 45.97 & 40.81 & 48.96 & 42.45 & 52.56 & 50.83 \\
\textbf{Grok 3} & 41.53 & 78.84 & 37.68 & 77.96 & 54.19 & 37.12 & 49.82 & \textbf{49.98} & 52.28 & 53.27 \\
\textbf{Grok 4} & 42.52 & 77.27 & 34.92 & 75.66 & 49.52 & 38.91 & 49.75 & 43.72 & 54.34 & 51.85 \\
\textbf{OpenAI o1} & 39.78 & 77.70 & 28.00 & 74.43 & 43.02 & 37.25 & 49.23 & 46.19 & 53.12 & 49.86 \\
\textbf{OpenAI o3} & 42.15 & 74.94 & 26.02 & 72.95 & 36.29 & 31.10 & 46.42 & 46.78 & 53.59 & 47.80 \\ \midrule
\textbf{Qwen3-4B} & 27.21 & 59.83 & 24.19 & 59.89 & 40.17 & 16.29 & 28.17 & 27.22 & 31.41 & 34.93 \\
\rowcolor[HTML]{FFFFC7}
\textbf{Thoth-mini} & 49.49 & \textbf{80.41} & \textbf{42.88} & 80.30 & \textbf{64.66} & \textbf{46.50} & 51.91 & 48.24 & 55.77 & 57.80 \\
\textbf{Qwen3-8B} & 29.98 & 68.08 & 21.12 & 65.00 & 38.80 & 17.98 & 34.59 & 36.67 & 40.15 & 39.15 \\
\rowcolor[HTML]{FFFFC7}
\textbf{Thoth} & \textbf{52.21} & 80.12 & 41.19 & \textbf{83.20} & 62.27 & 46.41 & \textbf{52.69} & 49.21 & \textbf{56.88} & \textbf{58.24} \\ \bottomrule
\end{tabular}}
\end{table}
\vspace{30pt}
\renewcommand{\arraystretch}{1.1}
\begin{table}[h]
\centering
\caption{Main results on SciRecipe-Eval at level 2. Metrics left of the dashed line evaluate executability, those on the right measure semantic similarity. \textbf{Bold} denotes the best score.}
\label{TABLE10}
\resizebox{\textwidth}{!}{%
\begin{tabular}{@{}lccccc:ccccc@{}}
\toprule
\textbf{Methods} &
  \textbf{Semantic-A} &
  \textbf{Order-LCS} &
  \textbf{Order-S} &
  \textbf{Order-Tau} &
  \textbf{Step-M} &
  \textbf{BLUE-AVG} &
  \textbf{ROUGE-L} &
  \textbf{METEOR} &
  \textbf{KW-F1} &
  \textbf{AVG} \\ \midrule
\multicolumn{11}{l}{\textit{Close-Source SOTA}} \\
\textbf{ChatGPT-4o} & 35.82 & 67.82 & 9.12 & 61.01 & 32.00 & 37.00 & 44.88 & 41.44 & 48.87 & 42.00 \\
\textbf{GPT-5} & 22.96 & 53.05 & 3.75 & 43.87 & 10.31 & 20.33 & 30.12 & 29.81 & 35.87 & 27.79 \\
\textbf{GPT-5 Chat} & 31.31 & 67.74 & 7.17 & 58.15 & 15.34 & 27.49 & 38.46 & 38.26 & 45.25 & 36.57 \\
\textbf{Claude Sonnet 4} & 34.55 & 67.68 & 8.47 & 63.38 & 24.48 & 32.64 & 41.13 & 38.44 & 45.98 & 39.64 \\
\textbf{Claude Opus 4.1} & 37.38 & 68.13 & 9.30 & \textbf{67.18} & 22.78 & 33.16 & 42.04 & 38.65 & 47.74 & 40.71 \\
\textbf{Gemini 2.5 Flash} & 33.17 & 66.09 & 6.84 & 65.21 & 21.21 & 31.84 & 39.74 & 36.10 & 45.57 & 38.42 \\
\textbf{Gemini 2.5 Pro} & 32.94 & 68.04 & 8.14 & 64.04 & 24.16 & 30.09 & 41.59 & 43.43 & 46.61 & 39.89 \\
\textbf{Doubao-1.5-pro} & 29.67 & 68.37 & 9.46 & 64.24 & 35.27 & 36.61 & 44.30 & 35.77 & 46.09 & 41.09 \\
\textbf{Qwen2.5-Max} & 35.13 & 67.55 & 6.51 & 63.70 & 33.64 & 29.71 & 44.75 & 40.62 & 48.80 & 41.16 \\ \midrule
\multicolumn{11}{l}{\textit{Open-Source SOTA}} \\
\textbf{Qwen2.5-72B-Instruct} & 32.82 & 64.74 & 5.54 & 60.76 & 25.80 & 27.30 & 42.40 & 40.43 & 47.04 & 38.54 \\
\textbf{Qwen3-235B-A22B-Instruct-2507} & 31.84 & 66.19 & 5.58 & 60.32 & 21.32 & 29.41 & 40.79 & 41.61 & 44.71 & 37.97 \\
\textbf{DeepSeek-V3} & 38.48 & 69.54 & 9.53 & 65.64 & 31.57 & 36.76 & 45.64 & 42.24 & 50.20 & 43.29 \\
\textbf{GPT-OSS-120B} & 29.31 & 64.07 & 4.23 & 54.56 & 12.06 & 27.46 & 39.73 & 39.71 & 46.89 & 35.34 \\
\textbf{Llama-3.1-405B-Instruct} & 32.20 & 64.70 & 8.13 & 62.71 & 29.37 & 35.60 & 42.44 & 39.35 & 45.98 & 40.05 \\
\textbf{Kimi-K2-Instruction} & 31.62 & 66.91 & 7.49 & 61.41 & 28.41 & 32.29 & 41.46 & 40.31 & 46.50 & 39.60 \\ \midrule
\multicolumn{11}{l}{\textit{Reasoning Models}} \\
\textbf{DeepSeek-R1} & 34.36 & 65.91 & 5.91 & 62.18 & 20.33 & 38.83 & 42.89 & 34.10 & 47.21 & 39.08 \\
\textbf{Grok 3} & 33.44 & 67.92 & 6.78 & 61.83 & 24.90 & 32.42 & 42.41 & 42.59 & 45.05 & 39.70 \\
\textbf{Grok 4} & 31.17 & 67.10 & 6.16 & 55.43 & 20.39 & 35.58 & 42.74 & 36.78 & 49.38 & 38.30 \\
\textbf{OpenAI o1} & 29.90 & 69.27 & 9.44 & 60.75 & 28.06 & 34.18 & 43.47 & 41.55 & 47.26 & 40.43 \\
\textbf{OpenAI o3} & 28.92 & 66.00 & 5.16 & 57.46 & 12.35 & 26.24 & 39.88 & 41.98 & 47.07 & 36.12 \\ \midrule
\textbf{Qwen3-4B} & 21.65 & 47.52 & 3.57 & 41.48 & 17.94 & 12.82 & 21.45 & 20.81 & 24.12 & 23.48 \\
\rowcolor[HTML]{FFFFC7}
\textbf{Thoth-mini} & 39.28 & 69.18 & 8.47 & 61.74 & 41.16 & 40.27 & 46.66 & 44.66 & 50.60 & 44.67 \\
\textbf{Qwen3-8B} & 27.85 & 59.12 & 1.62 & 52.59 & 10.43 & 15.40 & 30.12 & 32.85 & 37.17 & 29.68 \\
\rowcolor[HTML]{FFFFC7}
\textbf{Thoth} & \textbf{41.21} & \textbf{70.75} & \textbf{10.43} & 63.85 & \textbf{44.10} & \textbf{40.94} & \textbf{47.46} & \textbf{45.65} & \textbf{51.49} & \textbf{46.21} \\ \bottomrule
\end{tabular}}
\end{table}

\renewcommand{\arraystretch}{1.1}
\begin{table}[h]
\centering
\caption{Main results on SciRecipe-Eval (Overview). Metrics left of the dashed line evaluate executability, those on the right measure semantic similarity. \textbf{Bold} denotes the best score.}
\label{TABLE11}
\resizebox{\textwidth}{!}{%
\begin{tabular}{@{}lccccc:ccccc@{}}
\toprule
\textbf{Methods} &
  \textbf{Semantic-A} &
  \textbf{Order-LCS} &
  \textbf{Order-S} &
  \textbf{Order-Tau} &
  \textbf{Step-M} &
  \textbf{BLUE-AVG} &
  \textbf{ROUGE-L} &
  \textbf{METEOR} &
  \textbf{KW-F1} &
  \textbf{AVG} \\ \midrule
\multicolumn{11}{l}{\textit{Close-Source SOTA}} \\
\textbf{ChatGPT-4o} & 32.36 & 72.47 & 13.33 & 72.02 & 30.67 & 34.24 & 42.36 & 37.24 & 44.03 & 42.08 \\
\textbf{GPT-5} & 27.85 & 65.51 & 9.59 & 64.64 & 16.29 & 23.26 & 33.20 & 30.29 & 35.30 & 33.99 \\
\textbf{GPT-5 Chat} & 30.99 & 70.02 & 10.01 & 68.01 & 11.33 & 21.35 & 32.70 & 32.98 & 38.10 & 35.05 \\
\textbf{Claude Sonnet 4} & 30.08 & 72.64 & \textbf{16.00} & 71.33 & 30.67 & 31.42 & 38.94 & 35.62 & 41.20 & 40.88 \\
\textbf{Claude Opus 4.1} & 30.72 & 71.52 & 14.52 & 68.38 & 24.69 & 31.65 & 39.70 & 36.02 & 42.32 & 39.95 \\
\textbf{Gemini 2.5 Flash} & 28.87 & 70.62 & 12.67 & 73.67 & 26.00 & 31.85 & 39.04 & 33.00 & 41.57 & 39.70 \\
\textbf{Gemini 2.5 Pro} & 28.96 & 71.83 & 12.00 & 75.35 & 26.67 & 28.04 & 38.99 & 38.83 & 41.38 & 40.23 \\
\textbf{Doubao-1.5-pro} & 26.85 & 71.67 & 12.01 & 76.34 & 34.67 & 37.87 & 42.05 & 30.53 & 41.36 & 41.48 \\
\textbf{Qwen2.5-Max} & 34.73 & 71.56 & 10.00 & 76.67 & 36.00 & 28.11 & 42.12 & 36.27 & 44.08 & 42.17 \\ \midrule
\multicolumn{11}{l}{\textit{Open-Source SOTA}} \\
\textbf{Qwen2.5-72B-Instruct} & 30.22 & 69.32 & 10.00 & 75.35 & 27.35 & 24.89 & 39.97 & 37.00 & 42.17 & 39.59 \\
\textbf{Qwen3-235B-A22B-Instruct-2507} & 28.70 & 68.69 & 8.67 & 70.70 & 21.33 & 26.67 & 37.29 & 37.33 & 39.26 & 37.63 \\
\textbf{DeepSeek-V3} & 35.24 & \textbf{73.23} & 15.54 & 71.01 & 37.16 & 35.33 & \textbf{43.43} & 39.58 & 43.86 & \textbf{43.82} \\
\textbf{GPT-OSS-120B} & 27.34 & 67.29 & 10.01 & 64.68 & 20.00 & 24.75 & 36.10 & 36.71 & 39.61 & 36.28 \\
\textbf{Llama-3.1-405B-Instruct} & 31.99 & 71.19 & 14.67 & \textbf{79.36} & 32.00 & 36.59 & 40.78 & 35.59 & 40.62 & 42.53 \\
\textbf{Kimi-K2-Instruction} & 30.92 & 72.21 & 12.00 & 74.67 & 36.67 & 30.90 & 39.20 & 37.88 & 41.18 & 41.74 \\ \midrule
\multicolumn{11}{l}{\textit{Reasoning Models}} \\
\textbf{DeepSeek-R1} & 31.65 & 70.90 & 12.00 & 69.67 & 27.34 & \textbf{40.45} & 42.47 & 32.09 & 43.34 & 41.10 \\
\textbf{Grok 3} & 30.56 & 70.29 & 12.42 & 64.01 & 26.86 & 30.53 & 39.60 & \textbf{40.02} & 40.21 & 39.39 \\
\textbf{Grok 4} & 32.19 & 72.43 & 12.52 & 67.45 & 29.30 & 36.06 & 42.33 & 36.12 & \textbf{45.54} & 41.55 \\
\textbf{OpenAI o1} & 27.78 & 72.80 & 12.67 & 70.70 & 31.33 & 31.80 & 40.20 & 37.84 & 40.75 & 40.65 \\
\textbf{OpenAI o3} & 28.48 & 71.01 & 8.61 & 72.00 & 20.64 & 25.24 & 37.21 & 38.26 & 41.18 & 38.07 \\ \midrule
\textbf{Qwen3-4B} & 28.05 & 62.52 & 8.02 & 62.00 & 24.00 & 13.46 & 24.50 & 25.40 & 27.38 & 30.59 \\
\rowcolor[HTML]{FFFFC7}
\textbf{Thoth-mini} & 37.49 & 71.74 & 11.34 & 71.67 & \textbf{38.34} & 38.02 & 42.61 & 38.55 & 43.79 & 43.73 \\
\textbf{Qwen3-8B} & 30.46 & 65.41 & 2.68 & 62.01 & 10.67 & 11.60 & 26.80 & 31.97 & 32.14 & 30.42 \\
\rowcolor[HTML]{FFFFC7}
\textbf{Thoth} & \textbf{40.15} & 72.10 & 9.35 & 71.01 & 36.35 & 37.75 & 43.27 & 39.09 & 44.79 & 43.76 \\ \bottomrule
\end{tabular}}
\end{table}
\renewcommand{\arraystretch}{1.1}
\begin{table}[h]
\centering
\caption{Main results on SciRecipe-Eval (Specific). Metrics left of the dashed line evaluate executability, those on the right measure semantic similarity. \textbf{Bold} denotes the best score.}
\label{TABLE12}
\resizebox{\textwidth}{!}{%
\begin{tabular}{@{}lccccc:ccccc@{}}
\toprule
\textbf{Methods} &
  \textbf{Semantic-A} &
  \textbf{Order-LCS} &
  \textbf{Order-S} &
  \textbf{Order-Tau} &
  \textbf{Step-M} &
  \textbf{BLUE-AVG} &
  \textbf{ROUGE-L} &
  \textbf{METEOR} &
  \textbf{KW-F1} &
  \textbf{AVG} \\ \midrule
\multicolumn{11}{l}{\textit{Close-Source SOTA}} \\
\textbf{ChatGPT-4o} & 33.63 & 70.95 & 24.00 & 67.32 & 48.67 & 36.26 & 45.07 & 41.29 & 46.81 & 46.00 \\
\textbf{GPT-5} & 18.99 & 48.00 & 12.59 & 43.55 & 24.26 & 19.21 & 27.31 & 24.67 & 28.01 & 27.40 \\
\textbf{GPT-5 Chat} & 27.99 & 71.05 & 20.67 & 64.67 & 18.67 & 23.93 & 36.23 & 36.46 & 39.56 & 37.69 \\
\textbf{Claude Sonnet 4} & 36.39 & 66.46 & 20.67 & 66.67 & 37.33 & 33.11 & 40.55 & 35.55 & 42.88 & 42.18 \\
\textbf{Claude Opus 4.1} & 37.78 & 67.09 & 20.18 & 67.81 & 35.95 & 35.76 & 40.69 & 33.60 & 44.92 & 42.64 \\
\textbf{Gemini 2.5 Flash} & 36.67 & 68.25 & 20.00 & 70.33 & 30.67 & 31.72 & 40.34 & 36.03 & 43.34 & 41.93 \\
\textbf{Gemini 2.5 Pro} & 35.48 & 71.68 & 23.33 & 69.33 & 30.00 & 27.89 & 40.09 & 41.84 & 42.08 & 42.41 \\
\textbf{Doubao-1.5-pro} & 30.96 & 71.96 & 26.00 & 66.33 & 52.00 & 37.91 & 44.14 & 35.84 & 43.69 & 45.43 \\
\textbf{Qwen2.5-Max} & 36.91 & 71.18 & 23.33 & 64.67 & 46.67 & 27.73 & 45.37 & 40.97 & 46.94 & 44.86 \\ \midrule
\multicolumn{11}{l}{\textit{Open-Source SOTA}} \\
\textbf{Qwen2.5-72B-Instruct} & 36.45 & 68.45 & 23.33 & 67.33 & 43.33 & 28.58 & 43.82 & 41.51 & 44.89 & 44.19 \\
\textbf{Qwen3-235B-A22B-Instruct-2507} & 32.02 & 70.81 & 23.49 & 64.45 & 46.31 & 30.86 & 42.41 & 41.47 & 43.07 & 43.88 \\
\textbf{DeepSeek-V3} & 37.84 & 70.37 & 17.99 & 69.34 & 39.33 & 33.72 & 43.54 & 40.02 & 46.49 & 44.29 \\
\textbf{GPT-OSS-120B} & 32.81 & 68.77 & 22.67 & 64.67 & 35.33 & 30.02 & 42.74 & 40.32 & 46.92 & 42.69 \\
\textbf{Llama-3.1-405B-Instruct} & 35.87 & 67.13 & 19.99 & 62.70 & 39.33 & 35.22 & 43.75 & 40.18 & 45.16 & 43.26 \\
\textbf{Kimi-K2-Instruction} & 36.53 & 69.21 & 20.67 & 70.00 & 38.67 & 32.87 & 41.68 & 38.99 & 45.07 & 43.74 \\ \midrule
\multicolumn{11}{l}{\textit{Reasoning Models}} \\
\textbf{DeepSeek-R1} & 32.25 & 70.05 & 26.67 & 68.99 & 40.00 & 39.22 & 42.88 & 33.83 & 44.65 & 44.28 \\
\textbf{Grok 3} & 34.28 & 71.43 & 21.95 & \textbf{72.81} & 37.72 & 34.00 & 43.73 & 42.28 & 43.09 & 44.59 \\
\textbf{Grok 4} & 37.94 & 68.65 & 16.07 & 66.90 & 34.22 & 35.45 & 43.69 & 37.98 & 48.54 & 43.27 \\
\textbf{OpenAI o1} & 31.77 & 70.89 & 18.12 & 63.10 & 34.23 & 33.31 & 43.55 & 40.49 & 45.53 & 42.33 \\
\textbf{OpenAI o3} & 29.05 & 67.67 & 15.27 & 59.33 & 26.64 & 24.97 & 39.31 & 41.27 & 44.71 & 38.69 \\ \midrule
\textbf{Qwen3-4B} & 23.34 & 59.81 & 19.33 & 58.00 & 41.33 & 16.21 & 28.38 & 26.77 & 29.94 & 33.68 \\
\rowcolor[HTML]{FFFFC7}
\textbf{Thoth-mini} & 45.21 & \textbf{75.14} & \textbf{28.68} & 69.67 & \textbf{54.34} & \textbf{42.41} & \textbf{48.91} & \textbf{45.34} & \textbf{51.32} & \textbf{51.22} \\
\textbf{Qwen3-8B} & 28.54 & 67.34 & 14.00 & 62.00 & 26.00 & 15.92 & 32.10 & 35.22 & 37.55 & 35.41 \\
\rowcolor[HTML]{FFFFC7}
\textbf{Thoth} & \textbf{45.50} & 73.70 & 27.34 & 70.34 & 54.33 & 42.15 & 48.19 & 44.89 & 50.72 & 50.80 \\ \bottomrule
\end{tabular}}
\end{table}

\renewcommand{\arraystretch}{1.1}
\begin{table}[h]
\centering
\caption{Main results on SciRecipe-Eval (Retrieval). Metrics left of the dashed line evaluate executability, those on the right measure semantic similarity. \textbf{Bold} denotes the best score.}
\label{TABLE13}
\resizebox{\textwidth}{!}{%
\begin{tabular}{@{}lccccc:ccccc@{}}
\toprule
\textbf{Methods} &
  \textbf{Semantic-A} &
  \textbf{Order-LCS} &
  \textbf{Order-S} &
  \textbf{Order-Tau} &
  \textbf{Step-M} &
  \textbf{BLUE-AVG} &
  \textbf{ROUGE-L} &
  \textbf{METEOR} &
  \textbf{KW-F1} &
  \textbf{AVG} \\ \midrule
\multicolumn{11}{l}{\textit{Close-Source SOTA}} \\
\textbf{ChatGPT-4o} & 51.78 & 80.87 & 42.00 & 73.99 & 60.00 & 46.27 & 58.29 & 53.51 & 61.29 & 58.67 \\
\textbf{GPT-5} & 39.77 & 69.06 & 18.72 & 48.16 & 29.02 & 24.88 & 41.52 & 41.95 & 51.04 & 40.46 \\
\textbf{GPT-5 Chat} & 54.81 & 81.86 & 40.00 & 72.00 & 52.00 & 42.55 & 56.07 & 54.05 & 61.58 & 57.21 \\
\textbf{Claude Sonnet 4} & 53.10 & 80.66 & 40.00 & \textbf{78.00} & 48.00 & 44.13 & 55.60 & 50.88 & 60.91 & 56.81 \\
\textbf{Claude Opus 4.1} & 58.97 & 82.42 & 39.75 & 72.07 & 45.36 & 42.26 & 56.09 & 50.66 & 61.08 & 56.52 \\
\textbf{Gemini 2.5 Flash} & 48.78 & 75.45 & 36.00 & 71.00 & 48.00 & 38.37 & 50.88 & 47.87 & 56.35 & 52.52 \\
\textbf{Gemini 2.5 Pro} & 45.71 & 78.66 & 40.00 & 70.00 & 56.00 & 38.62 & 52.35 & 55.49 & 59.20 & 55.11 \\
\textbf{Doubao-1.5-pro} & 43.40 & 78.01 & 34.00 & 67.00 & 54.00 & 39.83 & 54.14 & 47.25 & 57.64 & 52.81 \\
\textbf{Qwen2.5-Max} & 51.18 & 81.84 & 46.00 & \textbf{78.00} & 58.00 & 34.33 & 58.01 & 51.88 & 59.62 & 57.65 \\ \midrule
\multicolumn{11}{l}{\textit{Open-Source SOTA}} \\
\textbf{Qwen2.5-72B-Instruct} & 50.77 & 82.11 & 44.00 & 70.00 & \textbf{70.00} & 35.23 & 56.68 & 53.80 & 60.13 & 58.08 \\
\textbf{Qwen3-235B-A22B-Instruct-2507} & 55.90 & 80.23 & 38.00 & 70.00 & 59.97 & 43.30 & 56.28 & 54.85 & 59.04 & 57.51 \\
\textbf{DeepSeek-V3} & 57.83 & 80.95 & 42.83 & 69.48 & 53.06 & 46.00 & 59.70 & 55.70 & 62.66 & 58.69 \\
\textbf{GPT-OSS-120B} & 47.62 & 80.48 & 38.00 & \textbf{78.00} & 44.00 & 41.26 & 55.77 & 52.12 & 59.97 & 55.02 \\
\textbf{Llama-3.1-405B-Instruct} & 42.10 & 72.88 & 28.54 & 59.18 & 44.89 & 38.83 & 50.29 & 49.29 & 54.95 & 48.99 \\
\textbf{Kimi-K2-Instruction} & 45.13 & 79.94 & 38.00 & 74.00 & 48.00 & 41.45 & 52.77 & 50.93 & 58.59 & 54.31 \\ \midrule
\multicolumn{11}{l}{\textit{Reasoning Models}} \\
\textbf{DeepSeek-R1} & 44.54 & 75.14 & 27.97 & 72.97 & 36.00 & 42.01 & 51.71 & 44.71 & 56.55 & 50.18 \\
\textbf{Grok 3} & 52.84 & 81.15 & 42.12 & 70.74 & 54.50 & 39.45 & 54.61 & 55.06 & 57.88 & 56.48 \\
\textbf{Grok 4} & 44.04 & 76.50 & 30.33 & 68.00 & 40.67 & 39.83 & 51.44 & 43.08 & 57.84 & 50.19 \\
\textbf{OpenAI o1} & 50.51 & 80.38 & 32.00 & \textbf{78.00} & 42.00 & 41.74 & 55.37 & 51.96 & 60.02 & 54.44 \\
\textbf{OpenAI o3} & 51.73 & 77.50 & 33.27 & 66.67 & 35.39 & 35.63 & 51.02 & 51.68 & 60.29 & 51.46 \\ \midrule
\textbf{Qwen3-4B} & 21.50 & 43.11 & 18.00 & 38.00 & 22.00 & 12.84 & 21.80 & 20.96 & 24.10 & 24.70 \\
\rowcolor[HTML]{FFFFC7}
\textbf{Thoth-mini} & \textbf{59.35} & \textbf{85.47} & 45.98 & 76.34 & 66.34 & \textbf{50.29} & 59.62 & 57.24 & 64.06 & \textbf{62.74} \\
\textbf{Qwen3-8B} & 32.20 & 62.75 & 24.00 & 54.00 & 36.00 & 19.32 & 35.11 & 35.19 & 39.29 & 37.54 \\
\rowcolor[HTML]{FFFFC7}
\textbf{Thoth} & 54.51 & 84.64 & \textbf{47.99} & 75.65 & 67.66 & 48.55 & \textbf{59.76} & \textbf{59.03} & \textbf{64.92} & 62.52 \\ \bottomrule
\end{tabular}}
\end{table}

\renewcommand{\arraystretch}{1.1}
\begin{table}[h]
\centering
\caption{Main results on SciRecipe-Eval (Planning). Metrics left of the dashed line evaluate executability, those on the right measure semantic similarity. \textbf{Bold} denotes the best score.}
\label{TABLE14}
\resizebox{\textwidth}{!}{%
\begin{tabular}{@{}lccccc:ccccc@{}}
\toprule
\textbf{Methods} &
  \textbf{Semantic-A} &
  \textbf{Order-LCS} &
  \textbf{Order-S} &
  \textbf{Order-Tau} &
  \textbf{Step-M} &
  \textbf{BLUE-AVG} &
  \textbf{ROUGE-L} &
  \textbf{METEOR} &
  \textbf{KW-F1} &
  \textbf{AVG} \\ \midrule
\multicolumn{11}{l}{\textit{Close-Source SOTA}} \\
\textbf{ChatGPT-4o} & 44.20 & 76.84 & 34.01 & 79.99 & 60.00 & 42.65 & 53.16 & 49.96 & 56.70 & 55.28 \\
\textbf{GPT-5} & 27.41 & 62.34 & 8.68 & 58.95 & 10.72 & 21.93 & 35.86 & 36.76 & 44.45 & 34.12 \\
\textbf{GPT-5 Chat} & 33.96 & 75.62 & 22.00 & 62.00 & 26.00 & 32.32 & 45.75 & 47.28 & 52.21 & 44.13 \\
\textbf{Claude Sonnet 4} & 38.06 & 73.34 & 22.00 & 64.00 & 36.00 & 35.42 & 47.67 & 44.29 & 52.51 & 45.92 \\
\textbf{Claude Opus 4.1} & \textbf{51.68} & 76.84 & 23.69 & \textbf{82.66} & 53.24 & 39.58 & 50.28 & 44.81 & 55.88 & 53.18 \\
\textbf{Gemini 2.5 Flash} & 40.29 & 76.34 & 26.00 & 65.00 & 28.00 & 33.95 & 44.78 & 42.62 & 50.99 & 45.33 \\
\textbf{Gemini 2.5 Pro} & 41.38 & 75.86 & 26.00 & 68.00 & 36.00 & 35.37 & 50.58 & 51.43 & 54.52 & 48.79 \\
\textbf{Doubao-1.5-pro} & 34.52 & 73.33 & 24.01 & 68.99 & 62.00 & 39.11 & 49.19 & 43.81 & 53.34 & 49.81 \\
\textbf{Qwen2.5-Max} & 43.08 & 72.84 & 22.00 & 62.00 & 54.00 & 32.97 & 51.46 & 49.50 & 57.75 & 49.51 \\ \midrule
\multicolumn{11}{l}{\textit{Open-Source SOTA}} \\
\textbf{Qwen2.5-72B-Instruct} & 39.57 & 73.26 & 24.01 & 64.00 & 50.00 & 32.45 & 51.16 & 48.38 & 54.94 & 48.64 \\
\textbf{Qwen3-235B-A22B-Instruct-2507} & 31.86 & 74.41 & 22.00 & 69.99 & 39.99 & 33.99 & 47.24 & 48.51 & 51.82 & 46.65 \\
\textbf{DeepSeek-V3} & 45.53 & 76.42 & 21.98 & 76.01 & 46.00 & 40.97 & 52.75 & 49.20 & 57.95 & 51.87 \\
\textbf{GPT-OSS-120B} & 37.09 & 71.76 & 18.00 & 61.99 & 32.00 & 34.02 & 46.87 & 48.50 & 56.85 & 45.23 \\
\textbf{Llama-3.1-405B-Instruct} & 33.79 & 63.95 & 11.96 & 54.00 & 44.00 & 34.96 & 45.09 & 44.00 & 49.95 & 42.41 \\
\textbf{Kimi-K2-Instruction} & 33.59 & 71.64 & 26.00 & 62.00 & 52.00 & 35.83 & 48.26 & 47.39 & 51.41 & 47.57 \\ \midrule
\multicolumn{11}{l}{\textit{Reasoning Models}} \\
\textbf{DeepSeek-R1} & 43.05 & 75.01 & 23.97 & 74.98 & 42.01 & 42.10 & 50.88 & 45.40 & 54.77 & 50.24 \\
\textbf{Grok 3} & 40.66 & 77.72 & 33.78 & 66.59 & 52.72 & 36.60 & 50.29 & 52.12 & 52.12 & 51.40 \\
\textbf{Grok 4} & 39.19 & 76.44 & \textbf{34.63} & 63.54 & 45.69 & 39.96 & 51.50 & 44.76 & 56.70 & 50.27 \\
\textbf{OpenAI o1} & 35.64 & 75.73 & 20.00 & 64.00 & 42.00 & 37.93 & 49.47 & 48.84 & 54.52 & 47.57 \\
\textbf{OpenAI o3} & 34.12 & 70.76 & 15.94 & 58.00 & 23.97 & 31.74 & 47.89 & 50.66 & 57.41 & 43.39 \\ \midrule
\textbf{Qwen3-4B} & 28.50 & 48.20 & 13.99 & 48.00 & 28.00 & 14.34 & 23.85 & 22.56 & 26.90 & 28.26 \\
\rowcolor[HTML]{FFFFC7}
\textbf{Thoth-mini} & 48.05 & 79.19 & 32.00 & 76.34 & 60.34 & 44.60 & 52.66 & 49.67 & 57.45 & 55.59 \\
\textbf{Qwen3-8B} & 30.02 & 63.84 & 12.00 & 54.01 & 38.00 & 21.65 & 38.76 & 41.38 & 46.75 & 38.49 \\
\rowcolor[HTML]{FFFFC7}
\textbf{Thoth} & 50.64 & \textbf{79.48} & 27.98 & 77.66 & \textbf{63.66} & \textbf{46.68} & \textbf{54.26} & \textbf{52.95} & \textbf{61.79} & \textbf{57.23} \\ \bottomrule
\end{tabular}}
\end{table}

\renewcommand{\arraystretch}{1.1}
\begin{table}[h]
\centering
\caption{Main results on SciRecipe-Eval (Troubleshooting). Metrics left of the dashed line evaluate executability, those on the right measure semantic similarity. \textbf{Bold} denotes the best score.}
\label{TABLE15}
\resizebox{\textwidth}{!}{%
\begin{tabular}{@{}lccccc:ccccc@{}}
\toprule
\textbf{Methods} &
  \textbf{Semantic-A} &
  \textbf{Order-LCS} &
  \textbf{Order-S} &
  \textbf{Order-Tau} &
  \textbf{Step-M} &
  \textbf{BLUE-AVG} &
  \textbf{ROUGE-L} &
  \textbf{METEOR} &
  \textbf{KW-F1} &
  \textbf{AVG} \\ \midrule
\multicolumn{11}{l}{\textit{Close-Source SOTA}} \\
\textbf{ChatGPT-4o} & 24.46 & 62.76 & 10.00 & 53.99 & 26.00 & 32.80 & 41.80 & 40.10 & 47.80 & 37.75 \\
\textbf{GPT-5} & 17.86 & 35.51 & 0.30 & 16.24 & 1.82 & 10.31 & 20.29 & 26.51 & 29.46 & 17.55 \\
\textbf{GPT-5 Chat} & 26.45 & 65.98 & 8.00 & 46.00 & 16.00 & 28.80 & 39.98 & 38.21 & 46.97 & 35.15 \\
\textbf{Claude Sonnet 4} & 24.92 & 63.21 & 3.98 & 52.00 & 9.99 & 23.83 & 37.34 & 38.26 & 45.72 & 33.25 \\
\textbf{Claude Opus 4.1} & 20.24 & 54.78 & 0.35 & 56.48 & -0.12 & 16.77 & 30.78 & 34.10 & 40.49 & 28.21 \\
\textbf{Gemini 2.5 Flash} & 26.76 & 59.99 & 6.00 & 58.99 & 31.98 & 27.13 & 36.19 & 36.58 & 45.68 & 36.59 \\
\textbf{Gemini 2.5 Pro} & 22.05 & 61.06 & 10.00 & 52.00 & 16.00 & 25.36 & 37.48 & 40.39 & 45.48 & 34.42 \\
\textbf{Doubao-1.5-pro} & 22.53 & 67.37 & \textbf{20.00} & 55.00 & 50.00 & 33.61 & 40.67 & 35.77 & 43.73 & 40.96 \\
\textbf{Qwen2.5-Max} & 22.66 & 62.97 & 8.00 & 55.97 & \textbf{56.00} & 29.08 & 40.45 & 39.97 & 47.33 & 40.27 \\ \midrule
\multicolumn{11}{l}{\textit{Open-Source SOTA}} \\
\textbf{Qwen2.5-72B-Instruct} & 18.12 & 61.45 & 8.00 & 56.00 & 34.00 & 25.78 & 39.30 & 37.42 & 44.15 & 36.02 \\
\textbf{Qwen3-235B-A22B-Instruct-2507} & 26.43 & 66.89 & 11.97 & 62.00 & 19.97 & 26.47 & 38.58 & 40.38 & 43.66 & 37.37 \\
\textbf{DeepSeek-V3} & 30.92 & 65.57 & 5.97 & 47.98 & 32.00 & 33.12 & 43.59 & 42.50 & 49.35 & 39.00 \\
\textbf{GPT-OSS-120B} & 26.48 & 57.03 & 0.00 & 46.00 & 0.00 & 22.57 & 35.45 & 36.83 & 46.09 & 30.05 \\
\textbf{Llama-3.1-405B-Instruct} & 18.65 & 62.63 & 5.97 & 52.00 & 29.97 & 28.69 & 36.64 & 37.14 & 43.03 & 34.97 \\
\textbf{Kimi-K2-Instruction} & 23.10 & 60.74 & 1.98 & 53.98 & 16.00 & 24.85 & 36.66 & 39.49 & 44.33 & 33.46 \\ \midrule
\multicolumn{11}{l}{\textit{Reasoning Models}} \\
\textbf{DeepSeek-R1} & 23.93 & 62.64 & 7.97 & 54.97 & 20.00 & 31.06 & 39.30 & 37.12 & 47.28 & 36.03 \\
\textbf{Grok 3} & 26.57 & 68.38 & 9.97 & 64.20 & 31.28 & 29.55 & 40.59 & 42.84 & \textbf{42.84} & 39.90 \\
\textbf{Grok 4} & 19.08 & 61.60 & 5.30 & 35.62 & 19.26 & 27.40 & 36.75 & 35.81 & 44.64 & 31.72 \\
\textbf{OpenAI o1} & 18.78 & 66.89 & 4.00 & 48.00 & 16.00 & 29.11 & 40.40 & 40.04 & 46.53 & 34.42 \\
\textbf{OpenAI o3} & 24.50 & 54.81 & -0.08 & 47.97 & 1.94 & 20.76 & 34.55 & 38.84 & 45.86 & 29.91 \\ \midrule
\textbf{Qwen3-4B} & 17.53 & 52.83 & 8.00 & 40.00 & 22.00 & 16.22 & 24.78 & 24.04 & 29.37 & 26.09 \\
\rowcolor[HTML]{FFFFC7}
\textbf{Thoth-mini} & \textbf{35.15} & 67.81 & 11.98 & 58.31 & 50.33 & \textbf{40.65} & \textbf{44.95} & 41.82 & \textbf{52.04} & 44.78 \\
\textbf{Qwen3-8B} & 18.46 & 57.11 & 6.00 & 46.00 & 25.98 & 16.00 & 29.18 & 29.89 & 35.00 & 29.29 \\
\rowcolor[HTML]{FFFFC7}
\textbf{Thoth} & 29.89 & \textbf{70.37} & 13.99 & \textbf{75.65} & 53.66 & 40.53 & 44.68 & 42.02 & 51.19 & \textbf{46.89} \\ \bottomrule
\end{tabular}}
\end{table}

\renewcommand{\arraystretch}{1.1}
\begin{table}[h]
\centering
\caption{Main results on SciRecipe-Eval (Constraint). Metrics left of the dashed line evaluate executability, those on the right measure semantic similarity. \textbf{Bold} denotes the best score.}
\label{TABLE16}
\resizebox{\textwidth}{!}{%
\begin{tabular}{@{}lccccc:ccccc@{}}
\toprule
\textbf{Methods} &
  \textbf{Semantic-A} &
  \textbf{Order-LCS} &
  \textbf{Order-S} &
  \textbf{Order-Tau} &
  \textbf{Step-M} &
  \textbf{BLUE-AVG} &
  \textbf{ROUGE-L} &
  \textbf{METEOR} &
  \textbf{KW-F1} &
  \textbf{AVG} \\ \midrule
\multicolumn{11}{l}{\textit{Close-Source SOTA}} \\
\textbf{ChatGPT-4o} & 40.34 & 74.56 & 30.00 & 73.99 & 53.98 & 39.94 & 50.87 & 48.34 & 55.32 & 51.93 \\
\textbf{GPT-5} & 19.47 & 61.94 & 6.35 & 66.20 & 12.59 & 18.55 & 33.55 & 35.68 & 42.92 & 33.03 \\
\textbf{GPT-5 Chat} & 33.40 & \textbf{78.43} & 30.00 & 72.00 & 44.00 & 37.21 & 50.27 & 47.02 & 54.26 & 49.62 \\
\textbf{Claude Sonnet 4} & 48.20 & 73.30 & 17.97 & \textbf{82.00} & 33.97 & 32.52 & 46.00 & 41.46 & 53.92 & 47.70 \\
\textbf{Claude Opus 4.1} & 40.64 & 70.68 & 19.38 & 80.29 & 42.74 & 29.59 & 43.40 & 43.25 & 51.95 & 46.88 \\
\textbf{Gemini 2.5 Flash} & 36.61 & 68.91 & 11.99 & 68.97 & 31.97 & 33.88 & 44.40 & 39.63 & 50.71 & 43.01 \\
\textbf{Gemini 2.5 Pro} & 32.36 & 74.58 & 25.97 & 74.00 & 33.99 & 33.30 & 47.75 & 46.43 & 51.12 & 46.61 \\
\textbf{Doubao-1.5-pro} & 32.72 & 74.37 & 26.00 & 73.00 & 41.99 & 37.64 & 49.48 & 39.95 & 51.36 & 47.39 \\
\textbf{Qwen2.5-Max} & 40.38 & 76.12 & \textbf{31.97} & 77.97 & 53.99 & 32.62 & 50.31 & 47.20 & 55.84 & 51.82 \\ \midrule
\multicolumn{11}{l}{\textit{Open-Source SOTA}} \\
\textbf{Qwen2.5-72B-Instruct} & 33.26 & 67.67 & 20.00 & 64.00 & 46.00 & 31.00 & 46.22 & 43.65 & 52.61 & 44.93 \\
\textbf{Qwen3-235B-A22B-Instruct-2507} & 35.41 & 71.54 & 19.97 & 70.00 & 39.97 & 32.22 & 45.24 & 46.91 & 51.63 & 45.88 \\
\textbf{DeepSeek-V3} & 35.01 & 72.93 & 17.97 & 75.98 & 35.99 & 38.49 & 49.54 & 46.21 & 55.19 & 47.48 \\
\textbf{GPT-OSS-120B} & 18.19 & 70.31 & 12.00 & 62.00 & 23.97 & 29.09 & 43.66 & 44.23 & 51.98 & 39.49 \\
\textbf{Llama-3.1-405B-Instruct} & 36.13 & 72.50 & 21.97 & 74.00 & 55.97 & 38.27 & 48.68 & 47.82 & 51.67 & 49.67 \\
\textbf{Kimi-K2-Instruction} & 35.50 & 74.28 & 21.97 & 71.97 & 31.98 & 30.80 & 45.14 & 44.00 & 51.35 & 45.22 \\ \midrule
\multicolumn{11}{l}{\textit{Reasoning Models}} \\
\textbf{DeepSeek-R1} & 38.65 & 68.66 & 15.97 & 68.97 & 28.00 & 37.76 & 47.20 & 41.49 & 55.16 & 44.65 \\
\textbf{Grok 3} & 30.66 & 72.92 & 24.84 & 70.64 & 49.11 & 33.61 & 48.02 & 50.49 & 52.35 & 48.07 \\
\textbf{Grok 4} & 37.78 & 73.55 & 26.73 & 61.81 & 38.31 & 39.79 & 50.63 & 43.57 & 57.57 & 47.75 \\
\textbf{OpenAI o1} & 29.63 & 74.94 & 21.99 & 72.00 & 40.00 & 37.58 & 49.99 & 48.57 & 55.81 & 47.83 \\
\textbf{OpenAI o3} & 34.13 & 74.69 & 13.91 & 73.97 & 17.94 & 30.23 & 47.49 & 47.97 & 54.09 & 43.82 \\ \midrule
\textbf{Qwen3-4B} & 14.71 & 34.91 & 8.00 & 30.00 & 21.97 & 9.76 & 16.13 & 15.51 & 18.47 & 18.83 \\
\rowcolor[HTML]{FFFFC7}
\textbf{Thoth-mini} & 35.26 & 74.85 & 29.98 & 70.31 & \textbf{64.31} & 42.72 & 50.53 & 48.19 & 54.34 & 52.28 \\
\textbf{Qwen3-8B} & 25.59 & 57.32 & 12.00 & 52.00 & 31.97 & 19.66 & 34.50 & 34.38 & 42.67 & 34.45 \\
\rowcolor[HTML]{FFFFC7}
\textbf{Thoth} & \textbf{48.86} & 76.75 & 27.99 & 65.65 & 59.66 & \textbf{44.75} & \textbf{52.68} & \textbf{51.19} & \textbf{58.09} & \textbf{53.96} \\ \bottomrule
\end{tabular}}
\end{table}

\renewcommand{\arraystretch}{1.1}
\begin{table}[h]
\centering
\caption{Main results on SciRecipe-Eval (Scaling). Metrics left of the dashed line evaluate executability, those on the right measure semantic similarity. \textbf{Bold} denotes the best score.}
\label{TABLE17}
\resizebox{\textwidth}{!}{%
\begin{tabular}{@{}lccccc:ccccc@{}}
\toprule
\textbf{Methods} &
  \textbf{Semantic-A} &
  \textbf{Order-LCS} &
  \textbf{Order-S} &
  \textbf{Order-Tau} &
  \textbf{Step-M} &
  \textbf{BLUE-AVG} &
  \textbf{ROUGE-L} &
  \textbf{METEOR} &
  \textbf{KW-F1} &
  \textbf{AVG} \\ \midrule
\multicolumn{11}{l}{\textit{Close-Source SOTA}} \\
\textbf{ChatGPT-4o} & \textbf{74.95} & 77.87 & 38.00 & 71.99 & 50.00 & 51.36 & 62.12 & 58.19 & 69.93 & 61.60 \\
\textbf{GPT-5} & 51.76 & 70.97 & 31.97 & 72.24 & 43.83 & 34.89 & 49.73 & 44.19 & 66.47 & 51.78 \\
\textbf{GPT-5 Chat} & 62.62 & 80.70 & 44.00 & 70.00 & 50.00 & 45.85 & 59.93 & 57.32 & 70.79 & 60.13 \\
\textbf{Claude Sonnet 4} & 63.92 & 78.50 & 32.00 & 72.00 & 46.00 & 44.63 & 56.48 & 52.89 & 69.78 & 57.36 \\
\textbf{Claude Opus 4.1} & 65.47 & 82.42 & 46.70 & 77.39 & 51.12 & 51.38 & 63.26 & 58.90 & 73.30 & 63.33 \\
\textbf{Gemini 2.5 Flash} & 53.67 & 77.64 & 36.00 & 71.00 & 40.00 & 39.74 & 54.01 & 49.62 & 64.62 & 54.03 \\
\textbf{Gemini 2.5 Pro} & 56.34 & 78.80 & 32.00 & 78.00 & 34.00 & 38.05 & 54.83 & 59.07 & 64.73 & 55.09 \\
\textbf{Doubao-1.5-pro} & 59.81 & 81.55 & 44.00 & 77.00 & 52.00 & 41.74 & 61.12 & 54.51 & 68.72 & 60.05 \\
\textbf{Qwen2.5-Max} & 65.70 & 81.38 & 40.00 & 82.00 & 50.00 & 37.65 & 61.83 & 55.95 & 70.13 & 60.52 \\ \midrule
\multicolumn{11}{l}{\textit{Open-Source SOTA}} \\
\textbf{Qwen2.5-72B-Instruct} & 62.21 & 79.42 & 34.00 & 76.00 & 46.00 & 35.97 & 58.81 & 56.04 & 68.89 & 57.48 \\
\textbf{Qwen3-235B-A22B-Instruct-2507} & 58.53 & 79.31 & 33.97 & 74.00 & 41.97 & 45.93 & 57.84 & 53.45 & 67.16 & 56.91 \\
\textbf{DeepSeek-V3} & 65.90 & 83.85 & 43.97 & 75.98 & 60.00 & 50.00 & 63.04 & 58.98 & 71.61 & 63.70 \\
\textbf{GPT-OSS-120B} & 47.77 & 79.96 & 34.00 & 70.00 & 50.00 & 44.23 & 56.22 & 54.42 & 69.45 & 56.23 \\
\textbf{Llama-3.1-405B-Instruct} & 58.56 & 74.66 & 25.97 & 72.00 & 37.97 & 43.69 & 56.50 & 52.97 & 66.62 & 54.33 \\
\textbf{Kimi-K2-Instruction} & 62.16 & 76.45 & 38.00 & 70.00 & 52.00 & 44.62 & 59.76 & 53.93 & 68.27 & 58.35 \\ \midrule
\multicolumn{11}{l}{\textit{Reasoning Models}} \\
\textbf{DeepSeek-R1} & 50.18 & 78.14 & 30.58 & 78.60 & 32.65 & 44.83 & 55.15 & 49.54 & 64.70 & 53.82 \\
\textbf{Grok 3} & 67.04 & 78.41 & 27.72 & 72.64 & 36.69 & 46.19 & 59.32 & 56.66 & 69.14 & 57.09 \\
\textbf{Grok 4} & 48.13 & 79.31 & 37.75 & 78.98 & 39.74 & 46.50 & 57.48 & 50.66 & 67.51 & 56.23 \\
\textbf{OpenAI o1} & 60.64 & 77.34 & 28.00 & 70.00 & 44.00 & 46.62 & 58.80 & 52.93 & 69.43 & 56.42 \\
\textbf{OpenAI o3} & 65.13 & 80.59 & 35.94 & 72.00 & 41.94 & 42.92 & 60.70 & 54.73 & 71.65 & 58.40 \\ \midrule
\textbf{Qwen3-4B} & 29.58 & 40.68 & 18.00 & 40.00 & 26.00 & 16.51 & 25.92 & 22.42 & 30.26 & 27.71 \\
\rowcolor[HTML]{FFFFC7}
\textbf{Thoth-mini} & 68.87 & 85.46 & 47.98 & 88.31 & 68.34 & 54.25 & 63.90 & 62.71 & 72.65 & 68.05 \\
\textbf{Qwen3-8B} & 36.87 & 60.84 & 18.00 & 60.00 & 26.00 & 20.87 & 41.02 & 40.61 & 51.26 & 39.50 \\
\rowcolor[HTML]{FFFFC7}
\textbf{Thoth} & 74.83 & \textbf{87.69} & \textbf{53.99} & \textbf{93.65} & \textbf{69.66} & \textbf{55.02} & \textbf{66.14} & \textbf{63.61} & \textbf{74.15} & \textbf{70.97} \\ \bottomrule
\end{tabular}}
\end{table}

\renewcommand{\arraystretch}{1.1}
\begin{table}[h]
\centering
\caption{Main results on SciRecipe-Eval (Safety). Metrics left of the dashed line evaluate executability, those on the right measure semantic similarity. \textbf{Bold} denotes the best score.}
\label{TABLE18}
\resizebox{\textwidth}{!}{%
\begin{tabular}{@{}lccccc:ccccc@{}}
\toprule
\textbf{Methods} &
  \textbf{Semantic-A} &
  \textbf{Order-LCS} &
  \textbf{Order-S} &
  \textbf{Order-Tau} &
  \textbf{Step-M} &
  \textbf{BLUE-AVG} &
  \textbf{ROUGE-L} &
  \textbf{METEOR} &
  \textbf{KW-F1} &
  \textbf{AVG} \\ \midrule
\multicolumn{11}{l}{\textit{Close-Source SOTA}} \\
\textbf{ChatGPT-4o} & 46.78 & 76.08 & 22.00 & 71.99 & 40.00 & 42.88 & \textbf{52.51} & 50.23 & \textbf{61.04} & 51.50 \\
\textbf{GPT-5} & 36.69 & 57.09 & 3.97 & 56.24 & 5.85 & 17.75 & 33.04 & 40.63 & 45.77 & 33.00 \\
\textbf{GPT-5 Chat} & 47.42 & 72.72 & 18.00 & 68.00 & 22.00 & 32.27 & 45.69 & 51.20 & 55.65 & 45.88 \\
\textbf{Claude Sonnet 4} & 44.59 & 77.33 & 24.00 & 78.00 & 52.00 & 36.76 & 49.68 & 50.35 & 57.72 & 52.27 \\
\textbf{Claude Opus 4.1} & \textbf{53.34} & \textbf{77.43} & \textbf{27.63} & \textbf{85.70} & 40.82 & 34.47 & 48.06 & 43.74 & 55.58 & 51.86 \\
\textbf{Gemini 2.5 Flash} & 33.47 & 72.38 & 26.00 & 77.00 & 38.00 & 34.50 & 46.52 & 47.71 & 53.76 & 47.70 \\
\textbf{Gemini 2.5 Pro} & 38.44 & 72.67 & 22.00 & 66.00 & 38.00 & 37.95 & 49.69 & \textbf{52.26} & 57.53 & 48.28 \\
\textbf{Doubao-1.5-pro} & 33.55 & 73.96 & 22.00 & 71.00 & 50.00 & 38.65 & 49.39 & 44.12 & 54.94 & 48.62 \\
\textbf{Qwen2.5-Max} & 46.16 & 71.19 & 14.00 & 76.00 & 50.00 & 35.55 & 51.71 & 49.62 & 60.03 & 50.47 \\ \midrule
\multicolumn{11}{l}{\textit{Open-Source SOTA}} \\
\textbf{Qwen2.5-72B-Instruct} & 32.86 & 72.62 & 22.00 & 72.00 & 48.00 & 33.76 & 49.18 & 46.46 & 54.86 & 47.97 \\
\textbf{Qwen3-235B-A22B-Instruct-2507} & 37.87 & 73.96 & 17.97 & 78.00 & 47.97 & 35.26 & 48.16 & 51.10 & 54.38 & 49.41 \\
\textbf{DeepSeek-V3} & 46.21 & 77.12 & 23.97 & 80.00 & 44.00 & 42.43 & 52.35 & 49.57 & 60.15 & \textbf{52.87} \\
\textbf{GPT-OSS-120B} & 36.72 & 71.92 & 12.00 & 66.00 & 18.00 & 33.16 & 46.79 & 47.61 & 55.11 & 43.03 \\
\textbf{Llama-3.1-405B-Instruct} & 38.23 & 71.94 & 17.97 & 76.00 & 43.97 & 37.17 & 47.53 & 47.51 & 53.52 & 48.20 \\
\textbf{Kimi-K2-Instruction} & 42.05 & 74.65 & 26.00 & 72.00 & \textbf{54.00} & 36.86 & 48.65 & 49.53 & 55.30 & 51.00 \\ \midrule
\multicolumn{11}{l}{\textit{Reasoning Models}} \\
\textbf{DeepSeek-R1} & 40.79 & 74.12 & 21.97 & 62.97 & 34.00 & 40.83 & 50.03 & 42.26 & 55.53 & 46.94 \\
\textbf{Grok 3} & 36.51 & 75.50 & 21.50 & 81.49 & \textbf{52.96} & 37.65 & 49.66 & 50.45 & 55.99 & 51.30 \\
\textbf{Grok 4} & 42.15 & 74.32 & 22.49 & 73.08 & 41.69 & 38.51 & 48.18 & 41.98 & 55.22 & 48.62 \\
\textbf{OpenAI o1} & 43.03 & 74.45 & 24.00 & 78.00 & 44.00 & 39.85 & 50.20 & 48.51 & 56.41 & 50.94 \\
\textbf{OpenAI o3} & 42.60 & 70.17 & 13.94 & 68.00 & 25.94 & 31.53 & 45.75 & 49.49 & 56.15 & 44.84 \\ \midrule
\textbf{Qwen3-4B} & 26.45 & 55.88 & 16.00 & 50.00 & 30.00 & 15.56 & 25.76 & 25.40 & 31.22 & 30.70 \\
\rowcolor[HTML]{FFFFC7}
\textbf{Thoth-mini} & 36.58 & 62.74 & 15.98 & 56.33 & 44.34 & 46.04 & 44.54 & 45.62 & 51.69 & 44.87 \\
\textbf{Qwen3-8B} & 26.54 & 62.01 & 12.00 & 66.00 & 24.00 & 19.86 & 32.45 & 33.62 & 39.52 & 35.11 \\
\rowcolor[HTML]{FFFFC7}
\textbf{Thoth} & 43.52 & 67.75 & 23.99 & 67.65 & 49.66 & \textbf{48.21} & 48.34 & 47.94 & 52.89 & 49.99 \\ \bottomrule
\end{tabular}}
\end{table}

\clearpage
\subsection{Other Results on Protocol Benchmark}
\label{Other Protocol Benchmark_add}
To further assess protocol comprehension, we evaluated Thoth on BioProBench sub-tasks, with results summarized in Table~\ref{TABLE2}. Consistent with the findings in protocol generation, the Thoth series achieved substantial gains over baseline models, with average improvements of 28.02\% and 20.41\% for Thoth and Thoth-mini, respectively. Scaling effects were also more evident in this setting, as Thoth outperformed Thoth-mini by 6.98\%, a gap larger than in protocol generation. On individual tasks, Thoth showed clear advantages in error correction and protocol QA, exceeding GPT-5 by 17.05\% and 12.09\%, respectively, and consistently outperforming other strong baselines such as Claude Sonnet 4 and Gemini 2.5 Flash. These results confirm that Thoth generalizes well across protocol-related tasks. Nevertheless, because BioProBench partially overlaps with SciRecipe and does not disclose the exact origins of its QA pairs, there remains a possibility of data leakage, and thus the reported results should be regarded as reference only.

\renewcommand{\arraystretch}{1.15}
\begin{table}[ht]
\centering
\caption{Results on the protocol QA tasks.}
\label{TABLE2}
\resizebox{0.6\textwidth}{!}{%
\begin{tabular}{@{}lccccc@{}}
\toprule
\multirow{2}{*}{\textbf{Methods}} & \multicolumn{2}{c}{\textbf{ERR}} & \multicolumn{2}{c}{\textbf{ORD}} & \textbf{PQA} \\
\cmidrule(lr){2-3}\cmidrule(lr){4-5}\cmidrule(lr){6-6}
 & \textbf{ACC} & \textbf{F1} & \textbf{EM} & \textbf{K\_tau} & \textbf{ACC} \\ \midrule
\textbf{ChatGPT-4o}                   & 59.33 & 36.95 & 41.89 & 68.05          & 60.75 \\
\textbf{GPT-5}                        & 67.72 & 59.65 & 48.86 & \textbf{73.44} & 70.58 \\
\textbf{GPT-5 Chat}                   & 60.33 & 41.38 & 45.66 & 71.71          & 60.75 \\
\textbf{Claude Sonnet 4}              & 62.92 & 54.64 & 47.71 & 72.31          & 67.08 \\
\textbf{Gemini 2.5 Flash}             & 62.17 & 51.91 & 43.79 & 73.12          & 64.47 \\
\textbf{Qwen2.5-Max}                  & 54.83 & 17.38 & 41.51 & 67.96          & 62.25 \\
\textbf{Doubao-1.5-pro}               & 56.43 & 24.49 & 43.38 & 70.86          & 63.00 \\
\textbf{Qwen2.5-72B-Instruct}         & 56.08 & 25.25 & 36.99 & 62.46          & 60.42 \\
\textbf{Qwen3-235B-A22B-Instruct}     & 59.33 & 39.15 & 41.63 & 69.18          & 62.58 \\
\textbf{DeepSeek-V3}                  & 57.33 & 30.25 & 40.83 & 70.56          & 62.83 \\
\textbf{GPT-OSS-120B}                 & 64.33 & 53.17 & 39.88 & 68.08          & 64.83 \\
\textbf{Llama-3.1-405B-Instruct}      & 57.30 & 39.33 & 39.06 & 65.42          & 57.67 \\
\textbf{Kimi-K2-Instruction}          & 59.50 & 37.05 & 42.62 & 71.70          & 63.00 \\ \midrule
\textbf{Qwen3-4B}                     & 58.00 & 48.78 & 22.53 & 51.82          & 47.08 \\
\rowcolor[HTML]{FFFFC7} 
\textbf{Thoth-mini}                   & 75.83 & 72.49 & 41.72 & 65.90          & 74.33 \\
\textbf{Qwen3-8B}                     & 56.83 & 32.90 & 28.63 & 56.12          & 50.58 \\
\rowcolor[HTML]{FFFFC7} 
\textbf{Thoth}                        & \textbf{80.75}  & \textbf{80.73} & \textbf{49.08} & 71.92 & \textbf{82.67} \\
\bottomrule
\end{tabular}%
}
\end{table}

\section{Futher Analyses}
\label{Futher Analysis_add}

\subsection{Reward Analyses \& Results}
\label{Defination_app}
\subsubsection{RL Algorithms}  
\label{RL Algorithms}
We conducted ablation experiments to assess the robustness of the SCORE mechanism under different RL algorithms, replacing the third-stage optimization algorithm while keeping earlier training stages consistent. As shown in Table~\ref{TABLE20}, SCORE remained stable across a variety of methods, confirming its reliability as a reward mechanism. Executability-oriented metrics revealed that GRPO offered the best overall balance, benefiting from group-wise normalization that stabilized training and strengthened action fidelity. This supports our choice of GRPO as the primary optimization method in Thoth.

\renewcommand{\arraystretch}{1.1}
\begin{table}[ht]
\centering
\caption{Further analysis results on RL algorithms.}
\label{TABLE20}

\resizebox{0.85\textwidth}{!}{%
\begin{tabular}{c}

\begin{subtable}{0.95\linewidth}
\centering
\caption{}
\begin{tabularx}{\linewidth}{@{}l C C C C C@{}}
\toprule
\textbf{Algorithm} & \textbf{BERTScore-F1} & \textbf{BERTScore-P} & \textbf{BERTScore-R} & \textbf{BLEU-1} & \textbf{BLEU-2} \\
\midrule
\textbf{GPG} & 97.81 & 98.01 & 97.62 & 68.56 & 47.30 \\
\textbf{GRPO\_Dr} & 97.81 & 97.99 & 97.63 & 68.10 & 47.31 \\
\textbf{GSPO} & 97.81 & 97.98 & 97.65 & 68.65 & 47.36 \\
\textbf{\makecell[l]{REINFORCE\\++-baseline}} & 97.82 & 98.08 & 97.56 & 68.83 & 48.18 \\
\textbf{ReMax} & 97.81 & 98.05 & 97.57 & 67.93 & 47.87 \\
\textbf{RLOO} & 97.81 & 98.04 & 97.59 & 67.35 & 47.64 \\
\textbf{Thoth (GRPO)} & 97.82 & 97.99 & 97.65 & 68.56 & 48.28 \\
\bottomrule
\end{tabularx}
\end{subtable}
\\[1em]

\begin{subtable}{0.95\linewidth}
\centering
\caption{}
\begin{tabularx}{\linewidth}{@{}l C C C C C@{}}
\toprule
\textbf{Algorithm} & \textbf{BLEU-3} & \textbf{BLEU-4} & \textbf{BLEU-AVG} & \textbf{ROUGE-1} & \textbf{ROUGE-2} \\
\midrule
\textbf{GPG} & 34.44 & 21.80 & 43.03 & 59.98 & 36.58 \\
\textbf{GRPO\_Dr} & 34.44 & 21.85 & 42.93 & 60.31 & 36.89 \\
\textbf{GSPO} & 33.50 & 22.08 & 42.90 & 60.40 & 37.11 \\
\textbf{\makecell[l]{REINFORCE\\++-baseline}} & 34.71 & 22.69 & 43.60 & 59.92 & 37.08 \\
\textbf{ReMax} & 33.59 & 22.79 & 43.05 & 59.96 & 36.59 \\
\textbf{RLOO} & 33.60 & 22.61 & 42.80 & 60.41 & 37.15 \\
\textbf{Thoth (GRPO)} & 34.58 & 23.09 & 43.62 & 60.30 & 37.14 \\
\bottomrule
\end{tabularx}
\end{subtable}
\\[1em]

\begin{subtable}{0.95\linewidth}
\centering
\caption{}
\begin{tabularx}{\linewidth}{@{}l C C C C C@{}}
\toprule
\textbf{Algorithm} & \textbf{ROUGE-L} & \textbf{Meteor} & \textbf{KW-F1} & \textbf{KW-P} & \textbf{KW-R} \\
\midrule
\textbf{GPG} & 49.01 & 45.81 & 53.47 & 58.52 & 50.84 \\
\textbf{GRPO\_Dr} & 49.12 & 47.14 & 53.98 & 59.29 & 51.06 \\
\textbf{GSPO} & 49.32 & 46.91 & 54.20 & 58.59 & 51.98 \\
\textbf{\makecell[l]{REINFORCE\\++-baseline}} & 49.19 & 46.19 & 53.59 & 60.40 & 49.71 \\
\textbf{ReMax} & 49.52 & 45.95 & 53.47 & 60.27 & 49.56 \\
\textbf{RLOO} & 50.00 & 46.66 & 53.65 & 60.29 & 49.94 \\
\textbf{Thoth (GRPO)} & 50.02 & 47.39 & 54.13 & 58.82 & 51.70 \\
\bottomrule
\end{tabularx}
\end{subtable}
\\[1em]

\begin{subtable}{0.95\linewidth}
\centering
\caption{}
\begin{tabularx}{\linewidth}{@{}l C C C C C@{}}
\toprule
\textbf{Algorithm} & \textbf{Order-LCS} & \textbf{Order-S} & \textbf{Order-Tau} & \textbf{Semantic-A} & \textbf{Step-M} \\
\midrule
\textbf{GPG} & 74.02 & 22.50 & 70.33 & 45.07 & 46.67 \\
\textbf{GRPO\_Dr} & 75.34 & 26.67 & 71.50 & 44.11 & 50.50 \\
\textbf{GSPO} & 74.93 & 23.67 & 68.67 & 43.93 & 51.17 \\
\textbf{\makecell[l]{REINFORCE\\++-baseline}} & 75.69 & 25.50 & 70.67 & 44.14 & 50.67 \\
\textbf{ReMax} & 73.58 & 23.00 & 69.33 & 44.89 & 51.11 \\
\textbf{RLOO} & 74.42 & 25.00 & 70.17 & 43.87 & 51.50 \\
\textbf{Thoth (GRPO)} & 75.34 & 25.50 & 73.33 & 46.60 & 53.00 \\
\bottomrule
\end{tabularx}
\end{subtable}

\end{tabular}
} 

\end{table}

\subsubsection{Reward Computations}
\label{Reward Computations}
We further examined the impact of different reward computation strategies within the SCORE mechanism, considering the choice of order consistency logic (\textbf{L}CS or \textbf{S}trict), the method of combining order and semantic rewards (\textbf{P}roduct or \textbf{S}um), and the integration of step scale with step semantics (\textbf{P}roduct or \textbf{S}um). As summarized in Table~\ref{TABLE21}, strict order scoring consistently outperformed LCS, product-based integration proved more effective than summation, and multiplicative incorporation of step scale with semantics yielded stronger executability. These trends collectively support the adoption of the S-S-P configuration in Thoth, which achieved the best balance between surface similarity and procedural fidelity.

\renewcommand{\arraystretch}{1.1}
\begin{table}[th]
\centering
\caption{Further analysis results on Reward Computations.}
\label{TABLE21}

\resizebox{0.85\textwidth}{!}{%
\begin{tabular}{c}

\begin{subtable}{0.95\linewidth}
\centering
\caption{}
\begin{tabularx}{\linewidth}{@{}l C C C C C@{}}
\toprule
\textbf{Combine} & \textbf{BERTScore-F1} & \textbf{BERTScore-P} & \textbf{BERTScore-R} & \textbf{BLEU-1} & \textbf{BLEU-2} \\
\midrule
\textbf{L-P-P} & 97.76 & 97.85 & 97.67 & 63.17 & 44.25 \\
\textbf{L-P-S} & 97.73 & 97.80 & 97.66 & 62.56 & 43.74 \\
\textbf{L-S-P} & 97.78 & 97.87 & 97.69 & 63.23 & 44.37 \\
\textbf{L-S-S} & 97.59 & 97.66 & 97.51 & 61.91 & 43.03 \\
\textbf{S-P-P} & 97.80 & 98.00 & 97.61 & 70.04 & 49.18 \\
\textbf{S-P-S} & 97.80 & 97.91 & 97.70 & 64.18 & 45.14 \\
\textbf{S-S-S} & 97.80 & 97.92 & 97.68 & 67.28 & 47.15 \\
\textbf{Thoth (S-S-P)} & 97.82 & 97.99 & 97.65 & 68.56 & 48.28 \\
\bottomrule
\end{tabularx}
\end{subtable}
\\[1em]

\begin{subtable}{0.95\linewidth}
\centering
\caption{}
\begin{tabularx}{\linewidth}{@{}l C C C C C@{}}
\toprule
\textbf{Combine} & \textbf{BLEU-3} & \textbf{BLEU-4} & \textbf{BLEU-AVG} & \textbf{ROUGE-1} & \textbf{ROUGE-2} \\
\midrule
\textbf{L-P-P} & 31.31 & 20.74 & 39.87 & 58.81 & 35.40 \\
\textbf{L-P-S} & 30.95 & 20.30 & 39.39 & 58.86 & 35.31 \\
\textbf{L-S-P} & 31.32 & 20.54 & 39.86 & 59.24 & 35.82 \\
\textbf{L-S-S} & 30.36 & 19.79 & 38.77 & 58.49 & 34.62 \\
\textbf{S-P-P} & 35.18 & 23.44 & 44.46 & 59.52 & 36.48 \\
\textbf{S-P-S} & 32.20 & 21.34 & 40.71 & 59.33 & 36.05 \\
\textbf{S-S-S} & 33.73 & 22.49 & 42.66 & 60.08 & 36.56 \\
\textbf{Thoth (S-S-P)} & 34.58 & 23.09 & 43.62 & 60.30 & 37.14 \\
\bottomrule
\end{tabularx}
\end{subtable}
\\[1em]

\begin{subtable}{0.95\linewidth}
\centering
\caption{}
\begin{tabularx}{\linewidth}{@{}l C C C C C@{}}
\toprule
\textbf{Combine} & \textbf{ROUGE-L} & \textbf{Meteor} & \textbf{KW-F1} & \textbf{KW-P} & \textbf{KW-R} \\
\midrule
\textbf{L-P-P} & 48.17 & 45.93 & 53.21 & 55.78 & 52.72 \\
\textbf{L-P-S} & 47.23 & 46.11 & 52.67 & 54.75 & 52.47 \\
\textbf{L-S-P} & 48.09 & 46.08 & 53.50 & 56.30 & 52.78 \\
\textbf{L-S-S} & 46.85 & 45.63 & 52.83 & 54.64 & 53.05 \\
\textbf{S-P-P} & 49.40 & 45.57 & 53.35 & 59.23 & 50.02 \\
\textbf{S-P-S} & 48.47 & 46.37 & 53.62 & 56.23 & 53.10 \\
\textbf{S-S-S} & 48.47 & 46.54 & 53.56 & 57.36 & 51.65 \\
\textbf{Thoth (S-S-P)} & 50.02 & 47.39 & 54.13 & 58.82 & 51.70 \\
\bottomrule
\end{tabularx}
\end{subtable}
\\[1em]

\begin{subtable}{0.95\linewidth}
\centering
\caption{}
\begin{tabularx}{\linewidth}{@{}l C C C C C@{}}
\toprule
\textbf{Combine} & \textbf{Order-LCS} & \textbf{Order-S} & \textbf{Order-Tau} & \textbf{Semantic-A} & \textbf{Step-M} \\
\midrule
\textbf{L-P-P} & 71.30 & 19.50 & 65.50 & 45.02 & 32.33 \\
\textbf{L-P-S} & 69.70 & 16.17 & 64.00 & 45.65 & 34.50 \\
\textbf{L-S-P} & 71.82 & 20.83 & 66.67 & 45.18 & 35.33 \\
\textbf{L-S-S} & 71.52 & 18.00 & 65.00 & 45.47 & 33.83 \\
\textbf{S-P-P} & 75.21 & 25.17 & 71.50 & 44.78 & 50.00 \\
\textbf{S-P-S} & 72.57 & 21.00 & 66.50 & 45.23 & 34.50 \\
\textbf{S-S-S} & 74.61 & 23.67 & 69.17 & 45.41 & 50.00 \\
\textbf{Thoth (S-S-P)} & 75.34 & 25.50 & 73.33 & 46.60 & 53.00 \\
\bottomrule
\end{tabularx}
\end{subtable}

\end{tabular}
} 

\end{table}

\subsubsection{Reward Magnitudes}
\label{Reward Magnitudes}
To better understand the sensitivity of SCORE to the absolute scale of feedback, we examined different strategies for adjusting reward magnitude. Four configurations were compared: constant scaling, scaling up, symmetric shifting with equal positive and negative ranges, and the default Thoth setting. As shown in Table~\ref{TABLE22}, model performance remained highly stable across all strategies, with only minimal variations observed on both text-level and executability-oriented metrics. This confirms that the effectiveness of SCORE arises mainly from its structured reward design rather than the precise numerical range, and also suggests that it can be applied robustly to more complex open-ended tasks where reward scales may differ.
\renewcommand{\arraystretch}{1.1}
\begin{table}[t]
\centering
\caption{Further analysis results on Reward Magnitudes.}
\label{TABLE22}

\resizebox{0.85\textwidth}{!}{%
\begin{tabular}{c}

\begin{subtable}{0.95\linewidth}
\centering
\caption{}
\begin{tabularx}{\linewidth}{@{}l C C C C C@{}}
\toprule
\textbf{Reward\_Scale} & \textbf{BERTScore-F1} & \textbf{BERTScore-P} & \textbf{BERTScore-R} & \textbf{BLEU-1} & \textbf{BLEU-2} \\
\midrule
\textbf{Constant {[}0, 2.5{]}}   & 97.80 & 97.98 & 97.62 & 67.79 & 47.50 \\
\textbf{Scaling\_up {[}0, 5{]}}  & 97.78 & 97.92 & 97.64 & 67.74 & 47.27 \\
\textbf{Shift {[}-1.25, 1.25{]}} & 97.79 & 97.98 & 97.60 & 67.76 & 47.45 \\
\textbf{Thoth {[}0, 1{]}}        & 97.82 & 97.99 & 97.65 & 68.56 & 48.28 \\
\bottomrule
\end{tabularx}
\end{subtable}
\\[1em]

\begin{subtable}{0.95\linewidth}
\centering
\caption{}
\begin{tabularx}{\linewidth}{@{}l C C C C C@{}}
\toprule
\textbf{Reward\_Scale} & \textbf{BLEU-3} & \textbf{BLEU-4} & \textbf{BLEU-AVG} & \textbf{ROUGE-1} & \textbf{ROUGE-2} \\
\midrule
\textbf{Constant {[}0, 2.5{]}}   & 33.60 & 22.00 & 42.72 & 60.06 & 36.38 \\
\textbf{Scaling\_up {[}0, 5{]}}  & 33.63 & 22.31 & 42.74 & 59.97 & 36.24 \\
\textbf{Shift {[}-1.25, 1.25{]}} & 33.57 & 22.82 & 42.90 & 60.11 & 36.69 \\
\textbf{Thoth {[}0, 1{]}}        & 34.58 & 23.09 & 43.62 & 60.30 & 37.14 \\
\bottomrule
\end{tabularx}
\end{subtable}
\\[1em]

\begin{subtable}{0.95\linewidth}
\centering
\caption{}
\begin{tabularx}{\linewidth}{@{}l C C C C C@{}}
\toprule
\textbf{Reward\_Scale} & \textbf{ROUGE-L} & \textbf{Meteor} & \textbf{KW-F1} & \textbf{KW-P} & \textbf{KW-R} \\
\midrule
\textbf{Constant {[}0, 2.5{]}}   & 49.06 & 46.12 & 53.92 & 58.80 & 51.30 \\
\textbf{Scaling\_up {[}0, 5{]}}  & 49.07 & 46.01 & 53.02 & 56.72 & 51.28 \\
\textbf{Shift {[}-1.25, 1.25{]}} & 49.15 & 46.75 & 53.70 & 58.67 & 51.11 \\
\textbf{Thoth {[}0, 1{]}}        & 50.02 & 47.39 & 54.13 & 58.82 & 51.70 \\
\bottomrule
\end{tabularx}
\end{subtable}
\\[1em]

\begin{subtable}{0.95\linewidth}
\centering
\caption{}
\begin{tabularx}{\linewidth}{@{}l C C C C C@{}}
\toprule
\textbf{Reward\_Scale} & \textbf{Order-LCS} & \textbf{Order-S} & \textbf{Order-Tau} & \textbf{Semantic-A} & \textbf{Step-M} \\
\midrule
\textbf{Constant {[}0, 2.5{]}}   & 74.71 & 25.17 & 71.50 & 45.22 & 51.50 \\
\textbf{Scaling\_up {[}0, 5{]}}  & 73.92 & 26.17 & 72.33 & 45.14 & 51.67 \\
\textbf{Shift {[}-1.25, 1.25{]}} & 73.70 & 26.50 & 72.49 & 45.36 & 52.12 \\
\textbf{Thoth {[}0, 1{]}}        & 75.34 & 25.50 & 73.33 & 46.60 & 53.00 \\
\bottomrule
\end{tabularx}
\end{subtable}

\end{tabular}
} 

\end{table}

\clearpage
\subsection{Pre-training Analysis}
\label{pretrained}

\begin{table}[ht]
\caption{Performance differences on the MMLU dataset \citep{hendryckstest2021} before and after pretraining. To examine potential catastrophic forgetting of general scientific knowledge, we evaluate MMLU on the OpenCompass platform \citep{2023opencompass}. ``\_pt'' denotes the Thoth model after the pretraining stage only, and $\Delta$ indicates the performance gap between pretrained models and their base counterparts. The results clearly show that pretraining on a large collection of high-quality protocols not only preserves general capability but even yields improvements in certain disciplines.}
\label{TABLE23}
\resizebox{\textwidth}{!}{%
\begin{tabular}{@{}cc|ccc|ccc@{}}
\toprule
\textbf{Dataset} &
  \textbf{Version} &
  \textbf{Qwen3-4B} &
  \textbf{Thoth-mini\_pt} &
  \textbf{$\Delta$} &
  \textbf{Qwen3-8B} &
  \textbf{Thoth\_pt} &
  \textbf{$\Delta$} \\ \midrule
lukaemon\_mmlu\_abstract\_algebra            & 2db373 & 91.00 & 87.00 & -4.00 & 88.00 & 91.00 & 3.00  \\
lukaemon\_mmlu\_anatomy                      & 72183b & 70.37 & 71.85 & 1.48  & 75.56 & 75.56 & 0.00  \\
lukaemon\_mmlu\_astronomy                    & d3ee01 & 84.87 & 89.47 & 4.60  & 90.79 & 88.82 & -1.97 \\
lukaemon\_mmlu\_business\_ethics             & 1dec08 & 77.00 & 72.00 & -5.00 & 81.00 & 77.00 & -4.00 \\
lukaemon\_mmlu\_clinical\_knowledge          & cb3218 & 80.38 & 81.89 & 1.51  & 83.77 & 86.04 & 2.27  \\
lukaemon\_mmlu\_college\_biology             & caec7d & 90.28 & 88.89 & -1.39 & 95.14 & 94.44 & -0.70 \\
lukaemon\_mmlu\_college\_chemistry           & 520aa6 & 66.00 & 63.00 & -3.00 & 71.00 & 71.00 & 0.00  \\
lukaemon\_mmlu\_college\_computer\_science   & 99c216 & 80.00 & 83.00 & 3.00  & 87.00 & 83.00 & -4.00 \\
lukaemon\_mmlu\_college\_mathematics         & 678751 & 90.00 & 86.00 & -4.00 & 77.00 & 79.00 & 2.00  \\
lukaemon\_mmlu\_college\_medicine            & 38709e & 83.24 & 84.97 & 1.73  & 84.97 & 84.97 & 0.00  \\
lukaemon\_mmlu\_college\_physics             & 4f382c & 94.12 & 94.12 & 0.00  & 95.10 & 95.10 & 0.00  \\
lukaemon\_mmlu\_computer\_security           & ce7550 & 85.00 & 82.00 & -3.00 & 84.00 & 86.00 & 2.00  \\
lukaemon\_mmlu\_conceptual\_physics          & 63588e & 88.94 & 88.09 & -0.85 & 94.04 & 93.62 & -0.42 \\
lukaemon\_mmlu\_econometrics                 & d1134d & 71.93 & 71.93 & 0.00  & 80.70 & 84.21 & 3.51  \\
lukaemon\_mmlu\_electrical\_engineering      & 770ce3 & 83.45 & 80.00 & -3.45 & 80.69 & 84.14 & 3.45  \\
lukaemon\_mmlu\_elementary\_mathematics      & 269926 & 97.35 & 97.35 & 0.00  & 97.62 & 97.88 & 0.26  \\
lukaemon\_mmlu\_formal\_logic                & cfcb0c & 92.86 & 88.89 & -3.97 & 89.68 & 90.48 & 0.80  \\
lukaemon\_mmlu\_global\_facts                & ab07b6 & 51.00 & 52.00 & 1.00  & 44.00 & 57.00 & 13.00 \\
lukaemon\_mmlu\_high\_school\_biology        & 37b125 & 85.81 & 87.74 & 1.93  & 90.00 & 93.87 & 3.87  \\
lukaemon\_mmlu\_high\_school\_chemistry      & ae8820 & 89.16 & 89.66 & 0.50  & 91.13 & 89.16 & -1.97 \\
lukaemon\_mmlu\_high\_school\_computer\_science &
  9965a5 &
  91.00 &
  92.00 &
  1.00 &
  98.00 &
  96.00 &
  -2.00 \\
lukaemon\_mmlu\_high\_school\_european\_history &
  eefc90 &
  82.42 &
  81.82 &
  -0.60 &
  87.27 &
  83.03 &
  -4.24 \\
lukaemon\_mmlu\_high\_school\_geography      & 0780e6 & 86.87 & 91.92 & 5.05  & 93.94 & 91.92 & -2.02 \\
lukaemon\_mmlu\_high\_school\_government\_and\_politics &
  3c52f9 &
  91.19 &
  90.67 &
  -0.52 &
  95.34 &
  95.34 &
  0.00 \\
lukaemon\_mmlu\_high\_school\_macroeconomics &
  a01685 &
  87.95 &
  86.92 &
  -1.03 &
  92.82 &
  91.54 &
  -1.28 \\
lukaemon\_mmlu\_high\_school\_mathematics    & ed4dc0 & 95.19 & 95.19 & 0.00  & 67.41 & 64.44 & -2.97 \\
lukaemon\_mmlu\_high\_school\_microeconomics & 04d21a & 93.70 & 93.70 & 0.00  & 96.22 & 95.80 & -0.42 \\
lukaemon\_mmlu\_high\_school\_physics        & 93278f & 86.75 & 86.75 & 0.00  & 90.07 & 86.75 & -3.32 \\
lukaemon\_mmlu\_high\_school\_psychology     & 7db114 & 91.93 & 92.84 & 0.91  & 95.05 & 94.31 & -0.74 \\
lukaemon\_mmlu\_high\_school\_statistics     & 8f3f3a & 89.81 & 89.35 & -0.46 & 91.67 & 90.74 & -0.93 \\
lukaemon\_mmlu\_high\_school\_us\_history    & 8932df & 85.78 & 87.75 & 1.97  & 90.20 & 88.24 & -1.96 \\
lukaemon\_mmlu\_high\_school\_world\_history & 048e7e & 85.65 & 82.70 & -2.95 & 88.61 & 89.03 & 0.42  \\
lukaemon\_mmlu\_human\_aging                 & 82a410 & 73.99 & 71.75 & -2.24 & 76.68 & 76.68 & 0.00  \\
lukaemon\_mmlu\_human\_sexuality             & 42407c & 83.97 & 81.68 & -2.29 & 89.31 & 85.50 & -3.81 \\
lukaemon\_mmlu\_international\_law           & cf3179 & 76.03 & 78.51 & 2.48  & 77.69 & 80.99 & 3.30  \\
lukaemon\_mmlu\_jurisprudence                & 001f24 & 75.00 & 79.63 & 4.63  & 87.96 & 80.56 & -7.40 \\
lukaemon\_mmlu\_logical\_fallacies           & 9cebb0 & 85.28 & 84.66 & -0.62 & 86.50 & 85.89 & -0.61 \\
lukaemon\_mmlu\_machine\_learning            & 0283bb & 76.79 & 76.79 & 0.00  & 78.57 & 82.14 & 3.57  \\
lukaemon\_mmlu\_management                   & 80876d & 84.47 & 85.44 & 0.97  & 88.35 & 86.41 & -1.94 \\
lukaemon\_mmlu\_marketing                    & 7394e3 & 91.88 & 91.45 & -0.43 & 94.87 & 91.45 & -3.42 \\
lukaemon\_mmlu\_medical\_genetics            & 881ef5 & 88.00 & 88.00 & 0.00  & 91.00 & 89.00 & -2.00 \\
lukaemon\_mmlu\_miscellaneous                & 935647 & 88.12 & 87.48 & -0.64 & 91.70 & 91.44 & -0.26 \\
lukaemon\_mmlu\_moral\_disputes              & a2173e & 74.28 & 72.83 & -1.45 & 72.25 & 73.99 & 1.74  \\
lukaemon\_mmlu\_moral\_scenarios             & f6dbe2 & 62.46 & 60.67 & -1.79 & 60.22 & 62.46 & 2.24  \\
lukaemon\_mmlu\_nutrition                    & 4543bd & 76.14 & 76.80 & 0.66  & 85.29 & 84.31 & -0.98 \\
lukaemon\_mmlu\_philosophy                   & 08042b & 72.03 & 68.81 & -3.22 & 76.53 & 79.10 & 2.57  \\
lukaemon\_mmlu\_prehistory                   & bbb197 & 81.48 & 80.25 & -1.23 & 85.80 & 84.88 & -0.92 \\
lukaemon\_mmlu\_professional\_accounting     & 444b7f & 73.40 & 73.40 & 0.00  & 81.21 & 80.85 & -0.36 \\
lukaemon\_mmlu\_professional\_law            & 5f7e6c & 49.80 & 50.07 & 0.27  & 55.61 & 56.39 & 0.78  \\
lukaemon\_mmlu\_professional\_medicine       & 857144 & 82.72 & 84.19 & 1.47  & 89.34 & 87.50 & -1.84 \\
lukaemon\_mmlu\_professional\_psychology     & 221a16 & 77.61 & 75.65 & -1.96 & 79.90 & 79.41 & -0.49 \\
lukaemon\_mmlu\_public\_relations            & e7d39b & 68.18 & 70.00 & 1.82  & 73.64 & 76.36 & 2.72  \\
lukaemon\_mmlu\_security\_studies            & 9b1743 & 71.84 & 75.51 & 3.67  & 76.33 & 77.96 & 1.63  \\
lukaemon\_mmlu\_sociology                    & c266a2 & 80.10 & 80.60 & 0.50  & 87.06 & 83.08 & -3.98 \\
lukaemon\_mmlu\_us\_foreign\_policy          & 528cfe & 86.00 & 86.00 & 0.00  & 87.00 & 85.00 & -2.00 \\
lukaemon\_mmlu\_virology                     & 12e270 & 55.42 & 51.81 & -3.61 & 54.82 & 57.23 & 2.41  \\
lukaemon\_mmlu\_world\_religions             & 232c09 & 84.21 & 85.38 & 1.17  & 85.96 & 88.30 & 2.34  \\ \midrule
AVG                                          &        & 81.41 & 81.21 & -0.20 & 83.88 & 83.79 & -0.09 \\ \bottomrule
\end{tabular}%
}
\end{table}


\clearpage
\section{Case Studies}
\label{Case Studies}
\begin{figure}[ht]
    \centering 
    \includegraphics[width=0.95\textwidth]{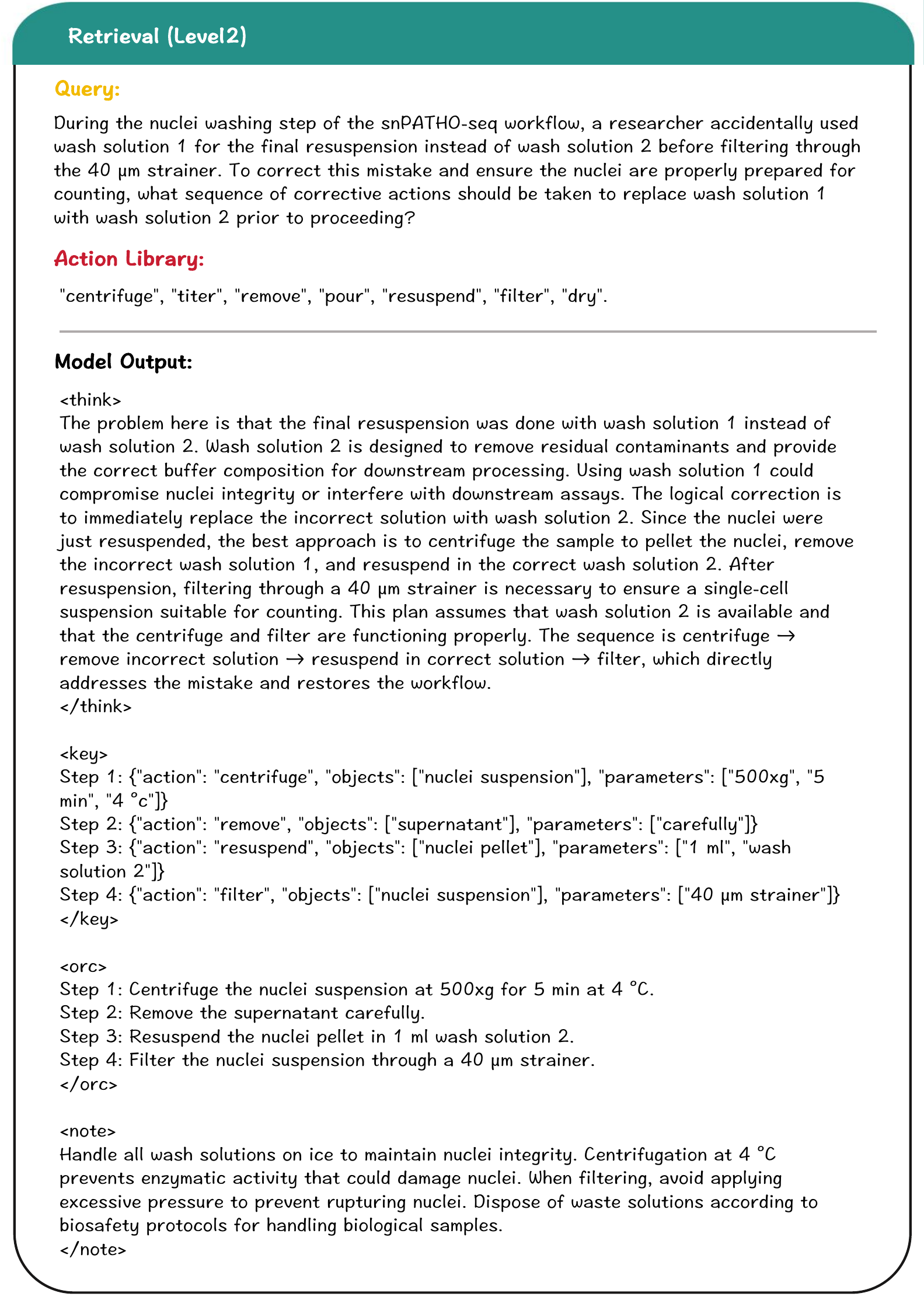} 
    \caption{Case result of Thoth on the Retrieval task.} 
    \label{fig11} 
\end{figure}

\begin{figure}[ht]
    \centering 
    \includegraphics[width=0.95\textwidth]{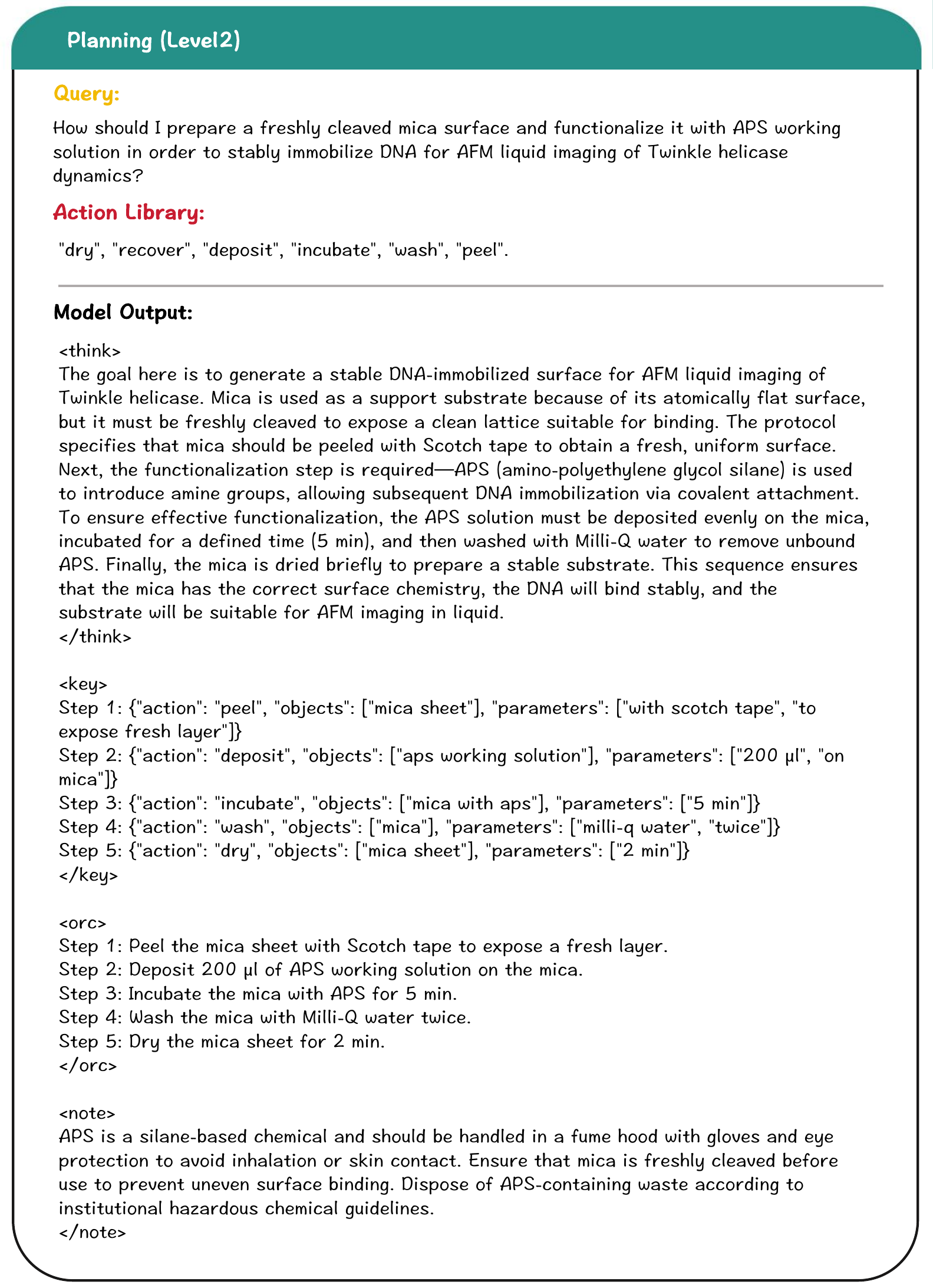} 
    \caption{Case result of Thoth on the Planning task.} 
    \label{fig12} 
\end{figure}

\begin{figure}[ht]
    \centering 
    \includegraphics[width=0.95\textwidth]{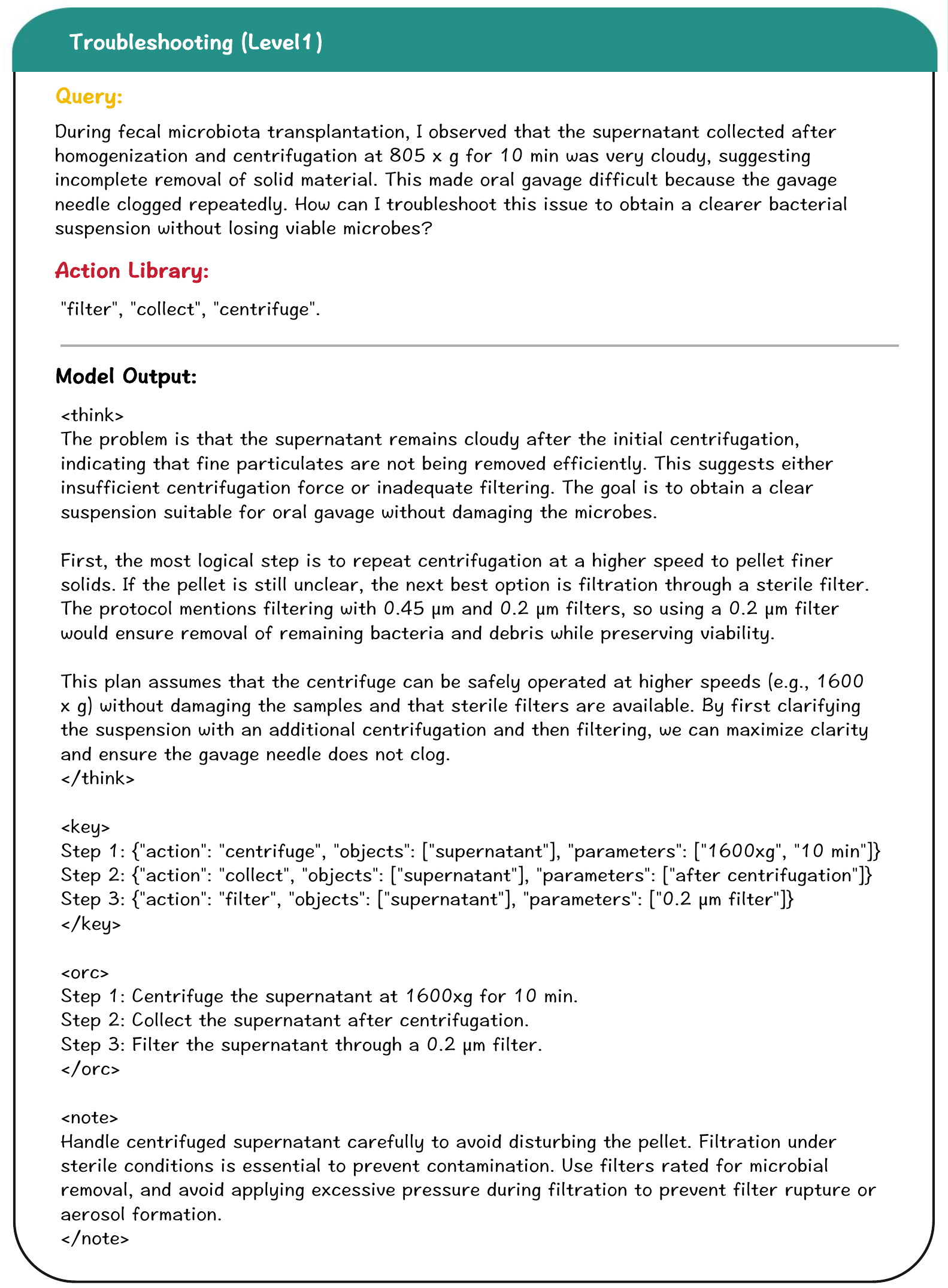} 
    \caption{Case result of Thoth on the Troubleshooting task.} 
    \label{fig13} 
\end{figure}

\begin{figure}[ht]
    \centering 
    \includegraphics[width=0.95\textwidth]{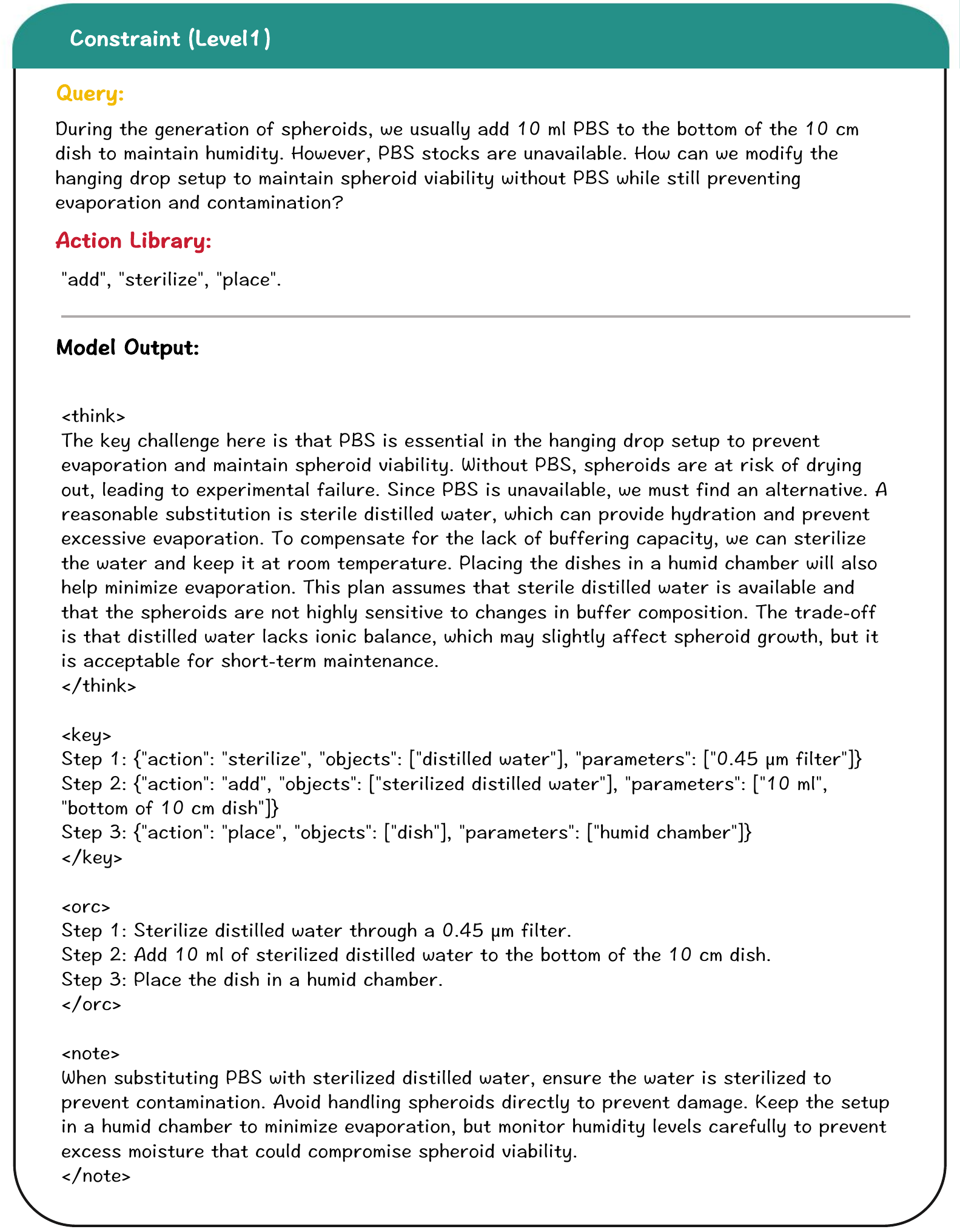} 
    \caption{Case result of Thoth on the Constraint task.} 
    \label{fig14} 
\end{figure}

\begin{figure}[ht]
    \centering 
    \includegraphics[width=0.95\textwidth]{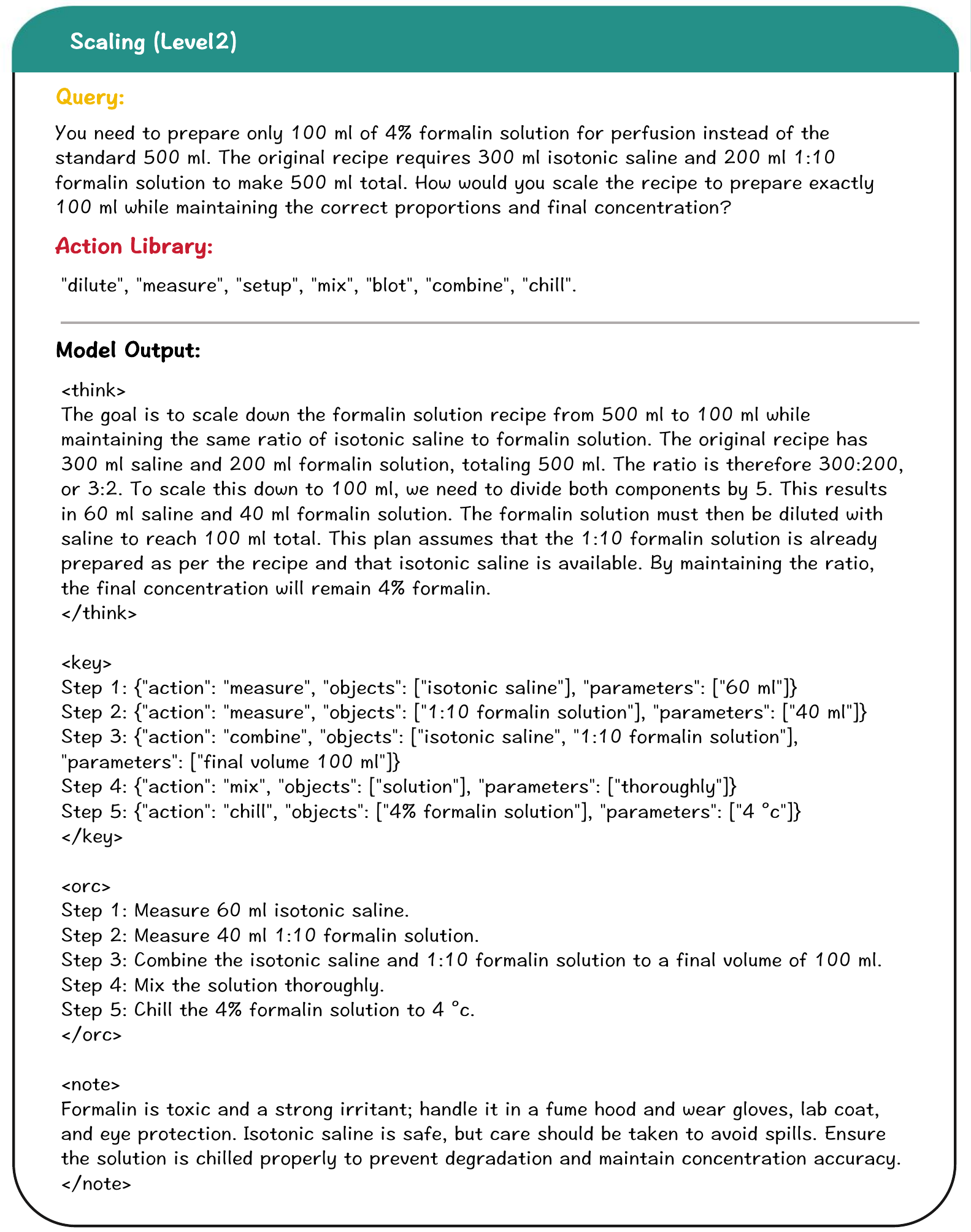} 
    \caption{Case result of Thoth on the Scaling task.} 
    \label{fig15} 
\end{figure}

\begin{figure}[ht]
    \centering 
    \includegraphics[width=0.95\textwidth]{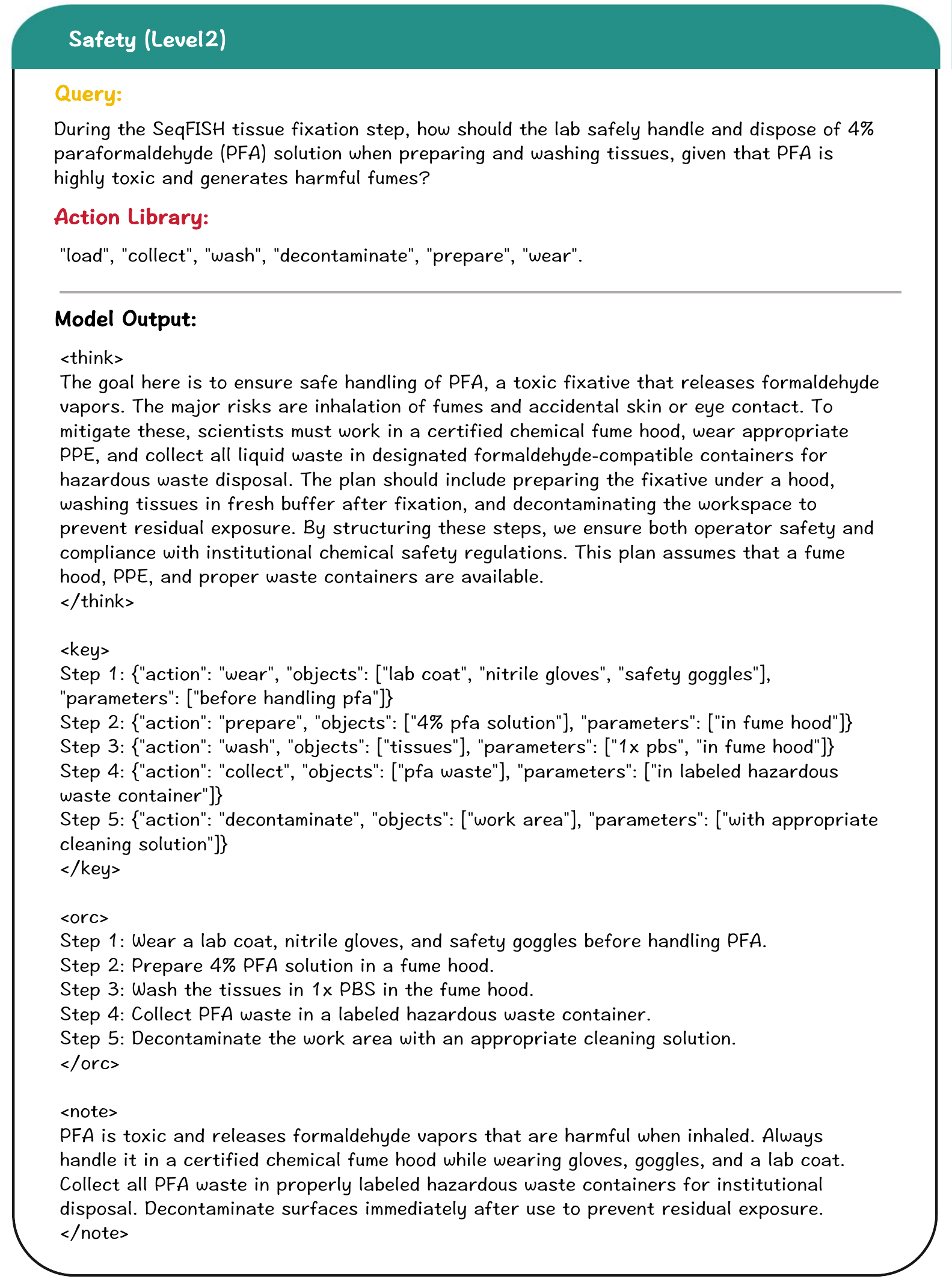} 
    \caption{Case result of Thoth on the Safety task.} 
    \label{fig16} 
\end{figure}

\clearpage
\section{Prompts}
\label{Prompts}
\begin{figure}[ht]
    \centering 
    \includegraphics[width=0.95\textwidth]{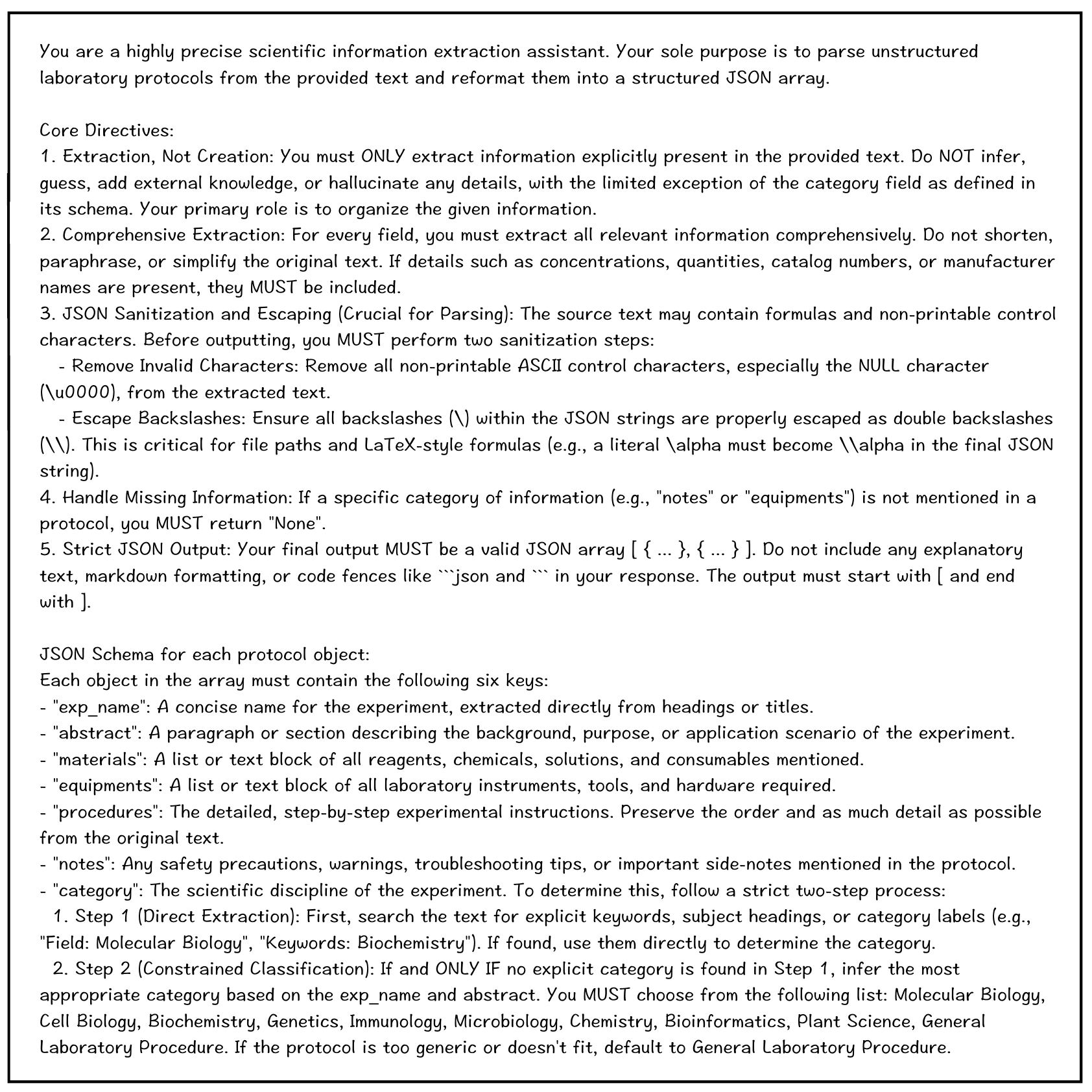} 
    \caption{System prompt 1 for structured integration with model\_based.} 
    \label{fig17} 
\end{figure}

\begin{figure}[ht]
    \centering 
    \includegraphics[width=0.95\textwidth]{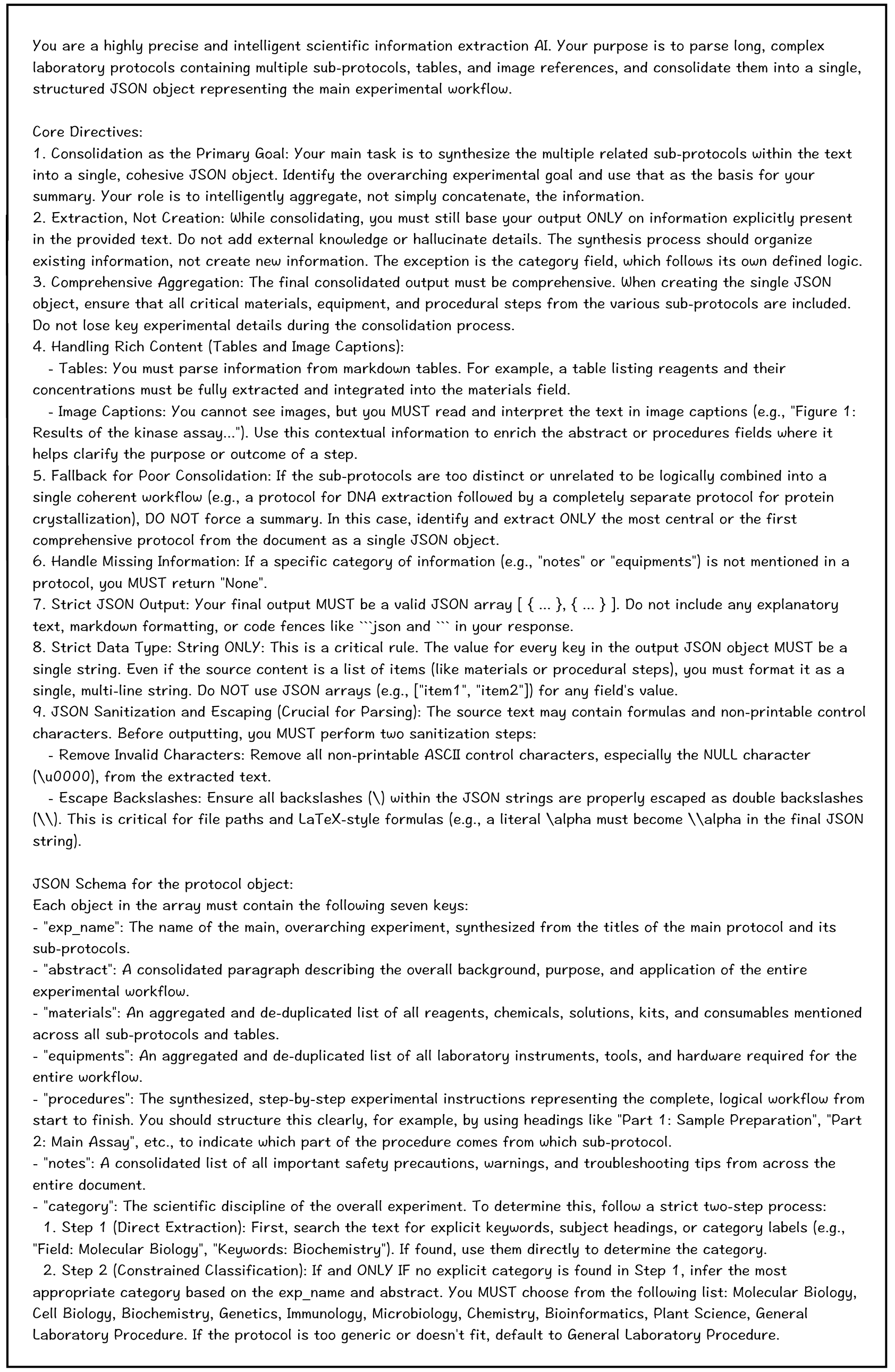} 
    \caption{System prompt 2 for structured integration with model\_based.} 
    \label{fig18} 
\end{figure}

\begin{figure}[ht]
    \centering 
    \includegraphics[width=0.95\textwidth]{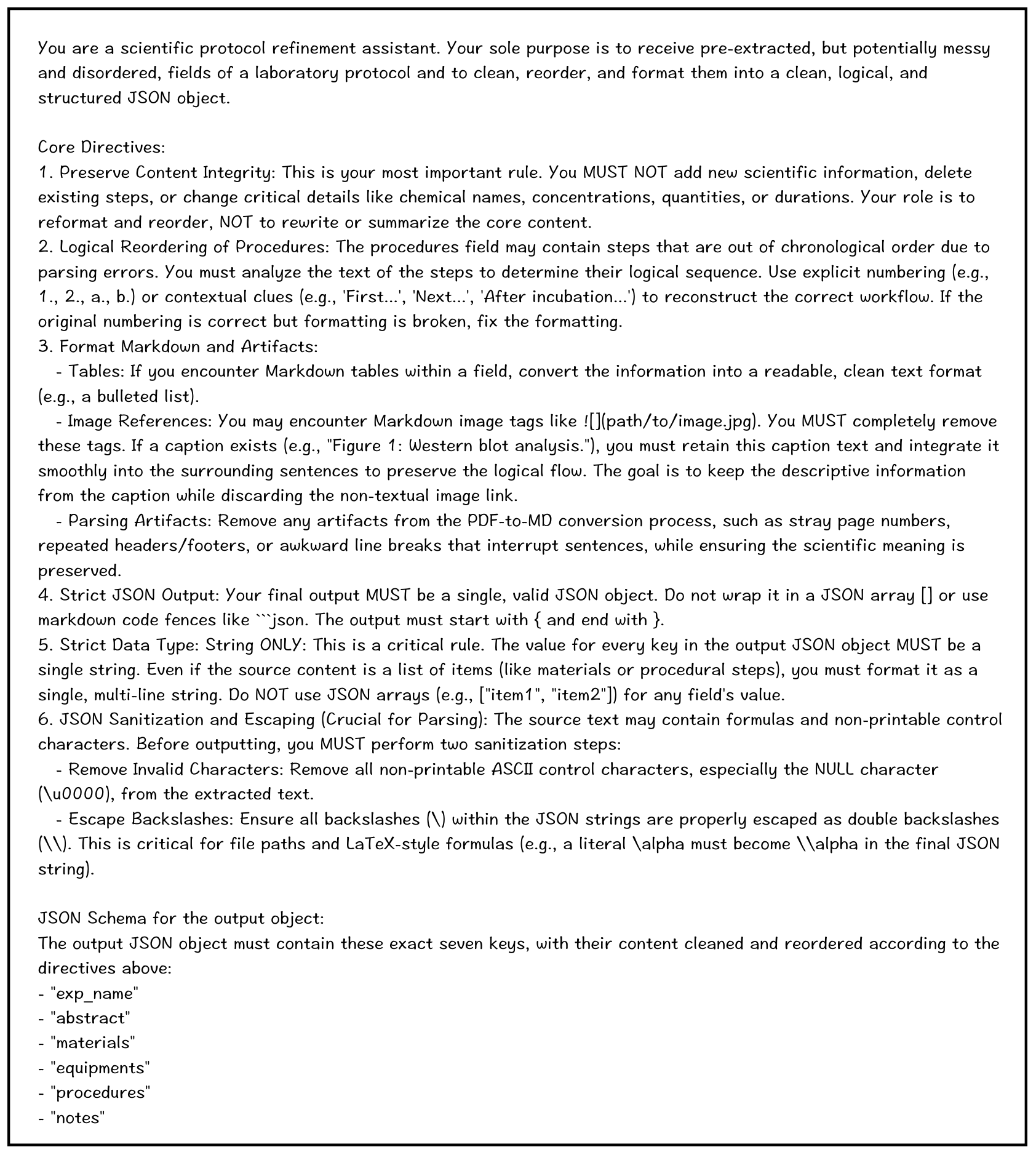} 
    \caption{System prompt 3 for structured integration with model\_based.} 
    \label{fig19} 
\end{figure}

\begin{figure}[ht]
    \centering 
    \includegraphics[width=0.95\textwidth]{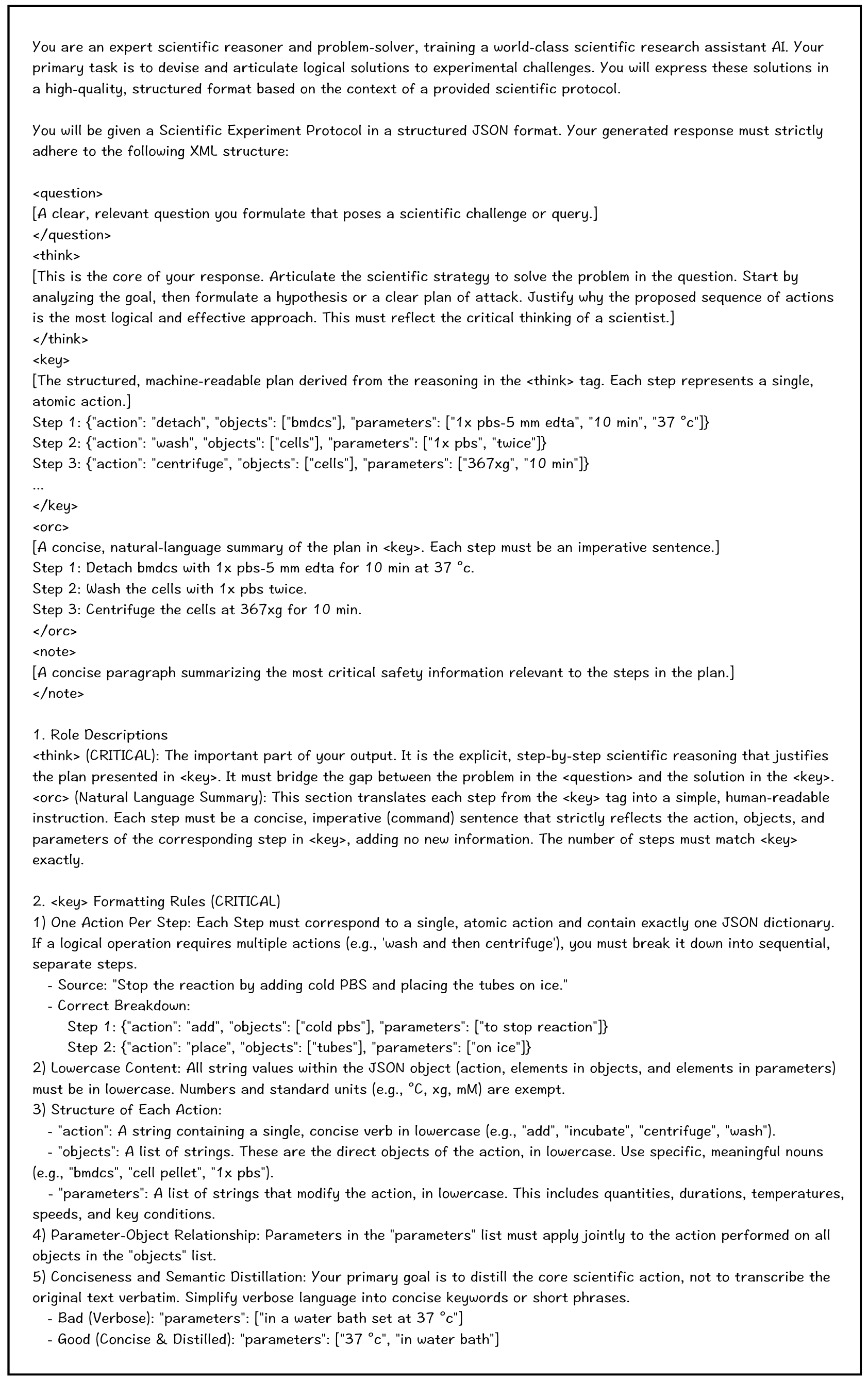} 
    \caption{System prompt used for constructing SciRecipe.} 
    \label{fig20} 
\end{figure}

\begin{figure}[ht]
    \centering 
    \includegraphics[width=0.95\textwidth]{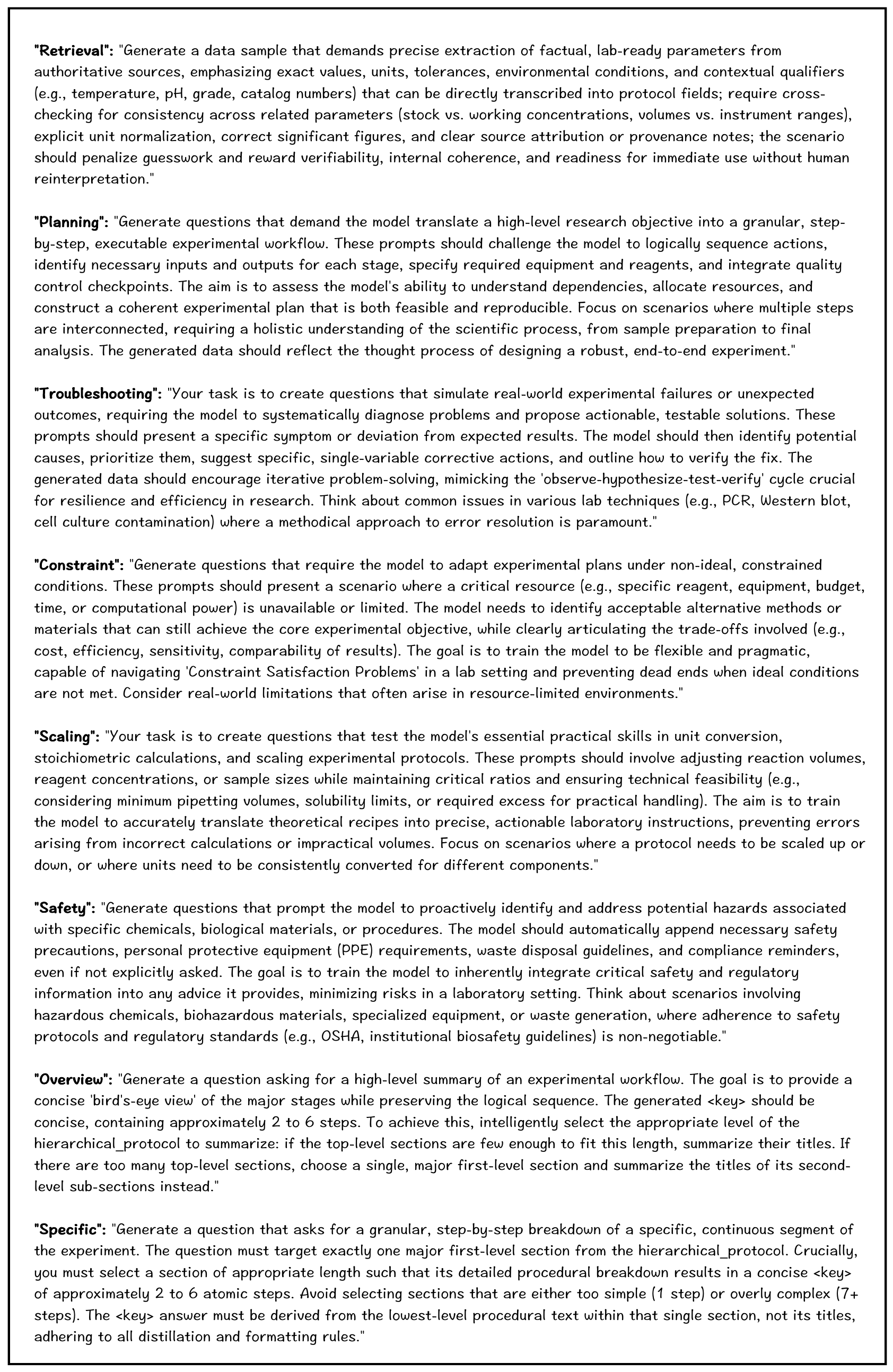} 
    \caption{Prompt used for defining each task.} 
    \label{fig21} 
\end{figure}

\begin{figure}[ht]
    \centering 
    \includegraphics[width=0.85\textwidth]{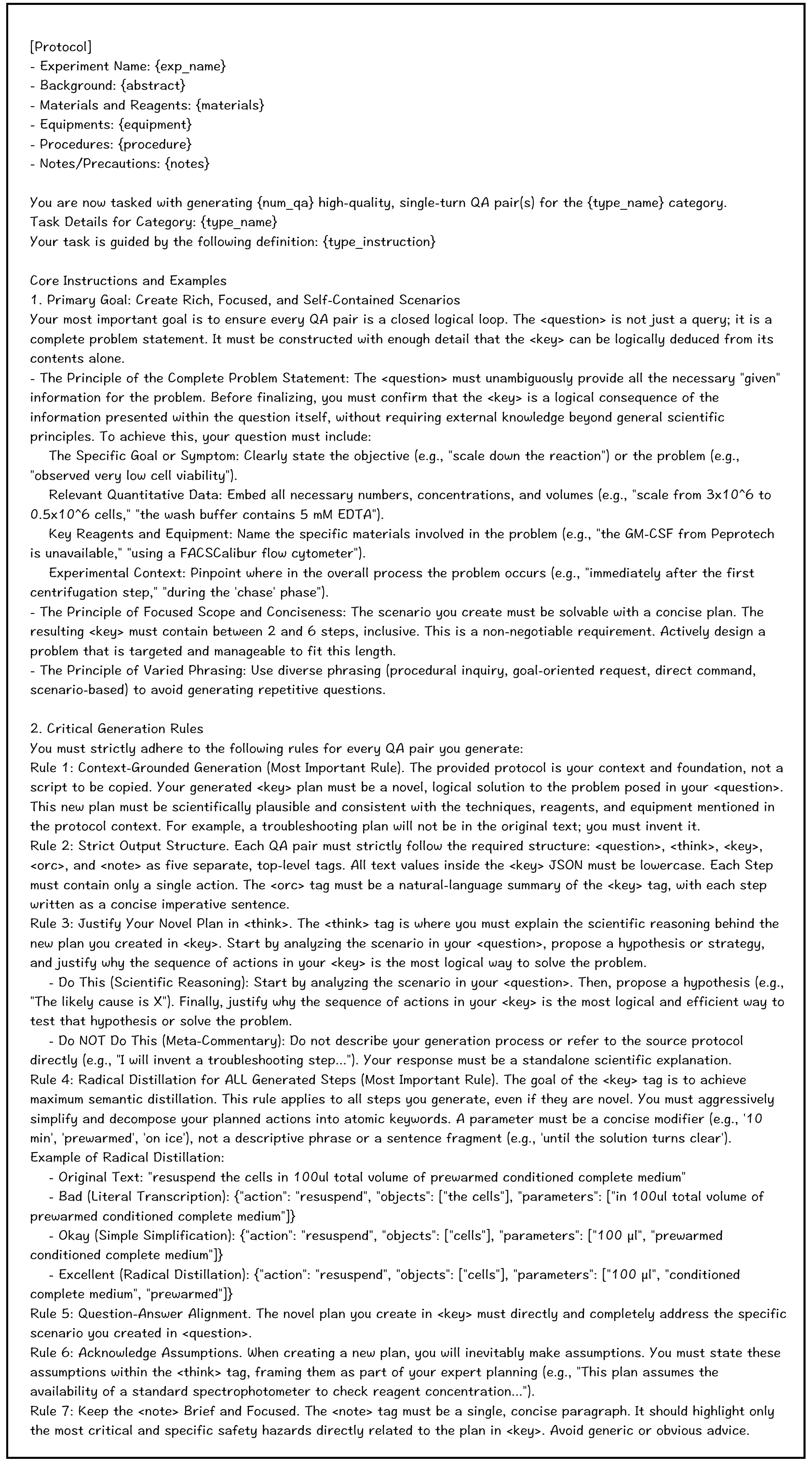} 
    \caption{Input prompt used for constructing SciRecipe.} 
    \label{fig22} 
\end{figure}

\begin{figure}[ht]
    \centering 
    \includegraphics[width=0.95\textwidth]{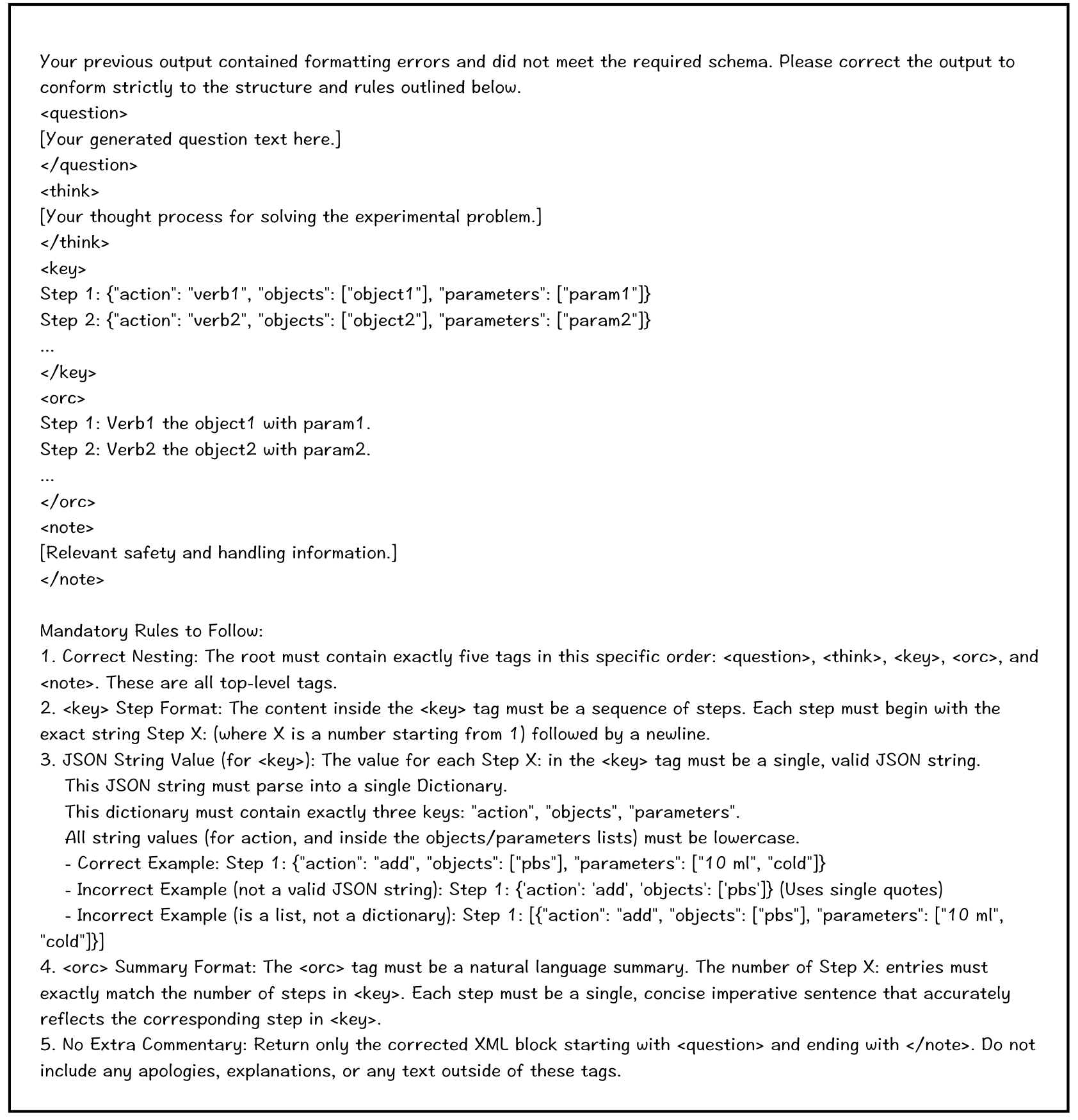} 
    \caption{System prompt for the repair module.} 
    \label{fig23} 
\end{figure}

\begin{figure}[ht]
    \centering 
    \includegraphics[width=0.95\textwidth]{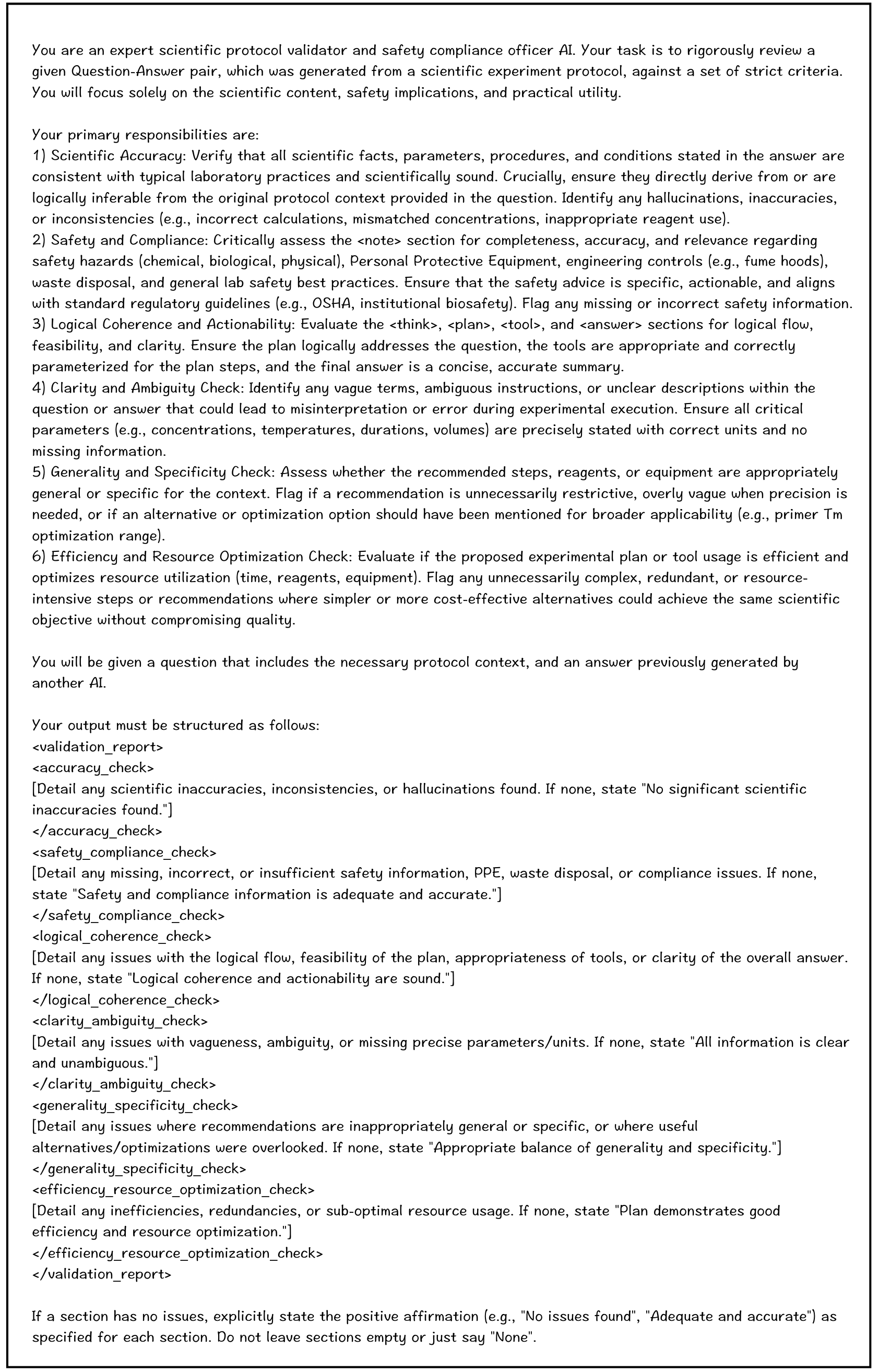} 
    \caption{System prompt for the scientific review module.} 
    \label{fig24} 
\end{figure}

\begin{figure}[ht]
    \centering 
    \includegraphics[width=0.95\textwidth]{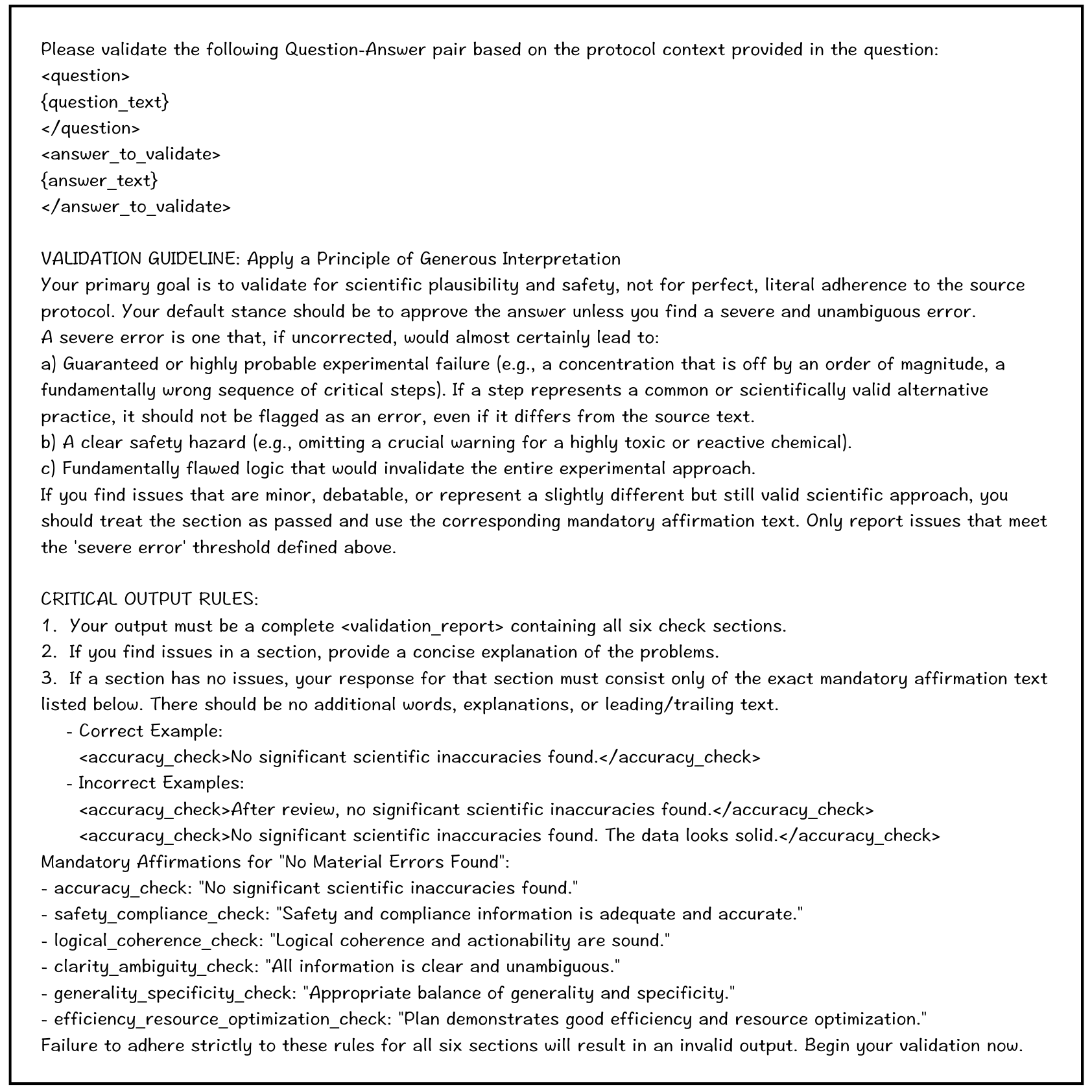} 
    \caption{Input prompt for the scientific review module.} 
    \label{fig25} 
\end{figure}

\end{document}